\documentclass[sigconf,screen]{acmart}

\usepackage{pifont}
\usepackage{microtype}
\usepackage{graphicx}
\usepackage{subfigure}
\usepackage{booktabs} %
\usepackage{listings} 
\usepackage{paralist}

\usepackage{amssymb}
\usepackage{amsmath}
\usepackage{empheq}
\usepackage{mdframed}
\usepackage[utf8]{inputenc}
\usepackage[english]{babel}
\usepackage{amsthm} 
\usepackage{xspace}
\usepackage{bm}
\usepackage{multirow}
\usepackage{array}
\usepackage{makecell}
\usepackage{longtable}
\usepackage{algorithm}
\usepackage[noend]{algorithmic}
\usepackage{graphicx} 
\usepackage{subcaption}
\usepackage{xcolor}
\usepackage{fvextra} 

\definecolor{lightgrey}{RGB}{240, 240, 240}

\usepackage{subcaption}
\usepackage{color, colortbl}
\usepackage{cellspace}
\usepackage{tikz}
\usetikzlibrary{shapes.geometric, positioning}

\usepackage{smartdiagram}
\usesmartdiagramlibrary{additions} 

\newcommand{\bmu}{\boldsymbol{\mu}}
\newcommand{\bSigma}{\mathbf{\Sigma}}

\newcommand{\outform}{{{\sc OutFormer}}\xspace}

\newcommand{\tabod}{{\sc TabPFN-OD}\xspace}

\theoremstyle{definition}

\newcommand{\hide}[1]{}

\makeatletter
\newcommand\footnoteref[1]{\protected@xdef\@thefnmark{\ref{#1}}\@footnotemark}
\makeatother

\renewcommand{\algorithmicrequire}{\textbf{Input:}}
\renewcommand{\algorithmicensure}{\textbf{Output:}}
\renewcommand{\algorithmiccomment}[1]{\hfill$\blacktriangleright$ #1}

\newcommand{\cbit}{\begin{compactitem}}
\newcommand{\ceit}{\end{compactitem}}
\newcommand{\cben}{\begin{compactenum}}
\newcommand{\ceen}{\end{compactenum}}

\newcommand{\beq}{\begin{equation}}
	\newcommand{\eeq}{\end{equation}}

\newcommand{\beqn}{\begin{equation*}}
\newcommand{\eeqn}{\end{equation*}}

\newcommand{\bit}{\begin{itemize}}
	\newcommand{\eit}{\end{itemize}}
\newcommand{\ben}{\begin{enumerate}}
	\newcommand{\een}{\end{enumerate}}

\newcounter{x}

\newcommand{\bx}{\mathbf{x}}

\definecolor{softorange}{HTML}{f4de96}
\definecolor{softblue}{HTML}{add7f6}

\newcommand{\silver}[1]{ {\colorbox{softblue!30}{{#1}}}}
\newcommand{\gold}[1]{ {\colorbox{softorange!30}{  #1}}}

\newcommand{\yessymb}{\textcolor{darkgreen}{\ding{51}}}
\newcommand{\nosymb}{\textcolor{red}{\ding{55}}}
\newcommand{\maybesymb}{\textcolor{orange}{(\ding{51})}}

\newcommand{\bench}{{\sc macrOData}\xspace}
\newcommand{\adb}{{ADBench}\xspace}

\newcommand{\fomo}{{\sc FoMo-0D}\xspace}
\newcommand{\dte}{{\sc DTE}\xspace}
\newcommand{\dtenp}{{\sc DTE-NP}\xspace}

\definecolor{darkgreen}{RGB}{0,139,0}
\definecolor{darkblue}{RGB}{0,0,139}
\definecolor{darkred}{RGB}{139,0,0}
\newcommand{\fraudbench}{\textsf{OddBench}\xspace}
\newcommand{\oddbench}{\fraudbench}
\newcommand{\onevsrestbench}{\textsf{OvRBench}\xspace}
\newcommand{\ovrbench}{\onevsrestbench}
\newcommand{\synbench}{\textsf{SynBench}\xspace}

\newcommand{\avg}{$^{\text{avg}}$}

\definecolor{best}{RGB}{255,220,180}
\definecolor{second}{RGB}{210,225,245}
\newcommand{\pmv}[2]{#1$_{\scriptsize\pm\,#2}$}

\newcommand{\lmn}[1]{{\color{blue}#1}}
\newcommand{\smn}[1]{{\color{purple}#1}}
\newcommand{\xueying}[1]{{\color{olive}#1}}
\newcommand{\todo}[1]{\textbf{\color{red}TO-DO: #1}}
\AtBeginDocument{%
	\providecommand\BibTeX{{%
			\normalfont B\kern-0.5em{\scshape i\kern-0.25em b}\kern-0.8em\TeX}}}

\copyrightyear{2026}
\acmYear{2026}
\setcopyright{acmlicensed}

\usepackage{xcolor}      
\usepackage{hyperref}     %

\definecolor{ACMPurple}{HTML}{003B8D}
\definecolor{ACMRed}{HTML}{c71111}
\definecolor{ACMDarkBlue}{HTML}{000000}
\definecolor{ACMBlack}{HTML}{000000}

\copyrightyear{2026}
\acmYear{2026}
\setcopyright{cc}
\setcctype{by}
\acmConference[KDD '26]{Proceedings of the 32nd ACM SIGKDD Conference on Knowledge Discovery and Data Mining V.2}{August 09--13, 2026}{Jeju Island, Republic of Korea}
\acmBooktitle{Proceedings of the 32nd ACM SIGKDD Conference on Knowledge Discovery and Data Mining V.2 (KDD '26), August 09--13, 2026, Jeju Island, Republic of Korea}
\acmDOI{10.1145/3770855.3817520}
\acmISBN{979-8-4007-2259-2/2026/08}

\begin{document}
\title{\bench: New Benchmarks {of Thousands of Datasets} for Tabular Outlier Detection}

\author{Xueying Ding}
\affiliation{%
	\institution{Carnegie Mellon University}
    \city{Pittsburgh}
	\country{USA}
}
\email{xding2@andrew.cmu.edu}

\author{Simon Kl\"{u}ttermann}
\affiliation{%
	\institution{Technical University of Dortmund}
    \city{Dortmund}
	\country{Germany}
}
\email{simon.kluettermann@cs.tu-dortmund.de}

\author{Haomin Wen}
\affiliation{%
	\institution{Carnegie Mellon University}
    \city{Pittsburgh}
	\country{USA}
}
\email{wenhaomin.whm@gmail.com}

\author{Yilong Chen}
\affiliation{%
	\institution{Carnegie Mellon University}
    \city{Pittsburgh}
	\country{USA}
}
\email{yilongch@andrew.cmu.edu}

\author{Leman Akoglu}
\affiliation{%
	\institution{Carnegie Mellon University}
    \city{Pittsburgh}
	\country{USA}
}
\email{lakoglu@andrew.cmu.edu}

\begin{abstract}

Quality benchmarks are essential for fairly and accurately tracking scientific progress and enabling practitioners to make informed methodological choices.
Outlier detection (OD) on tabular data underpins numerous real-world applications, yet existing OD benchmarks remain limited. 
The prominent OD  benchmark \adb \cite{han2022adbench} is the \textit{de facto} standard in the literature, yet comprises only 57 datasets. In addition to other shortcomings discussed in this work, its small scale severely restricts diversity and statistical power.
We introduce \bench, a large-scale benchmark suite for tabular OD comprising three carefully curated components: \oddbench, with 790 datasets containing real-world semantic anomalies; \ovrbench, with 856 datasets featuring real-world statistical outliers; and \synbench, with 800 synthetically generated datasets spanning diverse data priors and outlier archetypes.
Owing to its scale and diversity, \bench enables comprehensive and statistically robust evaluation of tabular OD methods.
Our benchmarks further satisfy several key desiderata: We provide standardized train/test splits for all datasets, 
public/private benchmark partitions with held-out test labels for the latter reserved toward an online leaderboard, and annotate our datasets with semantic metadata.
We conduct extensive experiments across all benchmarks, evaluating a broad range of OD methods 
 comprising classical, deep, and foundation models, over diverse hyperparameter configurations. We report detailed empirical findings, practical guidelines, as well as individual performances as references for future research. All benchmarks containing 2,446 datasets combined are open-sourced, along with a publicly accessible leaderboard hosted at \url{https://huggingface.co/MacrOData-CMU}.

\hide{
OD is important.
For years, OD algorithms were evaluated only on a \textbf{few}, \textbf{outdated} (KDD'99 \cite{}, UCI) and/or \textbf{arbitrary} (sometimes private) datasets, attributing it to anomalies being rare and such datasets not being public. 
This created a ground for \textit{not being able to track progress}.
Further, it
was exacerbated by hyperparameter hacking---algos are sensitive to HPs yet OD is unsupervised, leading to 
 \textit{unfair} evaluations.
The current de facto OD benchmark is ADBench, with only 57 datasets. Lots of shortcomings. But mainly its \textbf{size} is the biggest limitation: does offer  comprehensive evaluation (in terms of types of outliers as we show), not offering opportunities for meta-learning or FM pretraining. 
We introduce \bench, a repository of three new benchmarks for  tabular outlier detection. 
Our benchmarks are composed of both public and private testbeds. The latter maintain blind test sets whose labels are not released but accessible only to the organizers for evaluation. \todo{\textbf{Xueying}: We set up a submission site and leaderboard?}
}

\end{abstract}



\maketitle

\section{Introduction}
\label{sec:intro}

Reliable benchmarks are fundamental to accurately measuring scientific progress and enabling fair, reproducible comparison of methodologies. This is especially true for outlier detection (OD), a problem that arises across a wide range of high-impact real-world applications, such as fraud detection, healthcare, and environmental monitoring. 
Despite many methodological advances, progress in tabular OD has been difficult to assess due to the lack of large, diverse, and standardized benchmarks. Evaluation is largely centered around a small collection of datasets \cite{emmott2015meta,rayana2016sequential,Rayana2016ODDS,campos2016evaluation,steinbuss2021benchmarking,pang2021deep}, with the \textit{de facto} benchmark \adb \cite{han2022adbench} comprising only 57 datasets. This limited scale constrains dataset diversity, reduces statistical power,  and makes empirical comparisons highly sensitive to dataset selection, increasing the risk that reported improvements reflect benchmark idiosyncrasies rather than genuine methodological advances. These issues are further exacerbated by the hyperparameter sensitivity of OD methods \cite{ma2023need,ding2022hyperparameter,journals/tmlr/YooZA23} and the unsupervised nature of the problem, where tuning choices can substantially impact performance, yet are rarely evaluated in a statistically robust manner.
Together, these factors hinder fair evaluation, confound assessments of true progress, and complicate the reliable adoption of OD methods in practice.

While the small scale of the popular \adb readily limits its ability to capture the full spectrum of real-world inlier/outlier distributions, our detailed analysis (Sec. \ref{sec:related}) uncovers a specific limitation: \adb predominantly captures global outliers that resemble Gaussian noise. This idiosyncrasy enables simple, global distance-based methods such as KNN \cite{knn} to achieve state-of-the-art performance on \adb \cite{LivernocheJHR24,shen2025fomod}.
This and other limitations discussed in our work  underscore the need for more diverse and larger scale benchmark suites.

\begin{table*}
\centering
\caption{Comparison of Tabular Outlier Detection Benchmarks. Besides its scale with 2,446 datasets combined, \bench's three benchmarks -- \fraudbench, \onevsrestbench, \synbench -- compare favorably w.r.t. key desiderata: \underline{\#Datasets},  \underline{\#Samples and \#Dim.s}: range of number of points and features within datasets; \underline{Real/Syn. outliers}: whether the outliers are from the real world or synthesized; \underline{Seman./Stat. outliers}: whether the outliers are semantic anomalies (fraud, fault, etc.) or statistical deviations; \underline{Curation filters}: whether datasets are extensively vetted and filtered based on beyond-basic criteria;  \underline{Train/Test splits}: whether dataset provides standard splits; \underline{Public/Private parts}: whether benchmarks both publicly share as well as privately retain the test labels of datasets, the latter part used for maintaining a leaderboard; \underline{Semantic tags}: whether datasets contain semantic metadata and/or tags; \underline{Extensive HP conf.}: whether extensive hyperparameter configurations are employed for evaluation.}
\vspace{-0.1in}
 \resizebox{\textwidth}{!}{%
\begin{tabular}{lrrrrccccccc}
\toprule
 & \textbf{Year} & \textbf{\#Data}  & \textbf{\#Samples} & \textbf{\#Dim.s} & \textbf{Real/Syn.} & \textbf{Seman./Stat.} & \textbf{Curation}   & \textbf{Train/Test} & \textbf{Pub./Priv.} & \textbf{Semantic} & \textbf{Extensive }\\
\textbf{Benchmark} &  & \textbf{sets} & \textbf{range} & \textbf{range} & \textbf{outliers} & \textbf{outliers}  & \textbf{filters} &  \textbf{splits}  &\textbf{parts}  & \textbf{tags} & \textbf{HP conf.} \\
\midrule
Emmott \textit{et al.} \cite{emmott2015meta} & 2015 & 20 & [1000,6000] & [1,200] & Real & Statistical & \yessymb & \nosymb & \nosymb & \nosymb & \maybesymb \\
ODDS \cite{rayana2016sequential,Rayana2016ODDS} & 2016 & 31 & [148,7063] & [6,274] & Real & Mixed & \nosymb & \nosymb & \nosymb & \nosymb & \nosymb \\
DAMI \cite{campos2016evaluation} & 2016 & 23 & [80,60632] & [3,1555] & Real & Mixed & \nosymb & \nosymb & \nosymb & \nosymb & \yessymb \\
Steinbuss\&B\"{o}hm \cite{steinbuss2021benchmarking} & 2021 & 19 & [80,60632] & [5,38] & Synthetic & Statistical & \nosymb & \nosymb & \nosymb & \nosymb & \nosymb \\
ADBenchmarks \cite{pang2021deep} & 2021 & 21 & [405,619326] & [6,10000] & Real & Mixed & \yessymb & \nosymb & \nosymb & \nosymb & \nosymb \\
ADBench \cite{han2022adbench} & 2022 & 57 & [80,619326] & [3,1555] & Real & Mixed & \nosymb & \nosymb & \nosymb & \nosymb & \nosymb \\
\midrule 
\fraudbench [ours] & 2026 & 790 & [1000,1012180] & [3,1024] & \silver{Real} & \silver{Semantic} & \yessymb & \yessymb & Both & \yessymb & \yessymb \\
\onevsrestbench [ours] & 2026 & 856 & [1000,2215023] & [2,10935] & \silver{Real} & \gold{Statistical} & \yessymb & \yessymb & Both & \yessymb & \yessymb \\
\synbench [ours] & 2026 & 800 & [1000,6000] & [2,100] & \gold{Synthetic} & \gold{Statistical} & \yessymb & \yessymb & Pub. & \yessymb & \yessymb \\
\bottomrule
\end{tabular}
}
\label{tab:comparison}
\end{table*}

\begin{figure}[!ht]
    \centering
    \includegraphics[width=0.9 \linewidth]{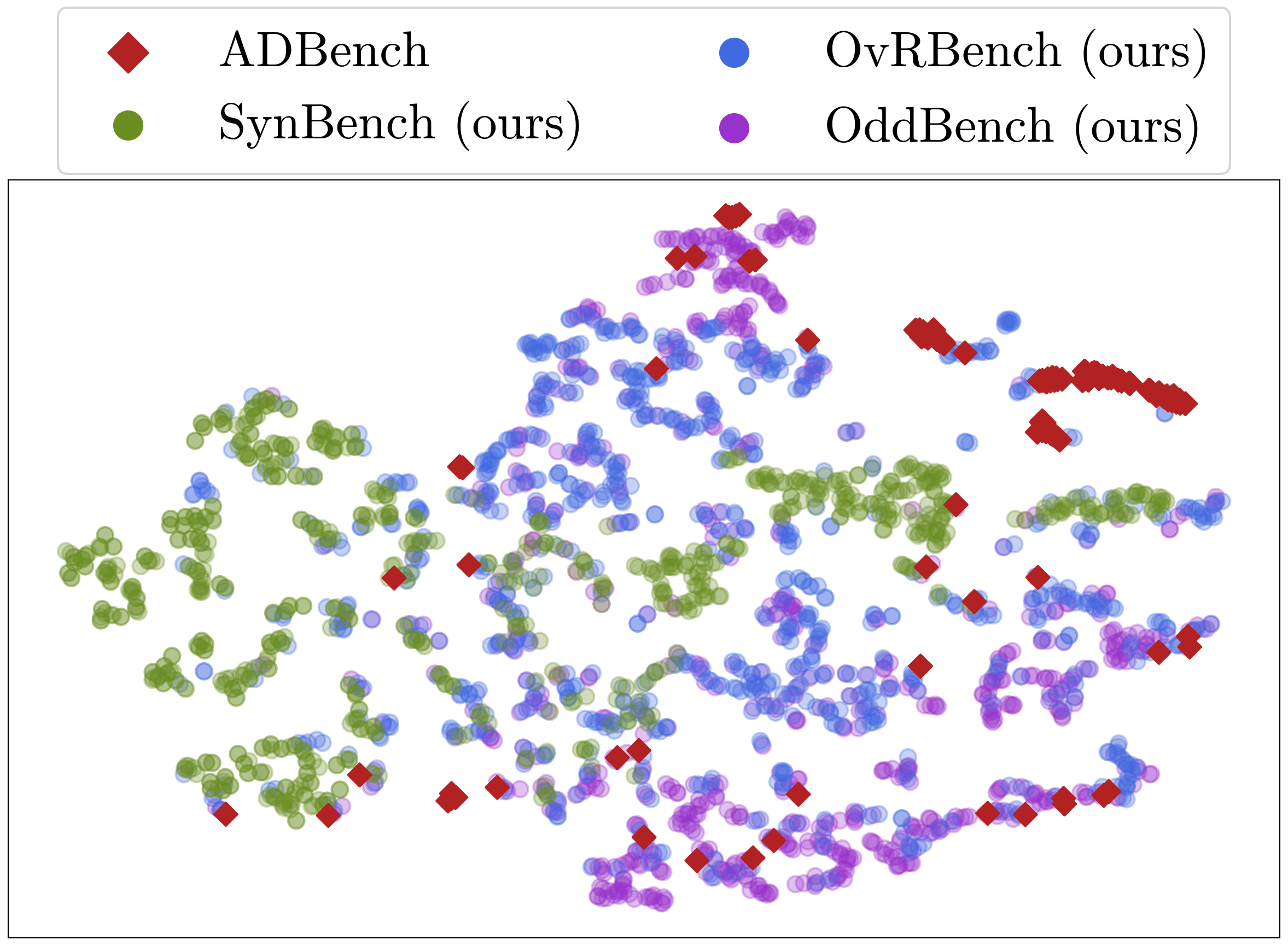}
    \vspace{-0.1in}
    \caption{t-SNE embedding of datasets (based on dataset-level metafeatures \cite{Vanschoren2018MetaLearningAS}) across three \bench benchmarks (ours) and from \adb \cite{han2022adbench} show that \bench offers significant boost in both scale and diversity. (best in color)}
    \label{fig:crownjewel}
    \vspace{-0.15in}
\end{figure}

To fill these gaps, we introduce \bench, a comprehensive suite of large-scale OD benchmarks. \bench sifts through 627 million tables in Tablib \cite{eggert2023tablib}, alongside other tabular sources, and carefully curates three distinct OD benchmarks (Sec. \ref{sec:new}): (1) \oddbench, containing 790 datasets with real-world semantic anomalies (fraud, fault, etc.) from diverse domains; (2) \ovrbench, which includes 856 datasets featuring real-world statistical outliers derived from One-vs-Rest repurposing of classification tasks; and (3) \synbench, with 800 synthetically generated datasets spanning diverse data priors and outlier archetypes. By incorporating multiple data sources, outlier types, and generation methods,  \bench  significantly improves both scale and diversity (Figure \ref{fig:crownjewel}), enabling more thorough and statistically robust evaluations. It also supports 
research directions previously limited by small-scale data, including meta-learning \cite{zhao2021automatic,zhao2022toward} and foundation models for OD \cite{shen2025fomod,tabpfnRW}.

Beyond scale and diversity, \bench incorporates several unique features that enhance its rigor and utility for the research community (Table \ref{tab:comparison}). First, we provide {standardized train/test splits} for each dataset, ensuring consistent evaluation across experiments. Second, we introduce {public/private benchmark partitions}, and withhold the private split test labels  to host an online OD leaderboard, enabling ongoing and fair competition. Third, we annotate datasets with {semantic metadata}, improving interpretability and providing deeper insights into the nature of the outliers. These features make \bench a powerful tool for benchmarking OD methods and a valuable resource for future research.

Finally,  we conduct extensive experiments on \bench across classical, deep, and foundation models for OD, systematically exploring a wide range of hyperparameter configurations. This large-scale evaluation provides a comprehensive assessment, highlighting statistically significant performance differences and providing insights for both researchers and practitioners.

The following summarizes our main contributions.
\makeatletter
\setlength{\leftmargini}{0.2in}
\setlength{\leftmarginii}{0.3in}
\setlength{\leftmarginiii}{0.2in}
\setlength{\leftmarginiv}{0.3in}
\makeatother
\begin{itemize}
\item \textbf{``Demystifying'' \adb:}
 Despite its small scale, which limits its representativeness and statistical power for evaluation, \adb has become the de facto standard for OD. We critically analyze its  limitations, revealing that it primarily captures global outliers resembling Gaussian noise.

\item \textbf{New Large-scale OD Benchmarks:}
We introduce \bench, a comprehensive, open-source repository featuring 2,446 datasets from three distinct testbeds: \fraudbench, with real-world datasets containing semantic anomalies; \onevsrestbench, with real-world datasets containing statistical outliers; and \synbench, containing synthetically generated datasets from diverse priors and outlier archetypes. Owing to its scale, diversity, train/test splits, public/private partitions, and rich metadata, \bench meets key desiderata, providing a standardized and robust foundation for OD evaluation.

\item \textbf{Evaluation and Practical Guidelines:~} 
We perform extensive evaluations across a wide range of OD methods and hyperparameter configurations, revealing statistically significant performance differences and offering actionable insights for future OD research and practice.

\item \textbf{Frontiers of OD and Future Opportunities:} 
\bench establishes a rich playground for exploring the future frontiers of OD, including meta-learning and foundation models for OD.

\end{itemize}

\noindent
\textbf{Availability:} We open-source (1)--(5), and host an online OD leaderboard  at \url{https://huggingface.co/MacrOData-CMU}:
(1) \textbf{Public datasets} (690+756+800) from \oddbench, \ovrbench, \synbench, plus  performance comparisons of prominent OD methods, 
(2) 
\textbf{Private datasets} (100+100 from \oddbench and \ovrbench) (w/out Test labels), plus Leaderboard metrics,
(3) \textbf{Code} for 
dataset curation with detailed filtering criteria,  
 (4) \textbf{Code} for dataset construction (given a table, without target; convert to OD task), and (5) \textbf{Individual performance results} of all methods on all 2,446 datasets.



\hide{
\textbf{Why New Benchmarks?}

ADBench 
	[Size] is small - 57 datasets 
[Representativeness (of reality + various known types)] 
	lacks diversity - kNN does well, Gaussian noise captures well
not sufficiently differentiating existing algos
[Tabular or not tabular? – embeddings ]

\textbf{Contributions.~}

“Demystifying ADBench”: 
Deep-dive on ADBench, back-tracking from SOTA: DTE-NP and Fomo (and why it does so well with GMM-only prior)
DTE-NP outliers are simply Gaussian-noise/high variance outliers – just like Fomo

Types of Outliers ADBench does NOT represent
And how SOTA fails on those Benchmarking Unsupervised Outlier Detection with Realistic Synthetic Data “With outliers in the dependency ..  (vine-vine), isolation Forest, “lof=100” and “wknn-100” do not perform well overall. Their AUC PR is close to 0 (!!!).

New public testbeds
Synthetic Corpora 
w/ known and varying anomaly types

Real testbeds - large, independent
	2.1) Real-one, Real-subsample-rest AT SCALE 
	2.2) Real normal, Real fraud,fault,error,bug… (maybe subsampled for imbalance)
w/ taxonomy:  Real-life Scenario: Adversarial / not (Planned/Unplanned)
		Outlier Type: Contextual / subspaces / sparse; 
Global/local; Dependency
		Tabular Representation: Embedding / Handcrafted / Raw
}

\hide{
Future opportunities:  Implications on tabular foundation models for OD
Supervised pretraining , thanks to :
Reliable labels (and size)
High diversity for better generalization: thanks to 
Size - no longer miniscule and/or Mix synthetic prior design
Open taxonomy
}

\hide{
As ADBench has become the de facto standard benchmark for OD, we dive deep into better understanding its composition and limitations in terms of its representativeness of real world outliers.
Through analyzing the inner-workings of the SOTA detector on ADBench, namely DTE-NP \cite{LivernocheJHR24} and its close rivals KNN \cite{knn} and \fomo \cite{shen2025fomod}, we characterize ADBench to exhibit global outliers that align with Gaussian noise, exposing the types of outliers it does not represent. \xueying{for onevrestbench and oddbench, kNN and DTENP still is the SOTA?}
}

\hide{
We introduce a new repository called \bench\footnote{\url{https://huggingface.co/MacrOData-CMU}} with a compilation of thousands of datasets. \bench consists of 3 testbeds: \fraudbench contains real-world datasets with pre-existing anomalies, \onevsrestbench has real-world points, some repurposed as outliers, and SynPriorBench contains synthetic datasets generated from various distinct data priors capturing diverse data distributions (multi-modal, skewed, etc.) as well as diverse types and severity of outliers (global, local, dependency, contextual, extreme-value, etc.).  
\smn{From Simon: Versions with private labels: e.g. \cite{lhcOlympics,MVTec2,roechnerPosition}}.
\todo{Leman: Highlight 1) scale, 2) diversity, 3) characterization; tags, 4) Settings; splits, 5) public-private: former allows meta-learning, latter provides fair record of progress}
}

\hide{
We compare top-performing OD algorithms on \bench, aiming to better quantify their performance and robustness across diverse distributions.  
\smn{From Simon: large-scale comparisons allow us to capture statistically significant differences.}
}

\hide{
\todo{Leman: Cover hot topics in OD, and how our benchmarks can assist} Our \bench offers several future opportunities. Such a large repository of labeled OD datasets unlocks supervised pretraining of OD foundation models on real datasets. It also facilitates leveraging (part of) these datasets for validation, hyper-parameter tuning, and various meta-learning approaches that can transfer learnings to new unlabeled tasks. 
}


\section{Related Work and ADBench}
\label{sec:related}

\subsection{A Brief  History on OD}
\label{ssec:history}


Outlier detection (OD) finds numerous real world applications across domains (finance, medicine, security, to name a few). Therefore, there exists a vast body of detection algorithms, from classical \cite{books/sp/Aggarwal2013,hodge2004survey} to deep architectures \cite{chalapathy2019deep,pang2021deep}. The plethora of detectors raises the question of which one to use on a new task  \cite{whichtouse18}.
Selecting an effective OD algorithm (\textbf{and} its hyperparameter values), however, is a challenging task. First, there is often no labeled data for model selection. Second, the track record of progress in the field is not reliable due to limited benchmarks and lack of governing over hyperparameter (HP) configuration, where manual dataset selection and manual HP tuning  plague the results, raising concerns for fair evaluation \cite{ding2022hyperparameter,journals/tmlr/YooZA23}. 

Various work in the OD literature pointed to the HP sensitivity of both classical methods \cite{aggarwal2015theoretical,goldstein2016comparative,ma2023need} and deep learning based methods with even more HPs \cite{ding2022hyperparameter}, offering workarounds, such as reporting performance averaged across multiple HPs \cite{goldstein2016comparative} or employing unsupervised model selection heuristics  \cite{zhao2021automatic,ding2023fast,ma2023need}. 
Several unsupervised hyperparameter optimization (HPO) methods rely on meta-learning on historical data repositories with labels \cite{zhao2021automatic,zhao2022toward,ding2023fast},  which are also needed for reliable evaluation. 
More recent foundation models for OD pretrain on labeled datasets, another form of meta-learning \cite{shen2025fomod}.
These underscore the need for large-scale, new OD benchmarks \cite{akoglu2021anomaly}, not only to keep a reliable record of progress, but also to foster effective HPO and future foundation models.


\vspace{-0.05in}
\subsection{Existing OD Benchmarks}
\label{ssec:makeup}

Table \ref{tab:comparison} shows a list of existing OD benchmarks. 
First, they are only a handful. Most strikingly,  they are minuscule; containing only a few tens of datasets.\footnote{We list the number of ``mother'' datasets although some works derive different variants.}  
Beyond size, they lack crucial properties that are standard for modern-era benchmarks. We defer a  detailed discussion of these properties to Section \ref{sec:new}, and briefly mention that unlike proposed \bench, they do not exhibit standard Train/Test splits, extensive curation, semantic tags and metadata, and public/private variants with redacted test labels in the latter for fair, blinded testing. In addition, characteristics and types of  outliers in these benchmarks are limited compared to the diversity covered by the three  benchmarks in \bench.

Curiously, these benchmarks 
 do not compare in popularity: ADBench \cite{han2022adbench}, specifically, stands out as the \textbf{de facto} benchmark that is frequently used for evaluation in the literature. That is possibly because it is the most recent (2022) and the relatively largest (57). Moreover, it combines some of the other earlier benchmarks; borrowing datasets from ODDS (26) \cite{rayana2016sequential}, DAMI (19) \cite{campos2016evaluation}, Emmott \textit{et al.} (11) \cite{emmott2015meta}, DevNet (6) \cite{devnet}, while also including embeddings of (6) image and (5) text datasets as additional tabular datasets.\footnote{\adb is criticized for its embeddings of other modalities as tabular data \cite{roechnerPosition}.} Appx. \ref{appx:adbench_stats} provides summary statistics of \adb datasets. 
 
 Due to its disproportionate popularity and prominence in the literature,  Section \ref{sec:demystify} provides a critical lens into \adb.










\vspace{-0.05in}
\subsection{Demystifying ADBench}
\label{sec:demystify}

In 2024, \citet{LivernocheJHR24} introduced the \textit{state-of-the-art} OD algorithm on ADBench called \dte (for \underline{D}iffusion \underline{T}ime \underline{E}stimation), inspired by the latest developments in denoising diffusion.  At the time, \dte was shown to outperform a record-breaking number (23) of baselines, from traditional shallow algorithms to deep learning-based methods, including the original DDPM \cite{ho2020denoising}.
Yet, curiously, the basic KNN algorithm by \citet{knn} is the \textbf{second top} performer as reported in \citet{LivernocheJHR24}.

Most recently, in 2025 \citet{shen2025fomod} introduced \fomo, the first \underline{fo}undation \underline{mo}del for zero(\underline{0})-shot \underline{OD}, which rivaled \dte \cite{LivernocheJHR24} (across hyperparameters) and performed statistically no different from KNN \cite{knn} (2nd top performer), while being pretrained solely on \textbf{synthetic} datasets and requiring \textbf{no} additional model training or (the notorious) hyperparameter tuning on a new OD task.

These recent developments and curious observations raise key questions: 
\textit{What drives the strong performance of these new approaches on ADBench?} 
\textit{What makes \fomo achieve as competitive results on \adb, considering it was pretrained purely on synthetic datasets?}
\textit{How could the basic KNN remain competitive against these modern techniques} (that use diffusion and extensive pretraining)? \textit{What could these observations reveal about ADBench?}

Appx. \ref{appx:adbench_demystify} provides our deep-dive on the innerworkings of \dte and \fomo to address these questions, and sheds light on \adb. We brief our key findings and implications as follows.

\vspace{-0.025in}
\subsubsection{\bf Key Findings.~} The analysis of \dte's \cite{LivernocheJHR24} analytical form and its approximation in Appx. \ref{ssec:dtedive} establishes intriguing \textit{connections between denoising diffusion time, Gaussian noise variance and nearest neighbors}. Specifically, \dte aims to estimate the posterior distribution over diffusion time of a stochastic process that creates noisy samples over time following a diffusion process. Then, \dte considers diffusion time as a surrogate for the distance from the data manifold, for which they derive an analytical formula under Gaussian noise. 
Effectively, {DTE identifies data points as outliers if their probability is high(er) under a Gaussian distribution with a large(r) variance.}
In other words, the \textbf{assumed outliers mimic Gaussian noise, and resemble global outliers---with diffusion noise variance increasing  in the \textit{full} $d$-dimensional feature space}, unlike \textit{sub}-space outliers.\footnote{See Eq. \eqref{eq:closed} in Appx.  where co-variance matrix is a scaled \textbf{I}dentity matrix in $\mathbb{R}^{d\times d}$.}

\dte proceeds to approximate the analytical form that yields a function of the distance from a data point to its nearest neighbor (NN) in the dataset. In practice, \dte uses not only the smallest but the {squared} average distance to {KNN}s for better approximation, especially for larger time steps. Therefore, their non-parametric approach coined as \textbf{\dtenp produces a ranking (of data points by outlierness) 
\textit{identical} to that by average squared distance to the $k$ nearest neighbors, a classical OD approach}. Notably, the KNN algorithm by \citet{knn} is slightly different as it uses the largest, i.e. $k$\textit{th} NN distance as the outlier score instead of the average.

On the other hand, \fomo \cite{shen2025fomod} is pretrained on  datasets synthesized from a 
Gaussian mixture model (GMM) based data prior. Outliers are sampled from the same GMM used to synthesize the inliers, but with \textit{inflated variances}. 
Then, \textbf{\fomo models outliers as generated by Gaussian noise} similar to \dte that treats outliers as generated by a diffusion process under Gaussian noise. 

\vspace{-0.025in}
\subsubsection{\bf Take-aways.~} The peering beneath \dte and \fomo, and understanding their makeup and inner-working have allowed us to reveal two key findings. \textbf{First}, \dte internally models outliers as generated by Gaussian noise, governed by a diffusion process. This is similar to \fomo, which is pretrained on datasets with outliers simulated from Gaussian noise. \textbf{Second}, \dte  estimates the distribution over diffusion time (i.e. Gaussian noise variance) approximately using average squared distances. Together, these findings unify our understanding of why three seemingly different algorithms --- DTE-NP \cite{LivernocheJHR24}, KNN \cite{knn}, \fomo \cite{shen2025fomod} --- are similarly competitive, and achieve state-of-the-art performance on \adb.

\vspace{-0.075in}
\subsubsection{\bf Remarks on ADBench.~}
\label{ssec:remarks}


These connections allow us to draw key conclusions about \adb. First is the revelation that \adb likely exhibits datasets with outliers that align well with \textit{Gaussian noise}. In addition, these are most likely \textit{global} outliers---points that stand out across many input features---as \dte models  noise diffusion in \textit{all} dimensions and KNN measures distances across \textit{all} features indiscriminately.

These insights expose key limitations of \adb, beyond its scale and other desiderata highlighted in Table \ref{tab:comparison}, suggesting that stress-testing OD algorithms likely requires going beyond global outliers that merely resemble Gaussian noise. 

We also critique \adb with regard to its dataset composition. 
First, its representativeness is severely hampered by its small number of (57) datasets,  hindering reliable ablation studies across  varying dataset characteristics; for example, dimensionality, domain, etc. 
Among its small body of datasets, it includes near-duplicates (e.g., \texttt{Cardio} and \texttt{Cardiography}, or \texttt{Satellite} and \texttt{Satimage-2} or \texttt{BreastW} and \texttt{Mammography}), which may inadvertently overweigh specific data distributions. In contrast, 
standard datasets like \texttt{Arrhythmia} are omitted despite their prevalence in the literature. Overall, its dataset selection criteria appear somewhat arbitrary.
Curiously, it also contains \textit{non-tabular} data embeddings, which can be criticized \cite{roechnerPosition} as not representing traditional tabular datasets with often raw or hand-crafted features. 
Outside of embeddings, \adb is dominated by low-dimensional data; nearly 50\% of datasets have fewer than 20 features---biasing evaluation toward methods less affected by the curse of dimensionality.

In addition, the utility of several constituent datasets is hampered by difficulty levels: ``trivial'' sets like \texttt{BreastW}, \texttt{Thyroid}, \texttt{Fraud} allow most algorithms to achieve >90\% ROC AUC, introducing evaluation noise due to performance saturation, while "very hard" sets like \texttt{Yeast}, \texttt{Vertebral}, \texttt{Speech}, \texttt{IMDb} often result in near-random performance (<60\% ROC AUC), presenting an unrealistic, overly pessimistic evaluation scenario. 

Other concerns include data integrity and reliability; for example, the \texttt{Thyroid} dataset exhibits only five features compared to the standard six found in the UCI repository, and the \texttt{Wine} dataset's reliance on a mere 10 anomalies makes it statistically unreliable.

{Collectively, these observations necessitate a cautious interpretation of results derived from \adb. The benchmark’s lack of structural diversity and over-representation of narrow outlier characteristics limit its ability to reflect the complexity of real-world anomaly scenarios.
Our work aims to fill these gaps by introducing an extensive evaluation suite called \bench that prioritizes scale, diversity, and rigorous curation. Furthermore, we adhere to established evaluation protocols; including standardized train/test splits and the use of both public and private datasets, where the latter enables a blinded, impartial evaluation of OD performance.}






\hide{
\todo{Simon: Take the first stab at this? Criticize based on 2.2}

\begin{itemize}
    \item Somewhat arbitrary choice of dataset selection. Why e.g. not arrhythmia even though used in multiple sources
    \item duplicate datasets. cardio and Cardiography, breastw and Mammography, satellite and satimage-2
    \item except embeddings: most datasets are low dimensional (50\% of datasets have $<20$ features), biases analysis towards algorithms that are usually limited by curse of dim
    \item both trivial datasets (almost all algorithms reach $>90\%$ ROC AUC on breastw, thyroid, fraud) and very hard datasets (yeast, vertebral, speech, imdb $<60\%$ ROC AUC on most algorithms) \cite{dean}. trivial datasets $\rightarrow$ evaluation noise, very hard datasets $\rightarrow$ unrealistic evaluation scenario
    \item Limited number of dataset clusters: Does not allow for reliable ablation studies (e.g. "Works well on high dim") \xueying{not sure what this means?}\smn{all high dimensional datasets are mnist/mvtec etc}
    \item Some processing lost to time. Thyroid usually 6 feature, source only 5 https://archive.ics.uci.edu/dataset/102/thyroid+disease
    \item Some datasets very unreliable. e.g. wine has only 10 anomalies, also often trivial: Taking anoscore=first feature reaches $92\%$ ROC AUC, more depending on splits
    \item Similar some trivial. E.g. breastW. Using lumpsize feature as anomaly score reaches $>97\%$ ROC AUC
    \item very uneven distribution of datasets (Fig. ~\ref{fig:adbench_stats}). Largely because of multiple embedding datasets
\end{itemize}
}

\vspace{-0.05in}
\section{Proposed \bench Benchmarks}
\label{sec:new}


Our  \bench consists of three separate benchmarks: 
\textbf{(1) \fraudbench} with 790 \textit{real world} datasets associated with various abnormalities, thus containing \textit{semantic anomalies}, carefully curated from Tablib \cite{eggert2023tablib};
\textbf{(2) \onevsrestbench} with 856 \textit{real world}  datasets curated from Tablib \cite{eggert2023tablib} and  classification libraries (TabZilla \cite{mcelfresh2023neural},  TabRepo~\cite{salinas2024tabrepo}, TabArena~\cite{erickson2025tabarena}, etc.), containing \textit{statistical outliers};
and  \textbf{(3) \synbench} with 800 \textit{synthetic} datasets generated from five  different data priors with \textit{diverse known outlier types}.

\bench provides \textbf{($i$) scale}; with over two thousand datasets in total\footnote{Respecting possible resource constraints for benchmarking, we provide 50 representative datasets each, for \fraudbench, \onevsrestbench and \synbench; see Sec. \ref{ssec:represent}.} that is two orders of magnitude larger than existing OD benchmarks with only several tens of datasets, and \textbf{($ii$) diversity}\footnote{We show both a high semantic diversity; see Fig.~\ref{fig:word_cloud_oddbench} and Fig.~\ref{fig:word_cloud_ovrbench}; and a high diversity in dataset characteristics; see Appx.~\ref{appx:fraudbench_distribution}, Fig.~\ref{fig:NvsD}.}; two real world benchmarks span multiple domains with various semantic tags\footnote{We augment each real world dataset with semantic tags, which the practitioners can leverage to scope their evaluation; see Sec. \ref{ssec:metadata}.} and one synthesized benchmark contains 
datasets with diverse inlier distributions and different known outlier types. 
Each dataset comes with a \textbf{standard Train/Test split}\footnote{We designate random 50\% of  inliers to Train, and  rest 50\% inliers plus all outliers to Test. Each dataset includes only one Train/Test split; however, the large number of single-split datasets offsets the effects of particularly un/lucky splits.}; with inlier-only training data. This enables \textbf{both one-class (inductive) and unsupervised (transductive) detection} (with Train/Test combined in the latter). 
Furthermore, we provide both a \textbf{Public\&Private benchmark}\footnote{We selected 100 random datasets  as private from the real-world benchmarks \fraudbench and \onevsrestbench, and publicly share the rest, 690 and 756 datasets respectively.} of the real-world \bench benchmarks, where private datasets' test labels are redacted on our end. These private test labels are used to maintain an \textbf{online leaderboard}
of detectors, enabling a fair and consistent record of progress tracking.\footnote{As \bench is sourced from public repos, we take preventive steps against reverse-engineering private dataset identities, and thereby their test labels; see Sec. \ref{ssec:deanon}.}

\begin{figure}[!t]
    \centering
    \includegraphics[width=1.0 \linewidth]{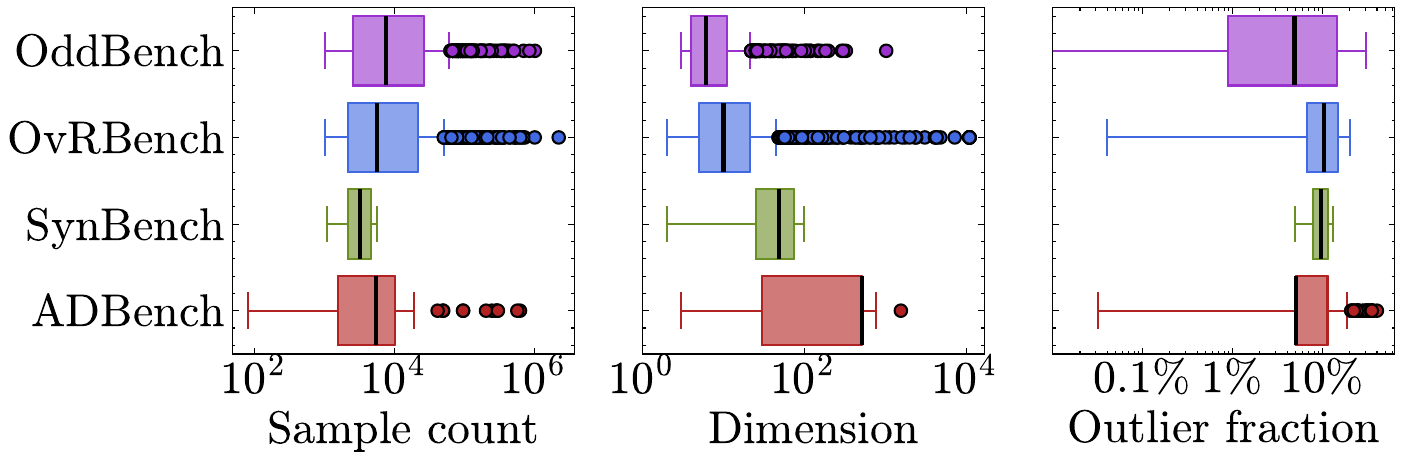}
    \vspace{-0.25in}
    \caption{Summary statistics of \bench vs. ADBench.}
    \label{fig:summary}
    \vspace{-0.15in}
\end{figure}

\bench improves upon \adb in representativeness (Fig. \ref{fig:crownjewel}, Fig. \ref{fig:summary}) and introduces key innovations for the first time (Table \ref{tab:comparison}), establishing a comprehensive benchmark that is designed to effectively assess and actively foster progress in the field of OD\footnote{Supervised algorithms significantly outperform unsupervised algorithms on our benchmarks, signaling large possible improvements; see Appx.~\ref{appx:fraudbench_distribution}, Fig.~\ref{fig:RFperformance}.}.

\begin{figure*}
\vspace{-0.1in}
    \centering
    \includegraphics[width=0.95\linewidth]{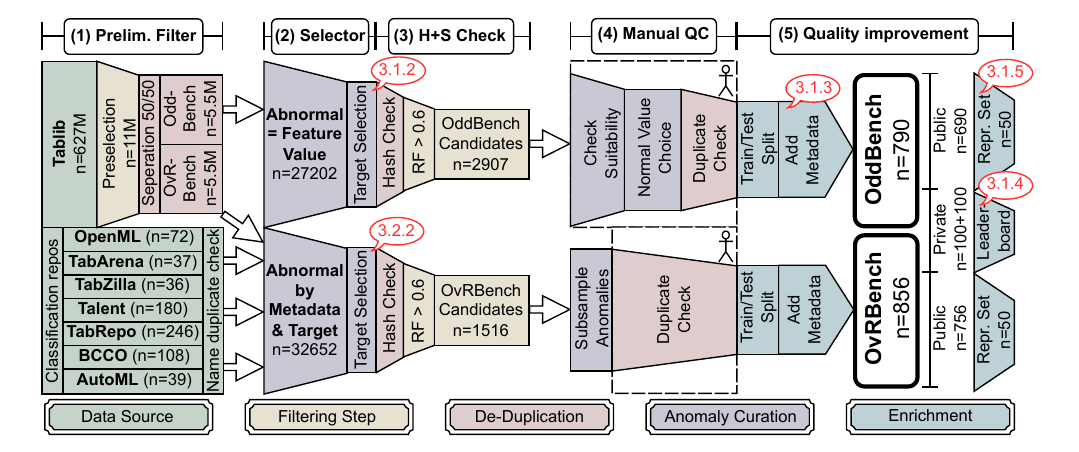}
    \vspace{-0.2in}
    \caption{Overview of curation steps used to build \oddbench and \ovrbench.}
    \label{fig:overview}
      \vspace{-0.1in}
\end{figure*}



\vspace{-0.05in}
\subsection{\oddbench} \label{sec:fraudbench}

\subsubsection{\textbf{Filtering Criteria}}
\label{ssec:filter}


The proposed \fraudbench is based on TabLib~\cite{eggert2023tablib}, which contains 627 million real-world tabular datasets with abundant metadata, extracted from numerous file formats, including CSV, HTML, SQLite, PDF, Excel, and others, sourced from GitHub and Common Crawl. 
We use half of TabLib to build \fraudbench, while the other half is used to build \onevsrestbench. Specifically, \fraudbench contains a subset of TabLib where 
 positive samples are associated with \textbf{semantic anomalies} in the real world. 
 We take the following steps to build \fraudbench, as outlined in Figure~\ref{fig:overview}.


  \vspace{0.025in}
(1) \textit{Preliminary Filter: } Tablib  contains raw Web-sourced tables. First, we remove all clearly unusable datasets; for each raw table, we drop the columns with ratio of null or infinity values above $25\%$, and any rows that contain null or infinity values. We also drop all monotonous features that might indicate indices. Next, we drop all datasets that contain fewer than $1000$ samples or $3$ numerical features, and no categorical feature that can be used to assign labels.

(2) \textit{Target-based Filter: } The preliminary selection criteria filter down Tablib by more than an order of magnitude (623M to 11M datasets). From the remaining, we take half for the semantic \fraudbench while the rest is used for \onevsrestbench. Next, we search for datasets where exactly one categorical feature value is associated with anomaly detection (e.g., ``fraud'', ``failure'', ``defect''; see the full list in Appx.~\ref{appx:anomaly_keywords}). We use samples with this feature value as anomalies and drop  other categorical features (Sec. \ref{ssec:target}).

(3) \textit{Hash and Separability Check: } We calculate a custom hash code (Appx.~\ref{appx:duplication}) for each dataset to remove duplicate datasets and also consider only such datasets where inlier samples can be separated from our anomalies. To that end, we demand that a supervised Random Forest classifier~\cite{randomForests} reaches at least $60\%$ ROC-AUC. 

(4) \textit{Manual Quality Control: } All remaining $2907$ candidate datasets are inspected manually. First, we check if the datasets are truly related to anomaly detection ($1300/2907$) and select the best-matching inlier categories for each dataset. We also manually remove all duplicate datasets that our earlier hash function missed ($790/1300$).

(5) \textit{Quality Improvements:} We add Train/Test splits using 50\% of inliers as the training and 50\% in test set with anomalies, add further metadata (Sec.~\ref{ssec:metadata}), retain part of our data as Private for an OD leaderboard (Sec.~\ref{ssec:anonym}), and create a representative set of $50$ datasets with similar  performance distribution (Sec.~\ref{ssec:represent}).

\begin{figure}[!ht]
\vspace{-0.075in}
    \centering
    \hspace{-0.075in}\subfigure[\fraudbench reflects \textbf{semantic} anomalies.]{
        \includegraphics[width=0.49\linewidth]{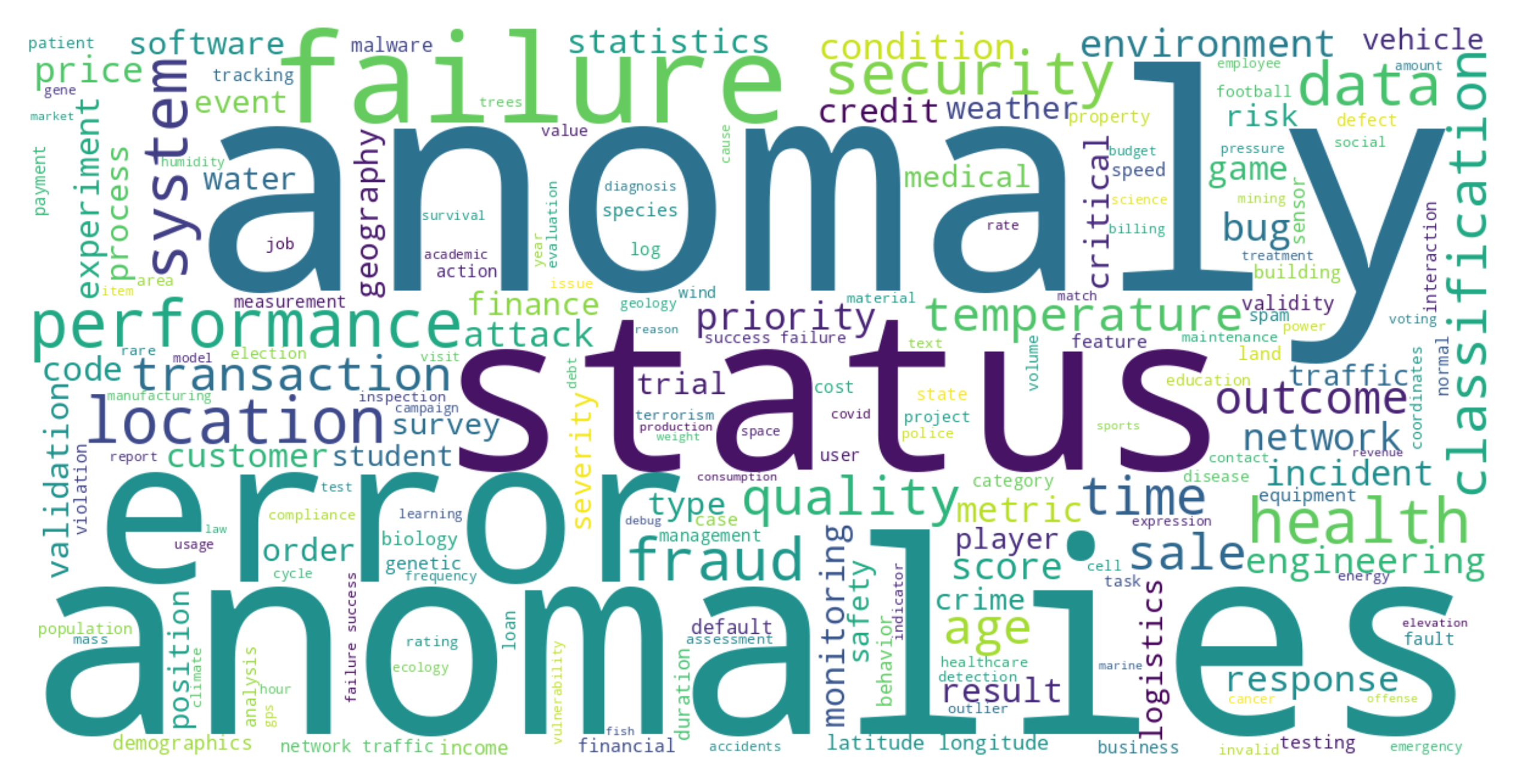}
        \label{fig:word_cloud_oddbench_b}
    }
    \hfill
    \hspace{-0.05in}
    \subfigure[\fraudbench spans \textbf{real-world domains}.]{
        \includegraphics[width=0.49\linewidth]{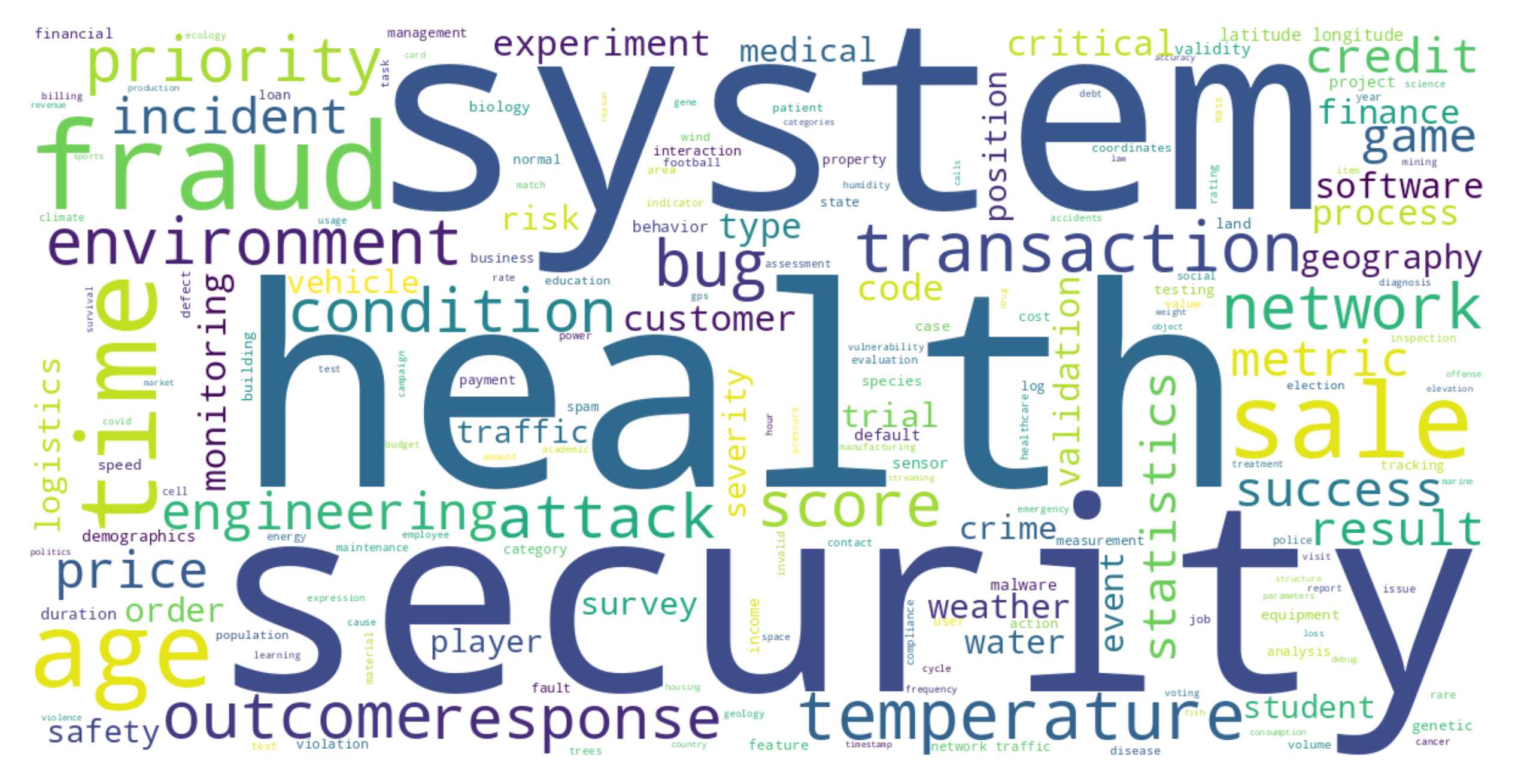}
        \label{fig:word_cloud_oddbench_a}
    }
    \vspace{-0.175in}
    \caption{Word clouds of metadata keywords across \fraudbench datasets showing (a) the semantic anomalies  and (b) a broad spectrum of domains covered. Full version is shown in (a), while overly frequent keywords including ``anomal'', ``status'', ``outlier'', ``classification'', ``data'', ``performance'', ``failure'', ``quality'', ``location'', and ``error'' are dropped in (b).}
    \label{fig:word_cloud_oddbench}
    \vspace{-0.2in}
\end{figure}

  \vspace{-0.125in}
\subsubsection{\textbf{Target Selection}}
\label{ssec:target}
\fraudbench comprises only those datasets with a categorical feature that explicitly identifies anomalous samples. Appx.~\ref{appx:anomaly_keywords} details a list of keywords used to select anomalous feature values (e.g., ``attack'', ``failure'', ``error''). Each candidate dataset undergoes manual verification to confirm semantic relevance to the anomaly detection domain and to select the most suitable subset of inlier feature values if multiple are present.



  \vspace{-0.05in}
\subsubsection{\textbf{Metadata}}\label{ssec:metadata}
To facilitate further research on anomaly detection, each dataset incorporates extensive metadata, comprising of feature names and original source links. We leverage a Large Language Model (LLM) (Appx.~\ref{appx:metadata}) to derive descriptive, human-readable titles and relevant keywords. Figure~\ref{fig:word_cloud_oddbench} illustrates that \fraudbench spans various types of semantic anomalies (a), from a variety of real-world domains (b). Furthermore, domain- and dataset-specific tags enable targeted investigations into specialized scenarios, such as high-dimensional or low-sample anomaly detection (Appx.~\ref{appx:fraudbench_distribution}, Fig.~\ref{fig:Tags}).

  \vspace{-0.065in}
\subsubsection{\textbf{Anonymization Efforts}}\label{ssec:anonym}
\label{ssec:deanon}


We designate $100$ private datasets for a leaderboard to maintain benchmark integrity. Our anonymization pipeline strips all metadata, leaving only raw training and test samples. We further obfuscate each dataset by randomly shuffling both feature and sample order.
These, combined with our manual dataset curation, ensure that reverse-engineering the source datasets is prohibitively time-consuming, therefore securing private test labels against de-anonymization.

  \vspace{-0.065in}
\subsubsection{\textbf{Representative Set}}\label{ssec:represent}

To curb the computation cost incurred by evaluating a possibly slow detector on hundreds of public datasets, we provide a representative set~\cite{representativesets} (denoted $\mathcal{S}$) with similar performance distribution across all OD methods (denoted $\mathcal{M}$) studied in this paper (Sec. \ref{ssec:setup}).   We construct this subset to match the first three moments (the mean, standard deviation, and skewness) of AUROC  performances. Specifically, we approximate\footnote{We use an evolutionary search for approximation, see ~\ref{appx:representative}.} 

\vspace{-0.2in}
\begin{equation}
    \min_{\mathcal{S}; \;|\mathcal{S}|=50}
    \; \sum_{m=1}^3\sum_{M\in \mathcal{M}} \left( \; \kappa_m(M|\oddbench) - \kappa_m(M|\mathcal{S}) \;\right)^2 \;
    \label{eqn:representative}
\end{equation} 
where $\kappa_m(M|\mathcal{D})$ denotes the $m$-th moment of method M's performance distribution over datasets in $\mathcal{D}$. 
This results in a subset of datasets that 
enable efficient benchmarking and rapid development. 







\vspace{-0.05in}
\subsection{\onevsrestbench} \label{sec:OnevsRestBench}


Common benchmarking practices in outlier detection~\cite{han2022adbench,deepsvdd} repurpose tabular classification datasets by designating a singular class as inlier and all remaining classes as outliers. Although \oddbench prioritizes real-world semantic outliers, this one-vs-rest approach remains a vital proxy for e.g., capturing concept shift and OOD detection~\cite{oodsurvey}.
Consequently, we introduce \ovrbench (\underline{O}ne-\underline{v}s-\underline{R}est Bench), which integrates commonly used classification benchmarks with the remaining Tablib datasets. We follow steps similar to those in Sec.~\ref{sec:fraudbench} and highlight only the differences in building \ovrbench.




  \vspace{-0.05in}
\subsubsection{\textbf{Filtering Criteria}}
\label{ssec:datasets_class}
\ovrbench incorporates the remaining half of TabLib datasets that satisfy the preliminary filtering criteria (Sec.~\ref{ssec:filter}). To ensure a high-quality benchmark, we integrate additional datasets from established tabular classification benchmarking repositories, including OpenML-CC18~\cite{bischl2017openml}, AutoML~\cite{gijsbers2019open}, TabZilla~\cite{mcelfresh2023neural}, Talent-CLS~\cite{liu2024talent}, TabRepo~\cite{salinas2024tabrepo}, BCCO-CLS~\cite{zhang2025limix}, and TabArena~\cite{erickson2025tabarena}. As depicted in Figure~\ref{fig:pie_ovrbench}, the final composition of \onevsrestbench reflects significant diversity in data sourcing. 

\hide{
\begin{figure}[htbp]
\vspace{-0.25in}
    \centering
    \includegraphics[width=0.85 \linewidth]{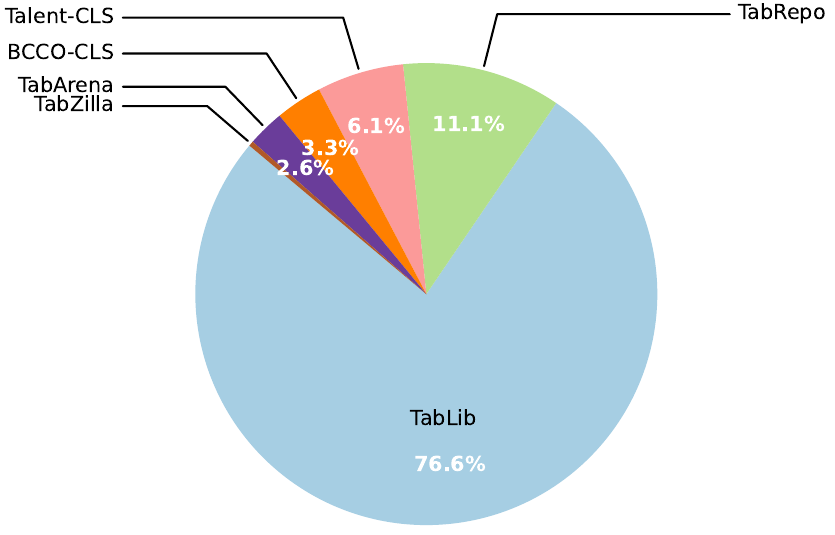}
   \vspace{-0.2in}
    \caption{Composition of \ovrbench, which exhibits significant data diversity by integrating a wide array of sources.}
    \label{fig:pie_ovrbench}
\end{figure}
}

\begin{figure}[htbp]
    \centering
    \includegraphics[width=0.75 \linewidth]{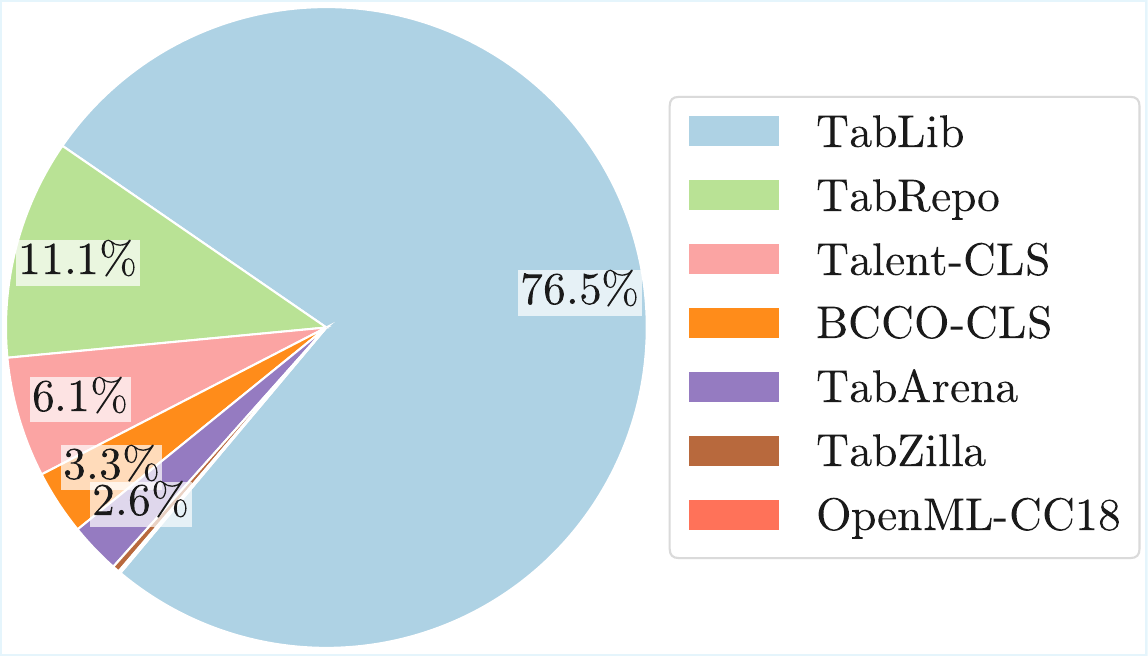}
    \caption{Composition of \ovrbench, which exhibits significant data diversity by integrating a wide array of sources. 
    }
    \label{fig:pie_ovrbench}
\end{figure}






Similar to \oddbench, we apply five steps of filtering as shown in Figure~\ref{fig:overview}.
(1) \textit{Preliminary Filter}: \ovrbench considers the second half of Tablib datasets that pass the aforementioned preliminary filtering, besides various classification benchmark sources.
(2) \textit{Target Designation and Anomaly Generation:} For Tablib tables, we select the target variable while others readily exhibit labels. We then designate inlier and anomalous classes (Sec. \ref{ssec:targetovr}).
(3) \textit{Hash and Separability Check}: We remove duplicate datasets as well as those datasets where anomalies are not separable from inliers.
(4) \textit{Manual Quality Control:} We manually remove duplicate datasets that were missed in earlier steps.
(5) \textit{Quality Improvement:} We apply the same data improvement steps as with \oddbench (Train/Test split, Metadata enrichment, Public/Private partitions and anonymization, Representative subset selection).


\begin{figure}[!t]
    \centering
    \hspace{-0.075in}\subfigure[Classification to Outlier detection]{
        \includegraphics[width=0.49\linewidth]{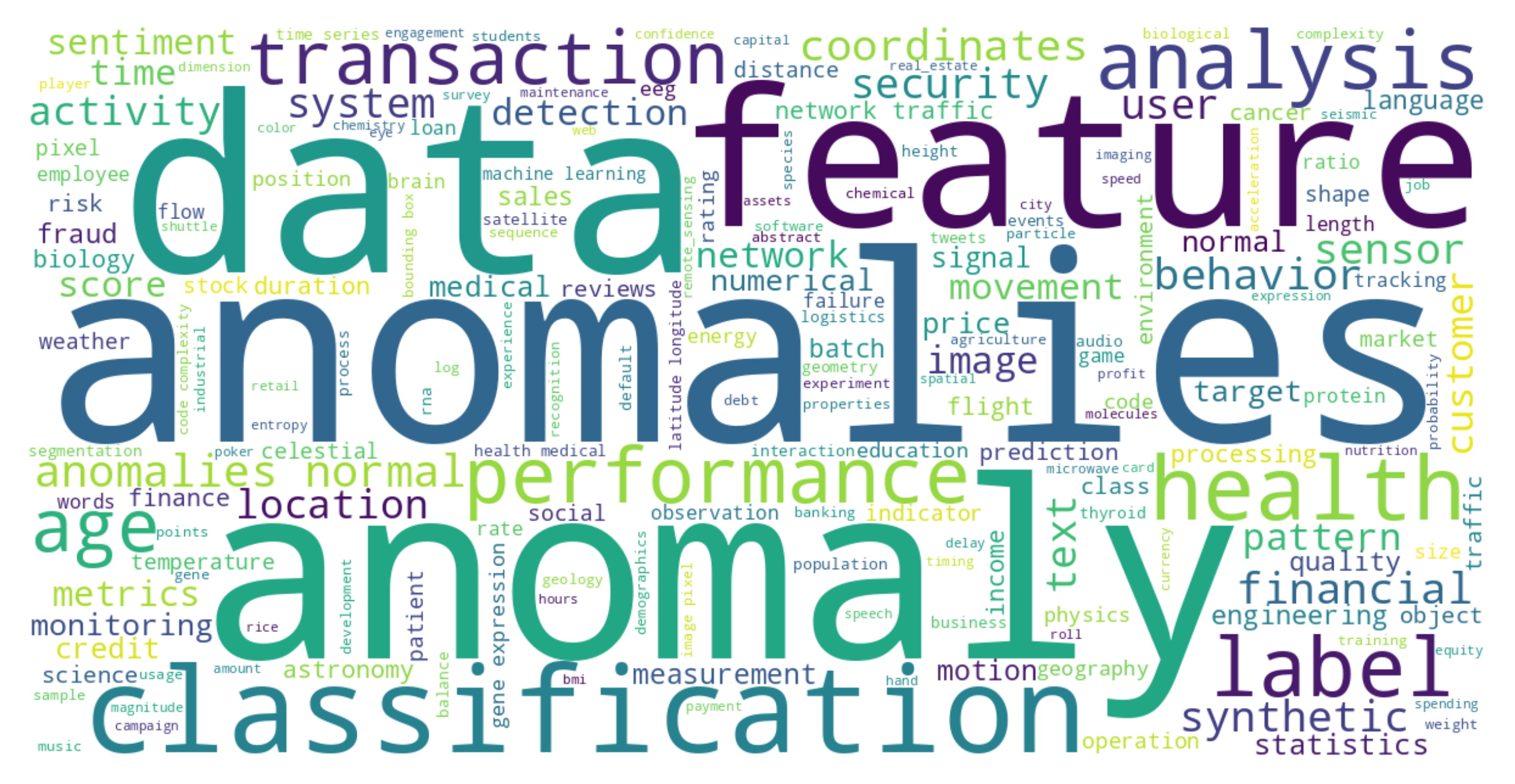}
        \label{fig:word_cloud_ovrbench_b}
    }
    \hfill
    \hspace{-0.05in}
    \subfigure[\ovrbench spans \textbf{real-world domains}.]{
        \includegraphics[width=0.49\linewidth]{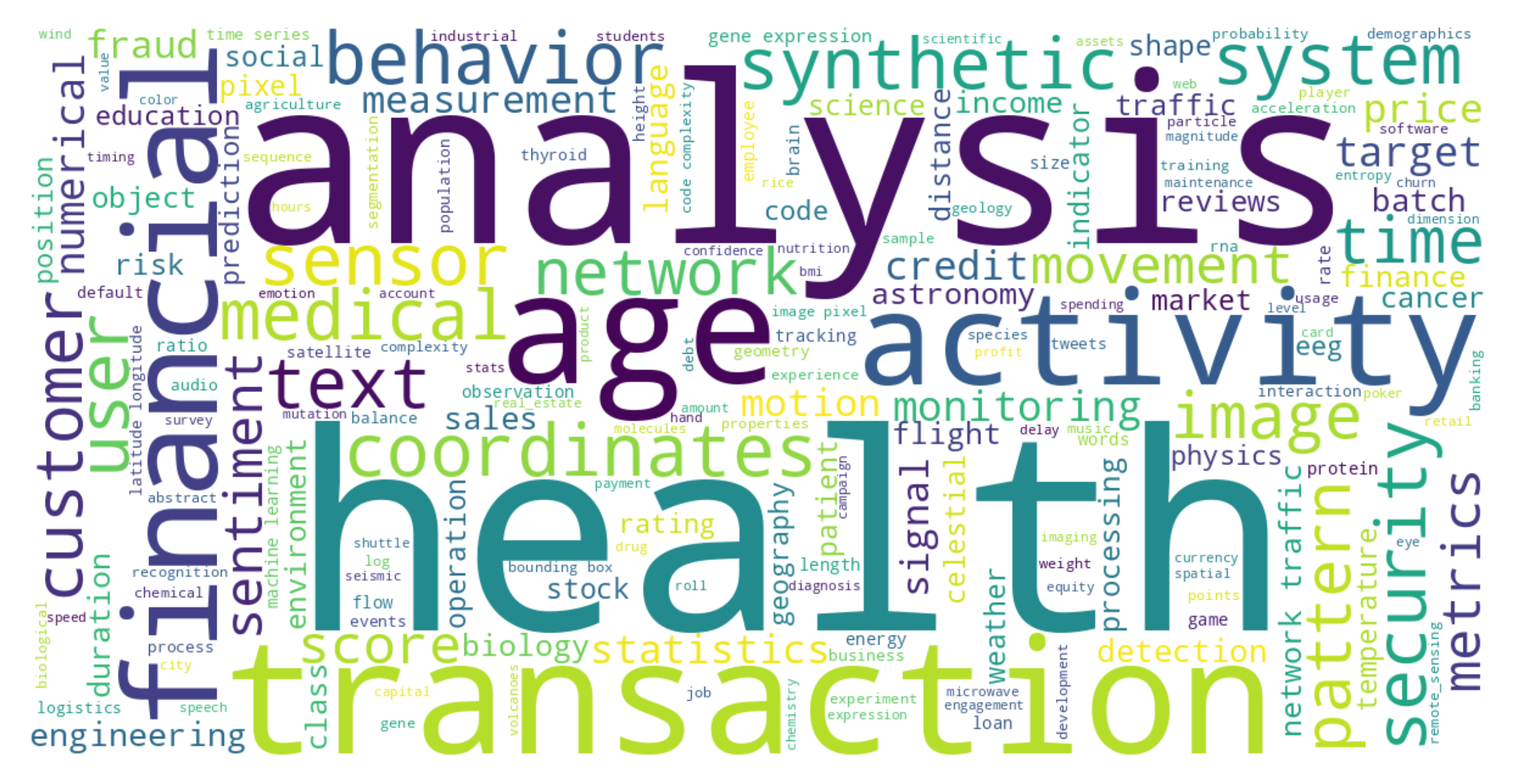}
        \label{fig:word_cloud_ovrbench_a}
    }
    \vspace{-0.175in}
    \caption{Word clouds of metadata keywords across \ovrbench datasets showing (a) it repurposes classification for OD  and (b) a broad spectrum of domains covered.  Full version is in (a), while overly frequent keywords ``anomal'',  ``classification'', ``data'',   ``label'', ``feature'',  etc. are dropped in (b).}
    \label{fig:word_cloud_ovrbench}
    \vspace{-0.15in}
\end{figure}

  \vspace{-0.05in}
\subsubsection{\textbf{Target Selection}}
\label{ssec:targetovr}
While datasets from existing classification benchmarks naturally include a label/target column,  TabLib lacks predefined target columns. To standardize these sources, we filter for datasets containing a single feature explicitly named ``label'' or ``target''. Our filtering criteria further require that the cardinality of the possible values of this feature be between 2 and 10 and that the largest category contain at least $50\%$ of the values. 
We designate this majority category as the inlier class. To ensure that anomalies are not unrealistically common, we subsample the remaining classes as anomalies at a given anomaly ratio randomly drawn from the range [0.05, 0.2]. 
The word clouds in Figure \ref{fig:word_cloud_ovrbench} (a) underscore the transformation of diverse classification tasks to \ovrbench datasets, covering distinct real-world domains (b).







 \vspace{-0.05in}
\subsubsection{\textbf{Metadata, Anonymization, Representatives}}
\label{ssec:rest}
We follow the same protocols as with \fraudbench,  described respectively in Sections \ref{ssec:metadata}, \ref{ssec:deanon} and \ref{ssec:represent}.

\subsection{\synbench}
\label{sec:synbench}

We simulate 800 datasets from three distinct data priors with five types of outliers, spanning feature dimensionalities in $[2, 100]$. We briefly describe each prior, with full details in the Appx. \ref{appx:synbench}. 

 \vspace{-0.05in}
\subsubsection{\textbf{Gaussian Mixture}}
We adopt Gaussian mixture models (GMMs) as a synthetic prior due to their ability to capture multimodal structure arising from heterogeneous subpopulations while remaining analytically tractable. As universal approximators of continuous densities \cite{carreira2002mode}, GMMs offer a flexible yet interpretable family that enables controlled variation in modality, correlation, and distributional complexity, facilitating systematic evaluation under diverse data regimes.

We simulate inliers from multivariate GMMs with varying numbers of components and dimensions, labeling samples within the $90^{\text{th}}$ percentile of the GMM likelihood as inliers \cite{shen2025fomod}. We generate \emph{contextual subspace outliers} by randomly selecting a mixture component and inflating variances along a subset of dimensions; samples outside the $90^{\text{th}}$ percentile are labeled as outliers. Outlier severity is governed by the subspace size and variance inflation scale, allowing fine-grained control over detection difficulty.

 \vspace{-0.05in}
\subsubsection{\textbf{Structural Causal Models (SCM)}} 
We adopt Structural Causal Models (SCMs) as a synthetic prior for their expressive power under mild assumptions \cite{bongers2021foundations}. By varying graph structures, structural functions, and noise, SCMs generate rich distributions with multimodality, nonlinearity, non-Gaussian marginals, and heteroscedasticity—properties common in real-world tabular data \cite{peters2017elements,scholkopf2022causality}. They also allow efficient sampling via a single topological pass through an MLP, with cost linear in the number of edges. We instantiate an SCM as a directed acyclic graph $G=(V,E)$, where each node corresponds to a variable and structural equations take the form $X_j = f_j(X_{Pa(X_j;G)}, \epsilon_j)$. We materialize $G$ using an MLP with randomly dropped edges, sample weights, inputs, and exogenous noise from standard Normal distributions, and define each $f_j$ as a weighted sum followed by a random activation. Propagating noise through the graph and reading out a selected subset of nodes $\mathcal{X} \subset V$ and target $Y$ yields inlier samples.

We generate two outlier types: {measurement} and {structural} outliers. \textit{Measurement outliers} are created by inflating the variance of the exogenous noise at a randomly selected variable, allowing the perturbation to propagate downstream while preserving the causal structure. \textit{Structural outliers} instead modify the causal graph by breaking or reversing edges, 
thereby altering the underlying data-generating mechanisms. While measurement outliers typically exhibit extreme values, structural outliers are characterized by violations of expected causal dependencies and may not be extreme in magnitude. Outlier severity is controlled via the noise inflation scale and the fraction of intervened edges.

 \vspace{-0.05in}
\subsubsection{\textbf{Copulas}} We use copula models as a synthetic prior to independently control feature marginals and dependence, motivated by evidence that real-world marginals often exhibit skewed or power-law behavior \cite{clauset2009power}. By Sklar’s theorem \cite{sklar1959fonctions}, any continuous multivariate distribution decomposes into marginal CDFs and a copula, enabling separate specification of marginals and dependencies. This flexibility allows us to model diverse tabular distributions with non-Gaussian marginals, nonlinear or asymmetric dependence, and tail dependence, beyond covariance-based models \cite{nelsen2006introduction,houssou2022generation}.

We draw each feature’s marginal independently from a pool of parametric distributions (Gaussian, Beta, Exponential, Student’s $t$, Power-law, and Log-logistic) and choose either a Gaussian or bivariate vine copula for the dependence structure. Inliers are generated by sampling $\mathbf{u} \sim C$ on $[0,1]^d$ and transforming each coordinate via $x_j = F_j^{-1}(u_j)$, yielding $\mathbf{x}=(x_1,\ldots,x_d)$. We vary marginals, parameters, and copula families to obtain a wide range of datasets. Leveraging the copula decomposition, we construct two outlier types. \emph{Probabilistic outliers} are created by pushing a subset of copula coordinates toward the boundaries (near $0$ or $1$), producing multivariate extremes that preserve approximate dependence after inverse marginal mapping. \emph{Dependence outliers} disrupt the joint structure by inverting ($u_j \leftarrow 1-u_j$) or permuting selected coordinates, violating dependencies without necessarily inducing extreme values. Outlier severity is controlled by the fraction of perturbed dimensions and the magnitude of the copula-space perturbations. 

\hide{
\subsection{Validation \lmn{change subtitle, has specific meaning in ML, weird here.}}
Fig.~\ref{fig:summary} illustrates the extensive statistical diversity of \bench, which encompasses a much wider range than the existing ADBench~\cite{han2022adbench} collection (Fig.~\ref{fig:crownjewel}) and is evenly distributed \lmn{evenly distributed means? uniform? weird/vague word choice.} (Appx.~\ref{appx:fraudbench_distribution}, Fig.~\ref{fig:nd_odd}). Furthermore, the word clouds in Fig.~\ref{fig:word_cloud_oddbench} and Fig.~\ref{fig:word_cloud_ovrbench} confirm that our proposed real-world datasets captures many different semantic origins. 
While many datasets exhibit high separability, Appx.~\ref{appx:fraudbench_distribution}, Fig.~\ref{fig:performance_odd} highlights that anomaly detection algorithms still underperform on a large fraction of the data. This gap underscores a substantial opportunity for algorithmic advancement and validates the need for better outlier detection frameworks like our \bench.
}



\vspace{-0.05in}
\section{Evaluation}
\label{sec:experiments}

We design experiments to investigate the following empirical questions, based on which we provide practical guidelines.
\makeatletter
\setlength{\leftmargini}{0.2in}
\setlength{\leftmarginii}{0.3in}
\setlength{\leftmarginiii}{0.2in}
\setlength{\leftmarginiv}{0.3in}
\makeatother
\begin{itemize}
\item \textbf{Q1) Performance:} How do prominent OD methods, from classical to deep to foundation models, compare on \bench? 

\item \textbf{Q2) Running time:} How does running time trade off against performance for different methods?

\item \textbf{Q3) Hyperparameter sensitivity:} How sensitive are different methods to their HP configuration choices? 
\end{itemize}
We conduct an extensive evaluation on all \textbf{2,446} \bench datasets using \textbf{14} popular OD methods. The methods are chosen from top-performing approaches identified in prior studies and evaluated across a range of hyperparameter (HP) configurations. 
We govern experiment design and HP configurations systematically to ensure fair evaluation and to set a standard for the future.

\vspace{-0.1in}
\subsection{Experiment Setup}
\label{ssec:setup}

\noindent
\textbf{Benchmarked Methods.}
For competitive evaluation, we take the union of the top--2 OD methods reported in established survey and benchmarking studies~\citep{emmott2013systematic,emmott2015meta,campos2016evaluation,goldstein2016comparative,domingues2018comparative,steinbuss2021benchmarking,han2022adbench,LivernocheJHR24}, covering both \textbf{classical} and \textbf{deep} models. We also include recent \textbf{foundation} models, namely \fomo~\citep{shen2025fomod} and  \outform  \cite{ding2026outformer} which are specialized in zero-shot tabular OD, and include TabPFN-OD; an adaptation of TabPFN \cite{hollmann2023tabpfn} for supervised prediction to OD, by designating each feature as a prediction target and averaging  errors across features as the outlier score. See Appx. \ref{appx:baselines} and Table~\ref{tab:baselines}.

\noindent
\textbf{HP Configurations.} Most classical  and deep models are sensitive to their HP choices as discussed in Sec. \ref{ssec:history}. For fair evaluation, we consider different HP configurations and report average performance and standard deviation to capture performance variability. For shallow models, we consider the full HP grid. For deep models, due to time and memory constraints, we randomly sample five HP settings from the predefined search space. Appx. Table~\ref{tab:hps} gives detailed HP configurations for classical and deep methods.
In contrast, foundation models require no model fitting, thereby no HP-tuning.


\noindent
\textbf{Metrics.} We extensively evaluate the baselines on \bench w.r.t. detection performance, runtime, and HP sensitivity.
For detection, we report results
using both AUROC and AUPRC. Instead of aggregating raw performances across hundreds of datasets with varying difficulty, we report five metrics that better capture relative performance across datasets: Average Rank, ELO~\cite{elo1967uscf}, Win Rate, rescaled AUROC or AUPRC (rAUC), and Champion Delta ($C_{\Delta}$)~\cite{zhang2025mitra}. In addition, we conduct pairwise statistical comparisons between methods and report $p$-values based on permutation tests. Appx. \ref{assec:metrics} gives implementation details. For total running time, we report inference time (only) for foundation models, and model fitting plus inference time for classical and deep models.
For HP sensitivity of a baseline, we measure the inter-quartile range of 25\%-75\% across HP configurations and report its distribution across datasets.

\vspace{-0.1in}
\subsection{Results \& Guidelines}
\label{ssec:results}
\subsubsection{\textbf{Main Results}}

\begin{figure}[!htbp]
    \centering
    \includegraphics[width=0.90\linewidth]{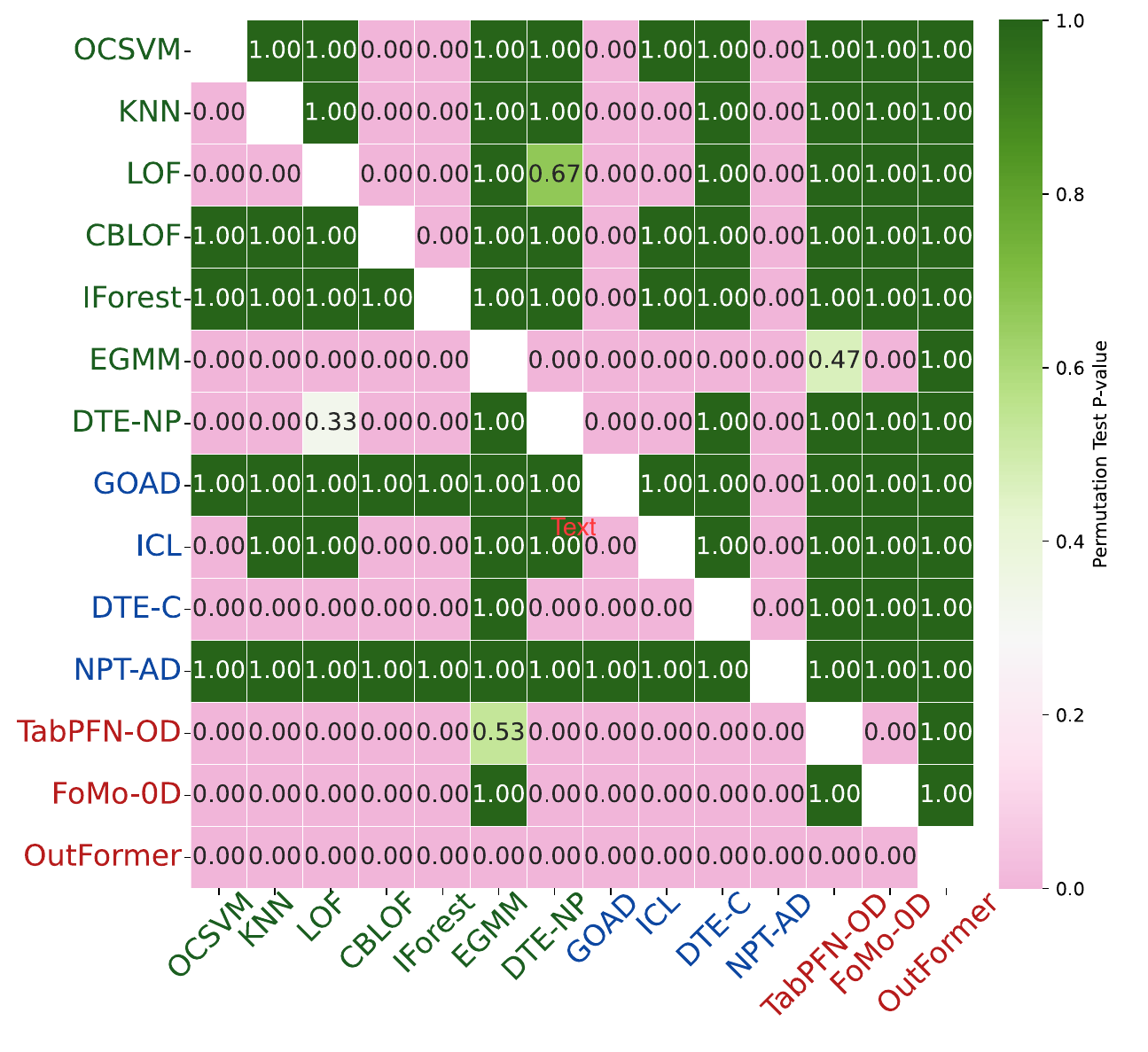}
    \vspace{-0.18in}
    \caption{Paired permutation tests on AUPRC across all datasets show the competitiveness of foundation models, with \outform significantly outperforming all baselines.}
    \label{fig:perm_summary}
\end{figure}

\textbf{Q1) Performance:~} Table \ref{tab:overall_auroc} provides relative AUPRC performance comparison of all baselines across all 2,446 \bench datasets combined. Figure \ref{fig:perm_summary} supplements these metrics with paired statistical tests between all 14 baselines. 
(See Appx. \ref{assec:indivpr} for results on individual benchmarks, and  Appx. \ref{assec:roc} for all corresponding results w.r.t. AUROC.)

\begin{table}[!t]
\centering
\caption{Overall AUPRC performance comparison of OD methods on all 2,446 datasets. Comparisons on individual benchmarks, \oddbench, \ovrbench, and \synbench, are respectively in Appx. \ref{assec:indivpr} Tables \ref{tab:oddbench_results_auprc},\ref{tab:ovrbench_results_auprc} and \ref{tab:synbench_results_auprc}. \textcolor{darkgreen}{green}, \textcolor{darkblue}{blue}, \textcolor{darkred}{red} depict classical, deep, foundation model categories. {Best} result within category is bolded, with \underline{overall best} {underlined}.}
\vspace{-0.1in}
\label{tab:overall_results}
\setlength{\tabcolsep}{1pt}
\resizebox{\linewidth}{!}{
\begin{tabular}{clccccc}
\toprule
& \textbf{Model} & \textbf{Avg. Rank ($\downarrow$)} & \textbf{ELO ($\uparrow$)} &
\textbf{Winrate ($\uparrow$)} & \textbf{rAUC ($\uparrow$)} &
\textbf{$C_{\Delta}$ ($\downarrow$)} \\
\midrule

\multirow{7}{*}{\rotatebox{90}{\textbf{Classical}}}
& \textbf{\textcolor{darkgreen}{OCSVM}} & \pmv{8.15}{3.2} & 370 & 0.44 & \pmv{0.724}{0.21} & 0.52 \\
& \textbf{\textcolor{darkgreen}{KNN}} & \pmv{6.68}{3.0} & 1036 & 0.55 & \pmv{0.766}{0.23} & 0.45 \\
& \textbf{\textcolor{darkgreen}{LOF}} & \pmv{6.20}{3.5} & 1265 & 0.58 & \pmv{0.773}{0.23} & 0.44 \\
& \textbf{\textcolor{darkgreen}{CBLOF}} & \pmv{8.42}{3.4} & 893 & 0.42 & \pmv{0.676}{0.28} & 0.49 \\
& \textbf{\textcolor{darkgreen}{IForest}} & \pmv{9.56}{3.9} & 633 & 0.34 & \pmv{0.628}{0.29} & 0.55 \\
& \textbf{\textcolor{darkgreen}{EGMM}} & \textbf{\pmv{4.88}{3.6}} & \textbf{1535} & \textbf{0.68} & \textbf{\pmv{0.833}{0.23}} & \textbf{0.34} \\
& \textbf{\textcolor{darkgreen}{DTE-NP}} & \pmv{6.12}{2.9} & 1026 & 0.59 & \pmv{0.783}{0.22} & 0.44 \\

\midrule
\multirow{4}{*}{\rotatebox{90}{\textbf{Deep}}}
& \textbf{\textcolor{darkblue}{GOAD}} & \pmv{11.17}{3.1} & 674 & 0.21 & \pmv{0.557}{0.28} & 0.57 \\
& \textbf{\textcolor{darkblue}{ICL}} & \pmv{7.89}{3.3} & \textbf{1180} & 0.44 & \pmv{0.721}{0.23} & 0.50 \\
& \textbf{\textcolor{darkblue}{DTE-C}} & \textbf{\pmv{6.78}{3.5}} & 1090 & \textbf{0.52} & \textbf{\pmv{0.781}{0.23}} & \textbf{0.45} \\
& \textbf{\textcolor{darkblue}{NPT-AD}} & \pmv{11.44}{3.7} & 55 & 0.19 & \pmv{0.517}{0.29} & 0.58 \\

\midrule
\multirow{3}{*}{\rotatebox{90}{\textbf{FM}}}
& \textbf{\textcolor{darkred}{TabPFN-OD}} & \pmv{5.22}{4.0} & \underline{\textbf{1604}} & 0.66 & \pmv{0.822}{0.24} & \underline{\textbf{0.26}} \\
& \textbf{\textcolor{darkred}{FoMo-0D}} & \pmv{6.20}{4.0} & 1283 & 0.59 & \pmv{0.800}{0.23} & 0.39
 \\
& \textbf{\textcolor{darkred}{OutFormer}} & \underline{\textbf{\pmv{4.84}{3.8}}} & 1355 & \underline{\textbf{0.69}} & \underline{\textbf{\pmv{0.854}{0.22}}} & 0.29 \\
\bottomrule
\end{tabular}
}
\label{tab:overall_auroc}
\vspace{-0.1in}
\end{table}

Together, Table \ref{tab:overall_auroc} and Figure \ref{fig:perm_summary} show that tabular foundation models (FMs) significantly outperform all classical and deep OD methods, except EGMM \cite{EGMM}. Ensemble-based EGMM is competitive with \tabod but much slower as shown next, and is significantly outperformed by \outform. Deep OD models underperform relative to many classical methods, largely due to their hyperparameter (HP) sensitivity, as detailed below. 

\begin{figure}[!ht]
    \centering
    \begin{minipage}[t]{0.49\linewidth}
        \centering
        \includegraphics[width=\linewidth]{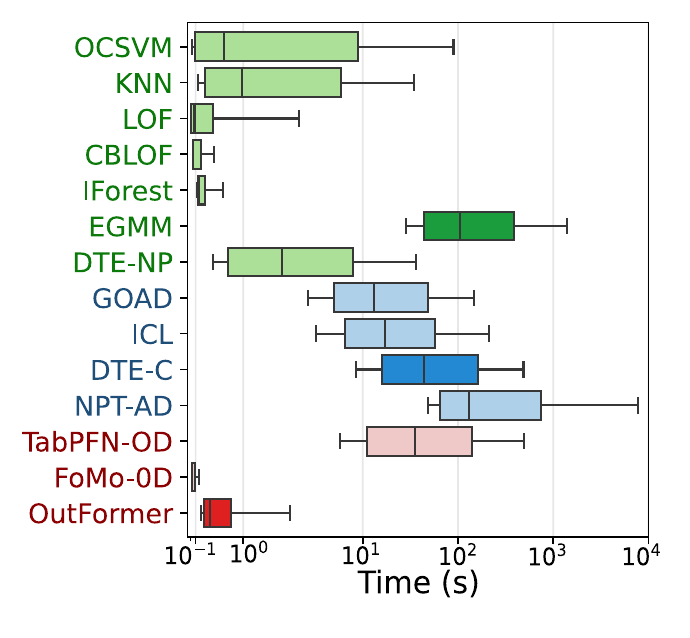}
    \end{minipage}
    \begin{minipage}[t]{0.5\linewidth}
        \centering
        \includegraphics[width=\linewidth]{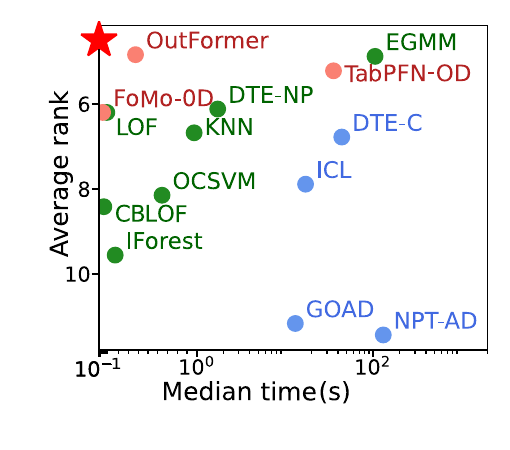}
    \end{minipage}
    \vspace{-0.24in}
    \caption{(left) Running time comparison of OD methods across all datasets. (right) Time-vs-Performance trade-off; FM-based \fomo and \outform occupy the Pareto front.}
    \label{fig:total_time}

\end{figure}

\textbf{Q2) Running time:}
Figure \ref{fig:total_time} (left) presents the total running time distribution over all datasets for all methods. In general, classical methods are faster than deep ones, with the exception of EGMM. Feed-forward-only FMs incur the least cost, except TabPFN-OD which iterates over features for prediction.
Ensemble-based EGMM that stands out performance-wise is relatively slow, underscoring the typical trade-off in Figure \ref{fig:total_time} (right). Remarkably, FMs like \fomo and \outform  take the lead  in \textit{both} performance and running time---offering the best of both worlds.


\begin{figure}[!ht]
\vspace{-0.15in}
    \centering
\includegraphics[width=0.8\linewidth]{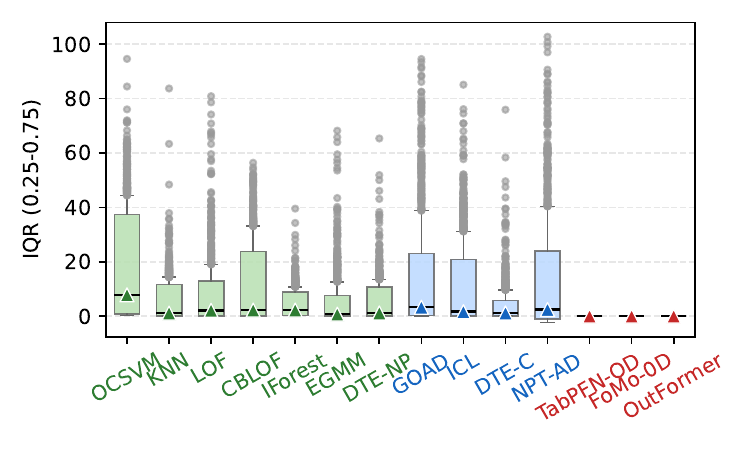}
    \vspace{-0.225in}
    \caption{Performance variation across HP configurations (as measured by inter-quartile 25\%-75\% AUPRC; the higher, the more HP-sensitive) across all datasets (boxplot) per baseline. }
    \label{fig:hp_sensitivity_overall}
    \vspace{-0.175in}
\end{figure}

\textbf{Q3) Hyperparameter (HP)  sensitivity:}
Figure \ref{fig:hp_sensitivity_overall} shows that performance varies substantially with HP choices across datasets, especially for deep models with many HPs, compared to classical methods with only a few. Ensemble approaches such as iForest \cite{Iforest} and EGMM \cite{EGMM} exhibit greater HP robustness, while FMs are especially notable for eliminating the need for tuning altogether.

\subsubsection{\textbf{Practical Guidelines}}
\label{ssec:guidelines}
Our extensive evaluations firmly position FMs as the dominant approach to OD, excelling in \textit{both} performance and runtime---a rare occurrence. 
Classical methods typically outperform deep models and offer greater robustness. Among those, EGMM, DTE-NP, and LOF define the performance-runtime Pareto front in Figure \ref{fig:total_time} (right), where  LOF is lightweight and well suited for CPU-only, low-resource settings. FM-based \fomo and \outform, however, dominate these and occupy the true Pareto front, with model sizes of roughly 5M and 45M parameters; far smaller than the billion-scale language models, and well within the limits of mainstream GPUs.
Ultimately,  FMs  stand at the forefront for practical OD owing to their strong performance, low-latency inference, and truly plug-and-play nature.




\section{Conclusion}
\label{sec:conclusion}

We introduce \bench, a large-scale benchmark suite for tabular outlier detection (OD), consisting of 2,446 datasets from diverse real-world domains and synthetic inlier/outlier priors, 
with 
standardized train/test splits and rich semantic metadata. Besides public datasets, we reserve private subsets and their test labels, supporting an  online OD leaderboard. We present extensive experiments across all benchmarks, analyzing performance, runtime, and hyperparameter sensitivity of a select set of OD methods.
With its scale and diversity, \bench provides a robust testbed for evaluating OD with enhanced statistical power. 
Beyond evaluation, our benchmarks, along with data curation and generation strategies, 
open new research avenues in meta-learning and foundation models for OD, which emerge as the leading-edge frontier.



\hide{
We introduce \bench, a large-scale benchmark suite for tabular OD, comprising 2,446 datasets including real-world datasets from diverse domains as well as synthetic datasets from diverse inlier/outlier data priors. 
Thanks to its scale and diversity, \bench offers a comprehensive testbed for
evaluating OD methods with greater statistical power.
Datasets have 
 standardized train/test splits  as well as rich semantic metadata. We partition benchmarks as 
public/private, reserve test labels of all private datasets and maintain an online leaderboard. 
We conduct extensive
experiments across all benchmarks and present analyses on performance, running time and hyperparameter sensitivity of a representative list of OD methods.
Our benchmarks, along with data curation and generation strategies, present research opportunities beyond evaluation; including meta learning and foundation models.
}









\bibliographystyle{ACM-Reference-Format}
\bibliography{00refs}

@inproceedings{rayana2016sequential,
  title={Sequential ensemble learning for outlier detection: A bias-variance perspective},
  author={Rayana, Shebuti and Zhong, Wen and Akoglu, Leman},
  booktitle={2016 IEEE 16th international conference on data mining (ICDM)},
  pages={1167--1172},
  year={2016},
  organization={IEEE}
}

@article{liu2024talent,
  title={TALENT: A tabular analytics and learning toolbox},
  author={Liu, Si-Yang and Cai, Hao-Run and Zhou, Qi-Le and Ye, Han-Jia},
  journal={arXiv preprint arXiv:2407.04057},
  year={2024}
}

@article{zhang2025limix,
  title={Limix: Unleashing structured-data modeling capability for generalist intelligence},
  author={Zhang, Xingxuan and Ren, Gang and Yu, Han and Yuan, Hao and Wang, Hui and Li, Jiansheng and Wu, Jiayun and Mo, Lang and Mao, Li and Hao, Mingchao and others},
  journal={arXiv preprint arXiv:2509.03505},
  year={2025}
}

@article{gijsbers2019open,
  title={An open source AutoML benchmark},
  author={Gijsbers, Pieter and LeDell, Erin and Thomas, Janek and Poirier, S{\'e}bastien and Bischl, Bernd and Vanschoren, Joaquin},
  journal={arXiv preprint arXiv:1907.00909},
  year={2019}
}

@article{erickson2025tabarena,
  title={Tabarena: A living benchmark for machine learning on tabular data},
  author={Erickson, Nick and Purucker, Lennart and Tschalzev, Andrej and Holzm{\"u}ller, David and Desai, Prateek Mutalik and Salinas, David and Hutter, Frank},
  journal={arXiv preprint arXiv:2506.16791},
  year={2025}
}

@article{mcelfresh2023neural,
  title={When do neural nets outperform boosted trees on tabular data?},
  author={McElfresh, Duncan and Khandagale, Sujay and Valverde, Jonathan and Prasad C, Vishak and Ramakrishnan, Ganesh and Goldblum, Micah and White, Colin},
  journal={Advances in Neural Information Processing Systems},
  volume={36},
  pages={76336--76369},
  year={2023}
}

@inproceedings{salinas2024tabrepo,
  title={TabRepo: A Large Scale Repository of Tabular Model Evaluations and its AutoML Applications},
  author={Salinas, David and Erickson, Nick},
  booktitle={International Conference on Automated Machine Learning},
  pages={19--1},
  year={2024},
  organization={PMLR}
}

@article{bischl2017openml,
  title={Openml benchmarking suites},
  author={Bischl, Bernd and Casalicchio, Giuseppe and Feurer, Matthias and Gijsbers, Pieter and Hutter, Frank and Lang, Michel and Mantovani, Rafael G and van Rijn, Jan N and Vanschoren, Joaquin},
  journal={arXiv preprint arXiv:1708.03731},
  year={2017}
}

@article{cblof, author = {He, Zengyou and Xu, Xiaofei and Deng, Shengchun}, title = {Discovering Cluster-Based Local Outliers}, year = {2003}, issue_date = {01 June 2003}, publisher = {Elsevier Science Inc.}, address = {USA}, volume = {24}, number = {9–10}, issn = {0167-8655}, journal = {Pattern Recogn. Lett.}, month = {jun}, pages = {1641–1650}, numpages = {10} }

@article{
shen2025fomod,
title={FoMo-0D: A Foundation Model for Zero-shot Tabular Outlier Detection},
author={Yuchen Shen and Haomin Wen and Leman Akoglu},
journal={Transactions on Machine Learning Research},
issn={2835-8856},
year={2025},
url={https://openreview.net/forum?id=XCQzwpR9jE},
note={}
}

@article{ho2020denoising,
  title={Denoising diffusion probabilistic models},
  author={Ho, Jonathan and Jain, Ajay and Abbeel, Pieter},
  journal={Advances in Neural Information Processing Systems},
  volume={33},
  pages={6840--6851},
  year={2020}
}

@inproceedings{Iforest,
  author={Liu, Fei Tony and Ting, Kai Ming and Zhou, Zhi-Hua},
  booktitle={2008 Eighth IEEE International Conference on Data Mining}, 
  title={Isolation Forest}, 
  year={2008},
  volume={},
  number={},
  pages={413-422},
}

@InProceedings{deepsvdd,
  title = 	 {Deep One-Class Classification},
  author =       {Ruff, Lukas and Vandermeulen, Robert and Goernitz, Nico and Deecke, Lucas and Siddiqui, Shoaib Ahmed and Binder, Alexander and M{\"u}ller, Emmanuel and Kloft, Marius},
  booktitle = 	 {International Conference on Machine Learning},
  pages = 	 {4393--4402},
  year = 	 {2018},
  }

@inproceedings{emmott2013systematic,
  title={Systematic construction of anomaly detection benchmarks from real data},
  author={Emmott, Andrew F and Das, Shubhomoy and Dietterich, Thomas and Fern, Alan and Wong, Weng-Keen},
  booktitle={Proceedings of the ACM SIGKDD workshop on outlier detection and description},
  pages={16--21},
  year={2013}
}

@inproceedings{goad,
  author = {Bergman, Lion and Hoshen, Yedid},
  booktitle = {International Conference on Learning Representations},
  description = {Classification-Based Anomaly Detection for General Data | OpenReview},
  title = {Classification-Based Anomaly Detection for General Data},
  year = 2020
}

@article{chalapathy2019deep,
  title={Deep learning for anomaly detection: A survey},
  author={Chalapathy, Raghavendra and Chawla, Sanjay},
  journal={arXiv preprint arXiv:1901.03407},
  year={2019}
}

@article{pang2021deep,
  title={Deep learning for anomaly detection: A review},
  author={Pang, Guansong and Shen, Chunhua and Cao, Longbing and Hengel, Anton Van Den},
  journal={ACM computing surveys (CSUR)},
  volume={54},
  number={2},
  pages={1--38},
  year={2021},
  publisher={ACM New York, NY, USA},
  url={https://github.com/mala-lab/ADBenchmarks-anomaly-detection-datasets}
}

@book{books/sp/Aggarwal2013,
  author = {Aggarwal, Charu C.},
  publisher = {Springer},
  title = {Outlier Analysis},
  year = 2013
}

@misc{Rayana2016ODDS,
  author       = {Rayana, Shebuti and Akoglu, Leman},
  title        = {ODDS Library: Outlier Detection DataSets},
  year         = {2016},
  howpublished = {\url{https://shebuti.com/outlier-detection-datasets-odds/}},
  institution  = {Stony Brook University, Department of Computer Science}
}

@inproceedings{Vanschoren2018MetaLearningAS,
  title={Meta-Learning: A Survey},
  author={Joaquin Vanschoren},
  booktitle={Automated Machine Learning},
  year={2018},
  url={https://api.semanticscholar.org/CorpusID:52938664}
}

@inproceedings{LivernocheJHR24,
  author = {Livernoche, Victor and Jain, Vineet and Hezaveh, Yashar and Ravanbakhsh, Siamak},
   booktitle = {ICLR},
  title = {On Diffusion Modeling for Anomaly Detection.},
  year = 2024
}

@article{ding2022hyperparameter,
  title={Hyperparameter sensitivity in deep outlier detection: Analysis and a scalable hyper-ensemble solution},
  author={Ding, Xueying and Zhao, Lingxiao and Akoglu, Leman},
  journal={Advances in Neural Information Processing Systems},
  volume={35},
  pages={9603--9616},
  year={2022}
}

@article{ding2023fast,
  title={Fast Unsupervised Deep Outlier Model Selection with Hypernetworks},
  author={Ding, Xueying and Zhao, Yue and Akoglu, Leman},
  journal={ACM SIGKDD},
  year={2024}
}

@article{steinbuss2021benchmarking,
  title={Benchmarking unsupervised outlier detection with realistic synthetic data},
  author={Steinbuss, Georg and B{\"o}hm, Klemens},
  journal={ACM Transactions on Knowledge Discovery from Data (TKDD)},
  volume={15},
  number={4},
  pages={1--20},
  year={2021},
  publisher={ACM New York, NY, USA}
}

@inproceedings{zhao2022toward,
  title={Toward unsupervised outlier model selection},
  author={Zhao, Yue and Zhang, Sean and Akoglu, Leman},
  booktitle={2022 IEEE International Conference on Data Mining (ICDM)},
  pages={773--782},
  year={2022},
  organization={IEEE}
}

@article{zhao2021automatic,
  title={Automatic unsupervised outlier model selection},
  author={Zhao, Yue and Rossi, Ryan and Akoglu, Leman},
  journal={Advances in Neural Information Processing Systems},
  volume={34},
  pages={4489--4502},
  year={2021}
}

@inproceedings{sohl2015deep,
  title={Deep unsupervised learning using nonequilibrium thermodynamics},
  author={Sohl-Dickstein, Jascha and Weiss, Eric and Maheswaranathan, Niru and Ganguli, Surya},
  booktitle={International conference on machine learning},
  pages={2256--2265},
  year={2015},
  organization={pmlr}
}

@article{aggarwal2015theoretical,
  title={Theoretical foundations and algorithms for outlier ensembles},
  author={Aggarwal, Charu C. and Sathe, Saket},
  journal={Acm sigkdd explorations newsletter},
  volume={17},
  number={1},
  pages={24--47},
  year={2015},
  publisher={ACM New York, NY, USA}
}

@article{campos2016evaluation,
  title={On the evaluation of unsupervised outlier detection: measures, datasets, and an empirical study},
  author={Campos, Guilherme O and Zimek, Arthur and Sander, J{\"o}rg and Campello, Ricardo JGB and Micenkov{\'a}, Barbora and Schubert, Erich and Assent, Ira and Houle, Michael E},
  journal={Data mining and knowledge discovery},
  volume={30},
  pages={891--927},
  year={2016},
  publisher={Springer}
}

@article{ding2026outformer,
  title={From Zero to Hero: Advancing Zero-Shot Foundation Models for Tabular Outlier Detection},
  author={Ding, Xueying and  Wen, Haomin and  Kl\"{u}ttermann, Simon and  Akoglu, Leman},
  journal={arXiv preprint arXiv:2602.03018},
  year={2026}
}

@article{emmott2015meta,
  title={A meta-analysis of the anomaly detection problem},
  author={Emmott, Andrew and Das, Shubhomoy and Dietterich, Thomas and Fern, Alan and Wong, Weng-Keen},
  journal={arXiv preprint arXiv:1503.01158},
  year={2015}
}

@inproceedings{whichtouse18,
  author={Ting, Kai Ming and Aryal, Sunil and Washio, Takashi},
  booktitle={2018 IEEE International Conference on Data Mining (ICDM)}, 
  title={Which Outlier Detector Should I use?}, 
  year={2018},
  volume={},
  number={},
  pages={8-8},
  doi={10.1109/ICDM.2018.00015}}

@article{domingues2018comparative,
  title={A comparative evaluation of outlier detection algorithms: Experiments and analyses},
  author={Domingues, R{\'e}mi and Filippone, Maurizio and Michiardi, Pietro and Zouaoui, Jihane},
  journal={Pattern recognition},
  volume={74},
  pages={406--421},
  year={2018},
  publisher={Elsevier}
}

@article{goldstein2016comparative,
  title={A comparative evaluation of unsupervised anomaly detection algorithms for multivariate data},
  author={Goldstein, Markus and Uchida, Seiichi},
  journal={PloS one},
  volume={11},
  number={4},
  year={2016},
}

@article{ma2023need,
  title={The need for unsupervised outlier model selection: A review and evaluation of internal evaluation strategies},
  author={Ma, Martin Q and Zhao, Yue and Zhang, Xiaorong and Akoglu, Leman},
  journal={ACM SIGKDD Explorations Newsletter},
  volume={25},
  number={1},
  pages={19--35},
  year={2023},
  publisher={ACM New York, NY, USA}
}

@inproceedings{
hollmann2023tabpfn,
title={Tab{PFN}: A Transformer That Solves Small Tabular Classification Problems in a Second},
author={Noah Hollmann and Samuel M{\"u}ller and Katharina Eggensperger and Frank Hutter},
booktitle={The Eleventh International Conference on Learning Representations},
year={2023},
}

@article{journals/tmlr/YooZA23,
  author = {Yoo, Jaemin and Zhao, Tiancheng and Akoglu, Leman},
  journal = {Trans. Mach. Learn. Res.},
  title = {Data Augmentation is a Hyperparameter: Cherry-picked Self-Supervision for Unsupervised Anomaly Detection is Creating the Illusion of Success.},
  volume = 2023,
  year = 2023
}

@article{han2022adbench,
  author = {Han, Songqiao and Hu, Xiyang and Huang, Hailiang and Jiang, Minqi and Zhao, Yue},
  journal = {Advances in Neural Information Processing Systems},
  title = {Adbench: Anomaly detection benchmark},
  volume = 35,
  year = 2022
}

@inproceedings{akoglu2021anomaly,
  title={Anomaly mining: Past, present and future},
  author={Akoglu, Leman},
  booktitle={Proceedings of the 30th ACM International Conference on Information \& Knowledge Management},
  pages={1--2},
  year={2021}
}

@article{hodge2004survey,
  title={A survey of outlier detection methodologies},
  author={Hodge, Victoria and Austin, Jim},
  journal={Artificial intelligence review},
  volume={22},
  number={2},
  pages={85--126},
  year={2004},
  publisher={Springer}
}

@article{eggert2023tablib,
  title={Tablib: A dataset of 627m tables with context},
  author={Eggert, Gus and Huo, Kevin and Biven, Mike and Waugh, Justin},
  journal={arXiv preprint arXiv:2310.07875},
  year={2023}
}

@misc{roechnerPosition,
      title={We Need to Rethink Benchmarking in Anomaly Detection}, 
      author={Philipp R\"{o}chner and Simon Kl\"{u}ttermann and Franz Rothlauf and Daniel Schl\"{o}r},
      year={2025},
      eprint={2507.15584},
      archivePrefix={arXiv},
      primaryClass={cs.LG},
      url={https://arxiv.org/abs/2507.15584}, 
}

@article{lhcOlympics,
   title={The LHC Olympics 2020 a community challenge for anomaly detection in high energy physics},
   volume={84},
   ISSN={1361-6633},
   url={http://dx.doi.org/10.1088/1361-6633/ac36b9},
   DOI={10.1088/1361-6633/ac36b9},
   number={12},
   journal={Reports on Progress in Physics},
   publisher={IOP Publishing},
   author={Kasieczka, Gregor and Nachman, Benjamin and Shih, David and Amram, Oz and Andreassen, Anders and Benkendorfer, Kees and Bortolato, Blaz and Brooijmans, Gustaaf and Canelli, Florencia and Collins, Jack H and Dai, Biwei and De Freitas, Felipe F and Dillon, Barry M and Dinu, Ioan-Mihail and Dong, Zhongtian and Donini, Julien and Duarte, Javier and Faroughy, D A and Gonski, Julia and Harris, Philip and Kahn, Alan and Kamenik, Jernej F and Khosa, Charanjit K and Komiske, Patrick and Le Pottier, Luc and Martín-Ramiro, Pablo and Matevc, Andrej and Metodiev, Eric and Mikuni, Vinicius and Murphy, Christopher W and Ochoa, Inês and Park, Sang Eon and Pierini, Maurizio and Rankin, Dylan and Sanz, Veronica and Sarda, Nilai and Seljak, Urŏ and Smolkovic, Aleks and Stein, George and Suarez, Cristina Mantilla and Szewc, Manuel and Thaler, Jesse and Tsan, Steven and Udrescu, Silviu-Marian and Vaslin, Louis and Vlimant, Jean-Roch and Williams, Daniel and Yunus, Mikaeel},
   year={2021},
   month=dec, pages={124201} }

@misc{MVTec2,
      title={The MVTec AD 2 Dataset: Advanced Scenarios for Unsupervised Anomaly Detection}, 
      author={Lars Heckler-Kram and Jan-Hendrik Neudeck and Ulla Scheler and Rebecca König and Carsten Steger},
      year={2025},
      eprint={2503.21622},
      archivePrefix={arXiv},
      primaryClass={cs.CV},
      url={https://arxiv.org/abs/2503.21622}, 
}

@inproceedings{devnet,
author = {Pang, Guansong and Shen, Chunhua and van den Hengel, Anton},
title = {Deep Anomaly Detection with Deviation Networks},
year = {2019},
isbn = {9781450362016},
publisher = {Association for Computing Machinery},
address = {New York, NY, USA},
url = {https://doi.org/10.1145/3292500.3330871},
doi = {10.1145/3292500.3330871},
booktitle = {Proceedings of the 25th ACM SIGKDD International Conference on Knowledge Discovery \& Data Mining},
pages = {353–362},
numpages = {10},
}

@inproceedings{dean,
author = {Kl\"{u}ttermann, Simon and Katzke, Tim and M\"{u}ller, Emmanuel},
title = {Unsupervised Surrogate Anomaly Detection},
year = {2025},
isbn = {978-3-032-05961-1},
publisher = {Springer-Verlag},
address = {Berlin, Heidelberg},
url = {https://doi.org/10.1007/978-3-032-05962-8_5},
doi = {10.1007/978-3-032-05962-8_5},
booktitle = {Machine Learning and Knowledge Discovery in Databases. Research Track: European Conference, ECML PKDD 2025, Porto, Portugal, September 15–19, 2025, Proceedings, Part I},
pages = {71–88},
numpages = {18},
location = {Porto, Portugal}
}

@inproceedings{knn,
author = {Ramaswamy, Sridhar and Rastogi, Rajeev and Shim, Kyuseok},
title = {Efficient algorithms for mining outliers from large data sets},
year = {2000},
isbn = {1581132174},
publisher = {Association for Computing Machinery},
address = {New York, NY, USA},
url = {https://doi.org/10.1145/342009.335437},
doi = {10.1145/342009.335437},
booktitle = {Proceedings of the 2000 ACM SIGMOD International Conference on Management of Data},
pages = {427–438},
}

@inproceedings{LOF,
author = {Breunig, Markus M. and Kriegel, Hans-Peter and Ng, Raymond T. and Sander, J\"{o}rg},
title = {LOF: identifying density-based local outliers},
year = {2000},
isbn = {1581132174},
publisher = {Association for Computing Machinery},
address = {New York, NY, USA},
url = {https://doi.org/10.1145/342009.335388},
doi = {10.1145/342009.335388},
booktitle = {Proceedings of the 2000 ACM SIGMOD International Conference on Management of Data},
pages = {93–104},
numpages = {12},
location = {Dallas, Texas, USA},
series = {SIGMOD '00}
}

@article{EGMM,
author = {Glodek, Michael and Schels, Martin and Schwenker, Friedhelm},
title = {Ensemble Gaussian mixture models for probability density estimation},
year = {2013},
issue_date = {February 2013},
publisher = {Kluwer Academic Publishers},
address = {USA},
volume = {28},
number = {1},
issn = {0943-4062},
url = {https://doi.org/10.1007/s00180-012-0374-5},
doi = {10.1007/s00180-012-0374-5},
journal = {Comput. Stat.},
month = feb,
pages = {127–138},
numpages = {12}
}

@inproceedings{OCSVM,
 author = {Sch\"{o}lkopf, Bernhard and Williamson, Robert C and Smola, Alex and Shawe-Taylor, John and Platt, John},
 booktitle = {Advances in Neural Information Processing Systems},
 editor = {S. Solla and T. Leen and K. M\"{u}ller},
 pages = {},
 publisher = {MIT Press},
 title = {Support Vector Method for Novelty Detection},
 url = {https://proceedings.neurips.cc/paper_files/paper/1999/file/8725fb777f25776ffa9076e44fcfd776-Paper.pdf},
 volume = {12},
 year = {1999}
}

@inproceedings{ICL,
title={Anomaly Detection for Tabular Data with Internal Contrastive Learning},
author={Tom Shenkar and Lior Wolf},
booktitle={International Conference on Learning Representations},
year={2022},
url={https://openreview.net/forum?id=_hszZbt46bT}
}

@InProceedings{NPTAD,
  title = 	 {Beyond Individual Input for Deep Anomaly Detection on Tabular Data},
  author =       {Thimonier, Hugo and Popineau, Fabrice and Rimmel, Arpad and Doan, Bich-Li\^{e}n},
  booktitle = 	 {Proceedings of the 41st International Conference on Machine Learning},
  pages = 	 {48097--48123},
  year = 	 {2024},
  editor = 	 {Salakhutdinov, Ruslan and Kolter, Zico and Heller, Katherine and Weller, Adrian and Oliver, Nuria and Scarlett, Jonathan and Berkenkamp, Felix},
  volume = 	 {235},
  series = 	 {Proceedings of Machine Learning Research},
  month = 	 {21--27 Jul},
  publisher =    {PMLR},
  url = 	 {https://proceedings.mlr.press/v235/thimonier24a.html},
}

@misc{qwen3technicalreport,
      title={Qwen3 Technical Report}, 
      author={An Yang and Anfeng Li and Baosong Yang and Beichen Zhang and Binyuan Hui and Bo Zheng and Bowen Yu and Chang Gao and Chengen Huang and Chenxu Lv and Chujie Zheng and Dayiheng Liu and Fan Zhou and Fei Huang and Feng Hu and Hao Ge and Haoran Wei and Huan Lin and Jialong Tang and Jian Yang and Jianhong Tu and Jianwei Zhang and Jianxin Yang and Jiaxi Yang and Jing Zhou and Jingren Zhou and Junyang Lin and Kai Dang and Keqin Bao and Kexin Yang and Le Yu and Lianghao Deng and Mei Li and Mingfeng Xue and Mingze Li and Pei Zhang and Peng Wang and Qin Zhu and Rui Men and Ruize Gao and Shixuan Liu and Shuang Luo and Tianhao Li and Tianyi Tang and Wenbiao Yin and Xingzhang Ren and Xinyu Wang and Xinyu Zhang and Xuancheng Ren and Yang Fan and Yang Su and Yichang Zhang and Yinger Zhang and Yu Wan and Yuqiong Liu and Zekun Wang and Zeyu Cui and Zhenru Zhang and Zhipeng Zhou and Zihan Qiu},
      year={2025},
      eprint={2505.09388},
      archivePrefix={arXiv},
      primaryClass={cs.CL},
      url={https://arxiv.org/abs/2505.09388}, 
}

@book{representativesets,
  title     = {Model Assisted Survey Sampling},
  author    = {S{\"a}rndal, Carl-Erik and Swensson, Bengt and Wretman, Jan},
  year      = {1992},
  publisher = {Springer-Verlag},
  doi       = {10.1007/978-1-4612-4378-6}
}

@Article{oodsurvey,
author={Yang, Jingkang
and Zhou, Kaiyang
and Li, Yixuan
and Liu, Ziwei},
title={Generalized Out-of-Distribution Detection: A Survey},
journal={International Journal of Computer Vision},
year={2024},
month={Dec},
day={01},
volume={132},
number={12},
pages={5635-5662},
url={https://doi.org/10.1007/s11263-024-02117-4}
}

@article{carreira2002mode,
  title={Mode-finding for mixtures of Gaussian distributions},
  author={Carreira-Perpinan, Miguel A.},
  journal={IEEE Transactions on Pattern Analysis and Machine Intelligence},
  volume={22},
  number={11},
  pages={1318--1323},
  year={2002},
  publisher={IEEE}
}

@article{bongers2021foundations,
  title={Foundations of structural causal models with cycles and latent variables},
  author={Bongers, Stephan and Forr{\'e}, Patrick and Peters, Jonas and Mooij, Joris M},
  journal={The Annals of Statistics},
  volume={49},
  number={5},
  pages={2885--2915},
  year={2021},
  publisher={Institute of Mathematical Statistics}
}

@book{peters2017elements,
  title={Elements of causal inference: foundations and learning algorithms},
  author={Peters, Jonas and Janzing, Dominik and Sch{\"o}lkopf, Bernhard},
  year={2017},
  publisher={The MIT press}
}

@incollection{scholkopf2022causality,
  title={Causality for machine learning},
  author={Sch{\"o}lkopf, Bernhard},
  booktitle={Probabilistic and causal inference: The works of Judea Pearl},
  pages={765--804},
  year={2022}
}

@article{clauset2009power,
  title={Power-law distributions in empirical data},
  author={Clauset, Aaron and Shalizi, Cosma Rohilla and Newman, Mark EJ},
  journal={SIAM review},
  volume={51},
  number={4},
  pages={661--703},
  year={2009},
  publisher={SIAM}
}

@inproceedings{sklar1959fonctions,
  title={Fonctions de r{\'e}partition {\`a} n dimensions et leurs marges},
  author={Sklar, M},
  booktitle={Annales de l'ISUP},
  volume={8},
  number={3},
  pages={229--231},
  year={1959}
}

@book{nelsen2006introduction,
  title={An introduction to copulas},
  author={Nelsen, Roger B},
  year={2006},
  publisher={Springer}
}

@article{houssou2022generation,
  title={Generation and simulation of synthetic datasets with copulas},
  author={Houssou, Regis and Augustin, Mihai-Cezar and Rappos, Efstratios and Bonvin, Vivien and Robert-Nicoud, Stephan},
  journal={arXiv preprint arXiv:2203.17250},
  year={2022}
}

@inproceedings{
zhang2025mitra,
title={Mitra: Mixed Synthetic Priors for Enhancing Tabular Foundation Models},
author={Xiyuan Zhang and Danielle C. Maddix and Junming Yin and Nick Erickson and Abdul Fatir Ansari and Boran Han and Shuai Zhang and Leman Akoglu and Christos Faloutsos and Michael W. Mahoney and Cuixiong Hu and Huzefa Rangwala and George Karypis and Bernie Wang},
booktitle={The Thirty-ninth Annual Conference on Neural Information Processing Systems},
year={2025},
}

@article{elo1967uscf,
  author  = {Elo, Arpad E.},
  title   = {The Proposed USCF Rating System, Its Development, Theory, and Applications},
  journal = {Chess Life},
  volume  = {22},
  pages   = {242--247},
  year    = {1967}
}

@article{randomForests,
author = {Breiman, Leo},
title = {Random Forests},
year = {2001},
issue_date = {October 1 2001},
publisher = {Kluwer Academic Publishers},
address = {USA},
volume = {45},
number = {1},
issn = {0885-6125},
url = {https://doi.org/10.1023/A:1010933404324},
doi = {10.1023/A:1010933404324},
journal = {Mach. Learn.},
month = oct,
pages = {5–32},
numpages = {28}
}

@misc{tabpfnRW,
      title={Real-TabPFN: Improving Tabular Foundation Models via Continued Pre-training With Real-World Data}, 
      author={Anurag Garg and Muhammad Ali and Noah Hollmann and Lennart Purucker and Samuel Müller and Frank Hutter},
      year={2025},
      eprint={2507.03971},
      archivePrefix={arXiv},
      primaryClass={cs.LG},
      url={https://arxiv.org/abs/2507.03971}, 
}

\clearpage
\newpage

\appendix 
\section*{Appendix}

\lstdefinestyle{custompython}{language=Python, basicstyle=\ttfamily\small, numbers=left, numberstyle=\tiny, stepnumber=1, numbersep=5pt, showstringspaces=false, breaklines=true, 
commentstyle=\color{gray},backgroundcolor=\color{gray!10},frame=single,keywordstyle=\rmfamily,breakindent=0pt,}

\section{ADBench Analyses}
\label{appx:adbench}

\subsection{Summary Statistics}

\begin{table}[h!]
\centering
\caption{\adb datasets summary statistics}
\vspace{-0.1in}
\begin{tabular}{lrrrr}
\toprule
 & \textbf{Min} & \textbf{Max} & \textbf{Mean} & \textbf{Median} \\
\midrule
\textbf{Samples} & 80 & 619,326 & 28,013 & 5,393 \\
\textbf{Features} & 3 & 1,555 & 371 & 512 \\
\textbf{Outlier fraction} & 0.03\% & 39.9\% & 10.0\% & 5.0\% \\
\bottomrule
\end{tabular}
\end{table}

\begin{figure}[!ht]
    \centering
\includegraphics[width=1.0\linewidth]{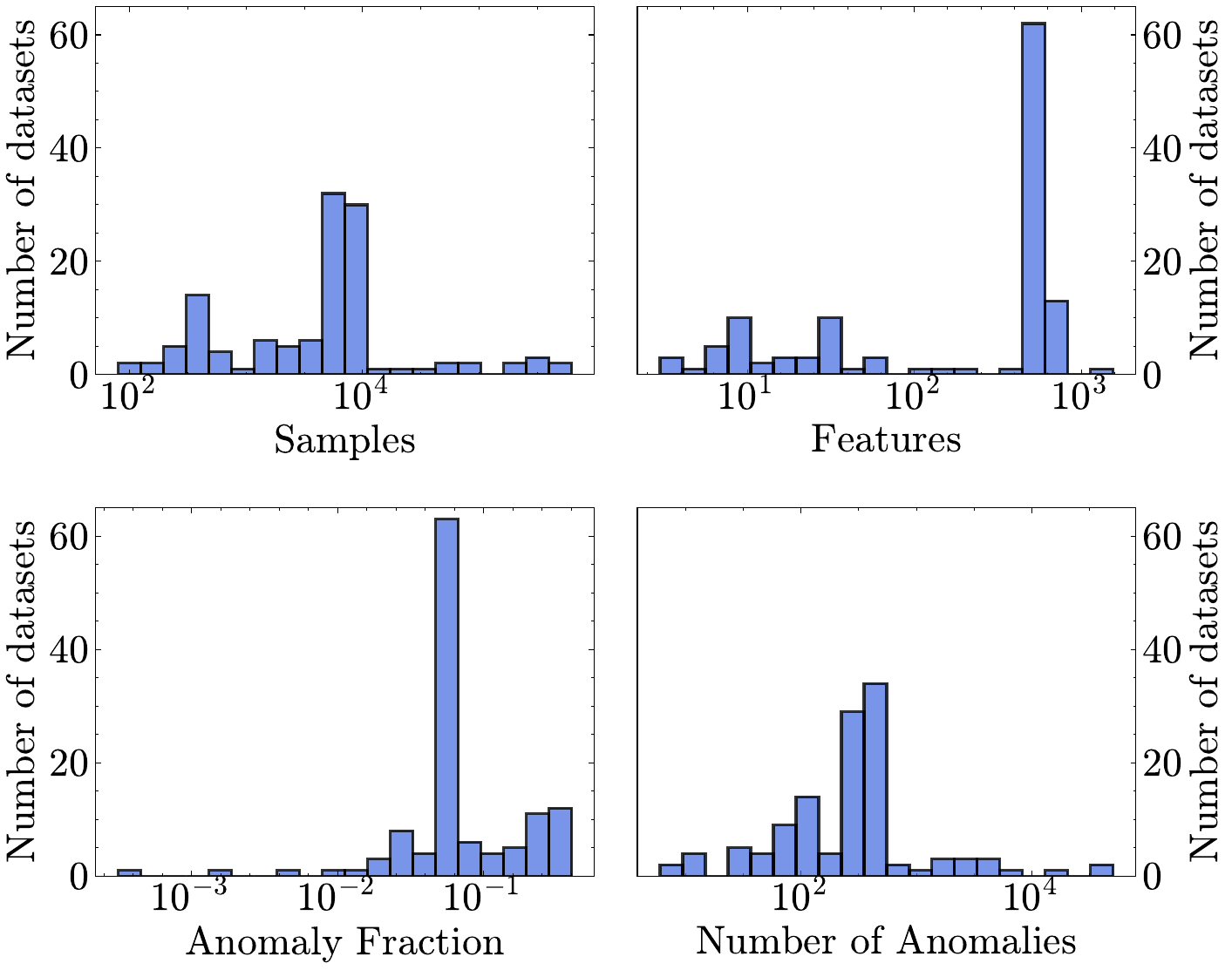}
    \caption{Summary statistics of \adb datasets.}
    \label{fig:adbench_stats}
\end{figure}

\label{appx:adbench_stats}

\subsection{Demystifying ADBench: Details}
\label{appx:adbench_demystify}

\subsubsection{\bf Delving into \dte (\citet{LivernocheJHR24})}
\label{ssec:dtedive}

$\;$

{\bf Background: Denoising Diffusion Probabilistic Models:~} A diffusion process describes a stochastic system in which the probability distribution evolves over time following the diffusion equation.
Denoising diffusion probabilistic models \cite{ho2020denoising,sohl2015deep} 
consider the states corresponding to time steps $t>0$ as latent variables.
Let \( \bx_0 \sim q(\bx_0) \) denote an observed data point sampled from the underlying data distribution and \( \bx_1, \dots, \bx_T \) denote the corresponding latent variables. The \textbf{forward diffusion} process typically adds (Gaussian) noise at each time step according to a noise (variance) schedule \( \beta_1, \dots, \beta_T \). Under Gaussian noise, the approximate posterior \( q(\bx_{1:T} \mid \bx_0) \) is written as
\beq
q(\bx_{1:T} \mid \bx_0) := \prod_{t=1}^T q(\bx_t \mid \bx_{t-1}), \; q(\bx_t \mid \bx_{t-1}) := \mathcal{N}(\bx_t; \sqrt{1 - \beta_t} \bx_{t-1}, \beta_t \mathbf{I})
\eeq

Using Gaussian distributions enables sampling \( \bx_t \)  in closed form at any time $t$ as
\beq
\label{eq:closed}
q(\bx_t \mid \bx_0) := \mathcal{N}(\bx_t; \sqrt{\bar{\alpha}_t} \bx_0, (1 - \bar{\alpha}_t) \mathbf{I}) \;,
\eeq
where \( \alpha_t := 1 - \beta_t \) and \( \bar{\alpha}_t := \prod_{s=1}^t \alpha_s \).

Diffusion probabilistic models learn the reverse process transitions, a.k.a. the \textbf{backward denoising}. 
Starting at \( p(\bx_T) = \mathcal{N}(\bx_T; 0, \mathbf{I}) \), the joint distribution of the reverse process \( p_\theta(\bx_0:T) \) decomposes into a product of terms
\beq
p_\theta(\bx_{t-1} \mid \bx_t) := \mathcal{N}(\bx_{t-1}; \; \mu_\theta(\bx_t, t),  \;\Sigma_\theta(\bx_t, t))
\eeq
After a finite sequence of transitions, the parameterized reverse process yields samples consistent with the data distribution.

{\bf Diffusion-based OD and DTE:~}
Typical diffusion-based OD approaches take as input a timestep (as hyperparameter\footnote{Note that \citet{LivernocheJHR24} show the sensitivity of diffusion-based OD algorithms to the choice of the starting timestamp hyperparameter in their Appendix A.}) from which they simulate the reverse diffusion chain to generate the reconstruction. The distance between a data point and its denoised reconstruction is used as its anomaly score. Motivated by computationally expensive reverse diffusion, \citet{LivernocheJHR24} embrace a simpler idea: \textbf{estimate the distribution over diffusion time}.
In essence, the diffusion time serves as a surrogate for the distance from the data manifold, for which they derive an analytical formula under Gaussian noise.

In the following, we analyze \dte's analytical form and its approximation in detail that establishes intriguing connections between denoising diffusion time, Gaussian noise variance and nearest neighbors. 
Consider $\bx_s \in \mathbb{R}^d$, a given input point, to be the outcome of a forward diffusion process starting from the data manifold (i.e. an inlier). How can we estimate (the distribution over) its diffusion time, that is the number of forward steps (the larger, the more outlying)? 

Assuming Gaussian noise as presented earlier in the Background, let $\sigma_t^2 = 1- \bar{\alpha}_t$ denote the variance at time $t$. Given dataset $\mathcal{D}$ sampled from the manifold, the posterior distribution over $\sigma_t^2$ can be written as
\begin{align}
    p(\sigma_t^2 & \mid \mathbf{x}_s)
 \propto p(\mathbf{x}_s \mid \sigma_t^2) p(\sigma_t^2)
= \sum_{\mathbf{x}_0} p(\mathbf{x}_s \mid \mathbf{x}_0, \sigma_t^2) p(\mathbf{x}_0)  \\
& = \sum_{\mathbf{x}_0 \in \mathcal{D}} \mathcal{N}(\mathbf{x}_s; \mathbf{x}_0, \sigma_t^2 \mathbf{I}) \;\propto  \;
\sigma_t^{-d}  \sum_{\mathbf{x}_0 \in \mathcal{D}} 
\exp\!\left(
    -\frac{\|\mathbf{x}_s - \mathbf{x}_0\|^2}{2\sigma_t^2}
\right) \label{eq:exact} \\
& \approx \;
\sigma_t^{-d} \exp\left( - \frac{1}{\sigma_t^2} \min_{x_0 \in D} \frac{\|x_s - x_0\|^2}{2} \right) \;. \label{eq:approx} 
\end{align}

The above approximation in Eq. \eqref{eq:approx}   is obtained by employing \texttt{exp} and then \texttt{log} functions preceding the \texttt{sum} in the second term of Eq. \eqref{eq:exact}, and approximating the resulting \texttt{log-sum-exp} term with the \texttt{max} function.

{\bf DTE and Gaussian Noise Outliers:~} The exact analytical posterior distribution, i.e. the first term in Eq. \eqref{eq:exact}, can be interpreted as the sum of Gaussian distribution likelihoods, each centered on a data point $\bx_0 \in \mathcal{D}$ with time-dependent variance.
Effectively, DTE identifies data points as outliers if their probability is high(er) under a Gaussian distribution with a large(r) variance. 
This is a byproduct of diffusion with Gaussian noise. Starting from an inlier $\bx_0$ on the data manifold, forward diffusion at time step $t$ is equivalent to creating a sample drawn from a Gaussian centered on $\bx_0$ and a variance that grows as a function of $t$. These are in essence simulated outliers; the larger the $t$, the more outlying.

In two variants of \dte, \citet{LivernocheJHR24} train a deep neural network to directly predict $t$ from $\bx_0$; \dte-IG assumes $t$ follows an Inverse Gamma distribution, while \dte-C categorizes $t$ into bins and simplifies the task as a classification one. To train such a network, they simulate noisy samples (i.e. outliers) from the given data points in $\mathcal{D}$ using the forward diffusion process while varying $t$, the number of diffusion steps. They argue that, thanks to the space-filling property of diffusion, different regions of the feature space are sampled at different rates mimicking a variety of potential outliers. Notably, however, those \textbf{assumed outliers are limited to Gaussian noise with variance in the \textit{full} $d$-dimensional feature space} as opposed to \textit{sub}-spaces. (See Eq. \eqref{eq:closed} where co-variance matrix is a scaled \textbf{I}dentity matrix in $\mathbb{R}^{d\times d}$.)

{\bf DTE and $k$ Nearest Neighbors:~}

The derivation of \eqref{eq:approx}  unearths an interesting connection between DTE and KNN: \textbf{the posterior over diffusion time is a function of the distance from a data point to its nearest neighbor in the dataset}. In practice, instead of the minimum squared distance to neighbors, \dte uses the average squared distance to the $k$ nearest neighbors to better approximate the log-sum-exp especially for larger time steps. 
Then, their anomaly score, the mean of this posterior distribution, is proportional to 
\beq
\frac{1}{K}\sum_{\mathbf{x}_0 \in \mathrm{KNN}(\mathbf{x}_s)}
\left\|\mathbf{x}_s - \mathbf{x}_0\right\|^2.
\eeq
The approximate approach, coined as \dte-NP, employs a non-parametric (NP) estimation.  As such, \textbf{\dte-NP produces a ranking (of data points by outlierness) 
\textit{identical} to that by KNN distance}, a classical OD approach. Notably, the KNN algorithm by \citet{knn} is slightly different as it uses the largest, i.e. $k$\textit{th} NN distance as the outlier score instead of the average. In fact, \textbf{KNN \cite{knn} is the second top algorithm reported on \adb}, following \dte-NP.
A deeper examination of \dte reveals why they perform similarly, despite appearing different.

\subsubsection{\bf Delving into \fomo (\citet{shen2025fomod})}
\label{ssec:fomodive}

$\;$

Next we analyze \fomo \cite{shen2025fomod} which performed competitively on \adb, closely rivaling  \dte (on average across hyperparameters) and KNN (2nd top performer).  For easy reference, we present the comparison in Table \ref{tab:aucrocD100}, which reports the average rank based on AUROC across \textsf{All} (57) datasets, along with the $p$-values of the one-sided Wilcoxon signed rank test, comparing \fomo to \dte-NP and KNN with author-recommended hyperparameter $k=5$, and 
 with \textbf{avg.} performance over varying HPs (denoted w/ superscript \avg). We refer to the original paper for the details on these results.

\begin{table}[H]
  \centering
  \caption{ \fomo vs.  \dte-NP and KNN  on \adb, based on the average rank by AUROC across \textsf{All} (57) datasets and $p$-values of the one-sided Wilcoxon signed rank test. }
  \vspace{-0.1in}
\setlength\tabcolsep{2 pt}
\begin{tabular}{c|c||cc|cc}
\toprule
 & \fomo   & DTE-NP & KNN &   DTE-NP\avg & KNN\avg  \\ \midrule

Rank(avg) & 11.886 & 7.553 & 9.018  & 9.079 & 11.105  \\ 
\hline

\textsf{All}  & - & \underline{0.016} & 0.106 & 0.112 & 0.315  \\ 

{\textsf{All}$\setminus$\{NLP, CV\}}  & - & 
{0.164} & {0.476} & {0.515} & 
{0.683}  \\

$d\leq 100$ & - & 0.415 & 0.700 & 0.752 & 0.860  \\

\bottomrule 
    \end{tabular}  \label{tab:aucrocD100}
  \end{table}

The $p$-values on \textsf{All} (57) datasets suggest that \fomo shows  no statistically significant difference from the \textbf{second top} model KNN on \adb at $p=0.106$. Further,
the differences vanish when (the embedded, high-dimensional) NLP (i.e. text) and CV (i.e. image) datasets are excluded. In fact, over datasets with up to 100 dimensions, which aligns with \fomo's pretraining, the $p$-values become notably large.
We also remark that \fomo performs similarly to \dte-NP\avg~ and KNN\avg, where their expected performance is computed over various $k$ values (as opposed to favorably chosen $k=5$ in \cite{LivernocheJHR24}) as suggested by the relatively larger ranks and $p$-values.  
Note that \fomo is a foundation model that labels test points via a single forward pass given training points as context, requiring no hyperparameters to be tuned (nor any model to be trained from scratch).

Revisiting our original goal: \textit{What makes \fomo achieve as competitive results as these SOTA methods on \adb, considering it was pretrained purely on synthetic datasets?} In fact, the astonishing performance of \fomo on real-world datasets has been an unresolved question until the detailed analysis of \dte; in particular, the type of outliers it captures. 

The key to the puzzle is the data prior from which the synthetic datasets used to pretrain \fomo are drawn. Specifically, Gaussian mixture models (GMMs) with \textit{diagonal} covariances are used to synthesize inliers, while the outliers are sampled from the same GMM with \textit{inflated variances} in random sub-spaces, ensuring they lie outside the 90th percentile of the original GMM \cite{shen2025fomod}. 
Overall, \textbf{\fomo performs alike \dte (and by extension KNN) as it likewise models outliers as generated by Gaussian noise.}

\section{Proposed Benchmark Details}
\label{appx:detail_bench}

\subsection{Real-World Benchmarks: \fraudbench and \ovrbench} \label{appx:detail_fraudbench}

\subsubsection{Keywords that indicate anomalousness:} \label{appx:anomaly_keywords}
To test whether a feature value or a metadata is likely related to outlier detection, we check whether such a string contains one of the following words:


\begin{Verbatim}[bgcolor=lightgrey, breaklines=true, fontfamily=courier]
"fraud", "intrusion", "attack", "malware", "spam", "phishing", "defect", "failure", "fault", "error", "bug", "outlier", "anomaly", "abnormal", "irregular", "rare",  "deviation", "exception",  "deviation", "irregularity", "abnormality", "flaw", "disturbance", "variance", "misfire", "oddity", "discrepancy", "dissonance", "unusual", "quirk", "oddity", "peculiarity", "nonconformity", "misfit", "aberration", "mistake", "fault", "glitch", "hiccup", "error", "breakdown", "defect", "anomalies", "invalid", "vulnerabl", "breach", "exception", "fail", "critical", "extreme"
\end{Verbatim}

\subsubsection{Duplicate Check} \label{appx:duplication}
Tablib~\cite{eggert2023tablib} contains a collection of all datasets from GitHub, without any duplication check. To solve this, we design a custom hashing function that is applied to each feature. When two datasets both contain a feature with the same hash value, we randomly drop one of these datasets. Our hashing function computes the average of five simple functions and considers two features equivalent if they share all five resulting values. The five functions we use are $\sin(x)$, $\cos(e\cdot x)$, $\tfrac{x}{1+|x|}$, $\arctan(x)$, $\log(|x|+1)$.

We design this hashing procedure so that dropping or shuffling features does not resolve potential hash conflicts. However, small variations (e.g., NaN imputation, Subset selection) still conceal hash conflicts. Thus, we also manually filter for duplicates.

\subsubsection{Metadata} \label{appx:metadata}
To improve the usability of \fraudbench, we add both feature names, a tablib identifier, and a link to the original dataset we processed to each dataset (when they are available). We also employ an LLM to generate short names for each dataset, and further let the LLM categorize the origin of each dataset into one of the following categories: "Engineering", "Science", "Logistics", "Human", or "Other", and add tags to the data that describe different scenarios. A word cloud of keywords generated by the same LLM is shown in Figure~\ref{fig:word_cloud_oddbench}, while a distribution of tags is given in Figure~\ref{fig:Tags}.

The Large Language Model used here is (Qwen 3- 8B\cite{qwen3technicalreport}). For each dataset it receives the following prompt:

\begin{lstlisting}[style=custompython]
I am trying to create anomaly detection datasets. For this, I parse the meta information of a large amount of tabular data and extract datasets that are suitable for anomaly detection. Now, given these datasets, I would like to expand each dataset slightly. 
For this, I would like to have short, memorable, meaningful names for each dataset ("dataset\_name"). Examples of these might by "PlanetaryMotion" , "MonsterType" or "LifeExpectancy". So these names should be in CamelCase without spaces.
Additionally, I would like to have 5-10 keywords that describe the specific context of the dataset ("keywords"). Examples of keywords might be ["planets", "orbit", "gravity"] or ["monsters", "type", "strength", "weakness"].
Finally, I would also like to categorize the datasets by the domain they cover/ their origin ("origin"). Valid answers here are ["Logistics","Engineering","Science","Human","Other"]. Answer only with a single category out of these options.
Please answer in json format like this:
{
  "reasoning": "Your reasoning here explaining why you chose these values",
  "keywords": ["keyword1", "keyword2", "..."],
  "dataset_name": "DatasetName",
  "origin": "label_to_use",
}
Do not include anything else except this json object in your answer.
\end{lstlisting}

\subsubsection{Finding representative sets}\label{appx:representative}
Finding the best representative set by minimizing Equation~\ref{eqn:representative} requires evaluating $\binom{690}{50}$ different subsets. As it is thus infeasible to find the best representative set, we instead use an evolutionary algorithm to approximate it. 

This evolutionary algorithm evaluates $1M$ different candidate sets $\mathcal{S}$. In each step, it either randomly generates a subset ($50\%$ probability) or modifies a previously generated subset by switching one element for another random one ($50\%$ probability). We retain both the best- and worst-performing subsets ($1\%$ probability) to improve the robustness of our optimization against local minima, but finally only report the best performing subset.




\subsubsection{Sources of datasets used in \ovrbench}\label{appx:onevsrestbench}

We state the origin of each dataset from established tabular qualification benchmarks in \ovrbench in Table~\ref{tab:onevsrest_data_with_label_log}.

    

    

\begin{table}[htbp]
\centering
\caption{Number of datasets used for \onevsrestbench excluding Tablib~\cite{eggert2023tablib}.}
    \begin{tabular}{lccc}
    \toprule
    \textbf{Data Source} & \textbf{Year} & \textbf{\# Original}  & \textbf{\# Used} \\
    \hline

    OpenML-CC18~\cite{bischl2017openml} & 2017 & 72 & 1 \\
    AutoML~\cite{gijsbers2019open}& 2019 &  39 & 0   \\
    TabZilla~\cite{mcelfresh2023neural} & 2023 & 36 & 3\\
    Talent-CLS~\cite{liu2024talent} & 2024 & 180 & 52\\
     TabRepo~\cite{salinas2024tabrepo} & 2024 & 246  & 95\\
    
    BCCO-CLS~\cite{zhang2025limix} & 2025 & 106& 28\\
    TabArena~\cite{erickson2025tabarena} & 2025 & 37 & 22\\ 
    \midrule
     Total & - & 716 & 201\\
    \bottomrule
    \end{tabular}
\label{tab:onevsrest_data_with_label_log}
\end{table}

Note that datasets can overlap between different data sources. If a dataset (by dataset name) appears in multiple benchmarks, we keep only the version from the newest source.

\subsubsection{Analysis of \fraudbench and \ovrbench} \label{appx:fraudbench_distribution}

This section assesses the characteristics and performance baselines of the proposed real-world benchmarks.

\begin{figure}[!t]
    \centering
    \hspace{-0.075in}\subfigure[\oddbench]{
        \includegraphics[width=0.79\linewidth]{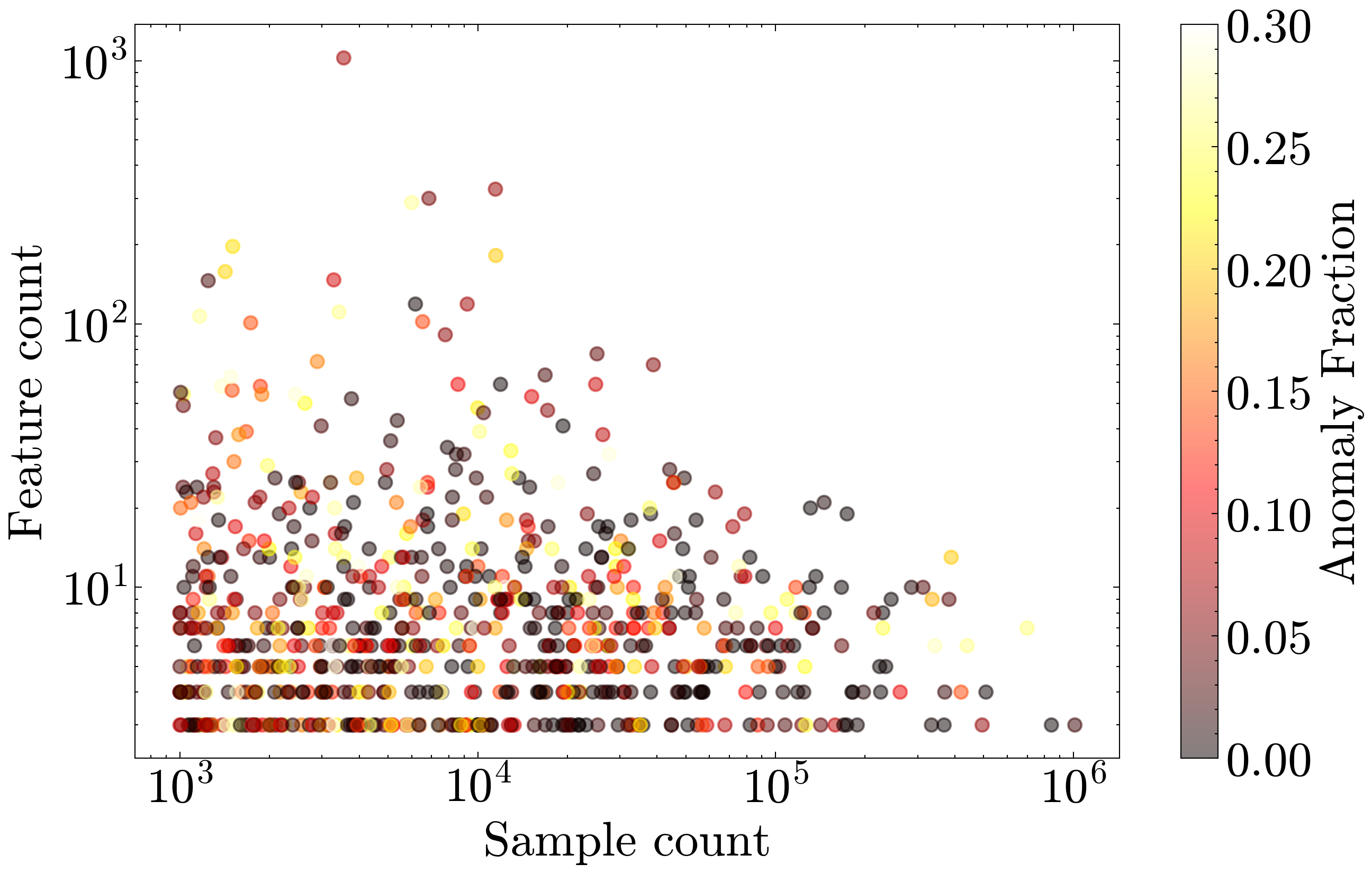}
        \label{fig:NvsD_odd}
    }
    \hfill
    \hspace{-0.05in}
    \subfigure[\ovrbench]{
        \includegraphics[width=0.79\linewidth]{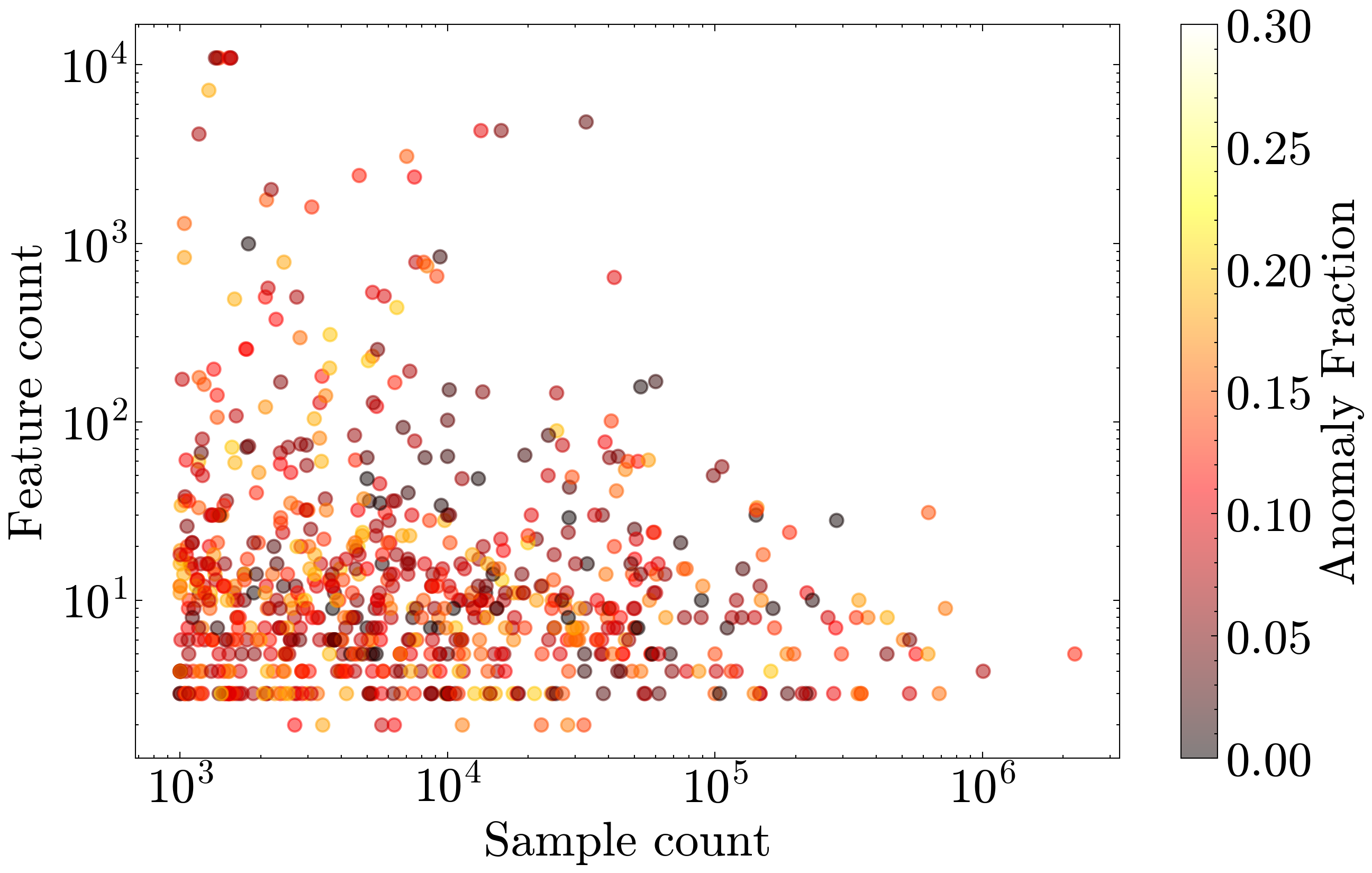}
        \label{fig:NvsD_ovr}
    }
    \vspace{-0.15in}
    \caption{Dataset statistics of (a) \oddbench and (b) \ovrbench, showing a high range of anomaly detection settings.}
    \label{fig:NvsD}
    \vspace{-0.1in}
\end{figure}

\begin{figure}[!t]
    \centering
    \hspace{-0.075in}\subfigure[\oddbench]{
        \includegraphics[width=0.79\linewidth]{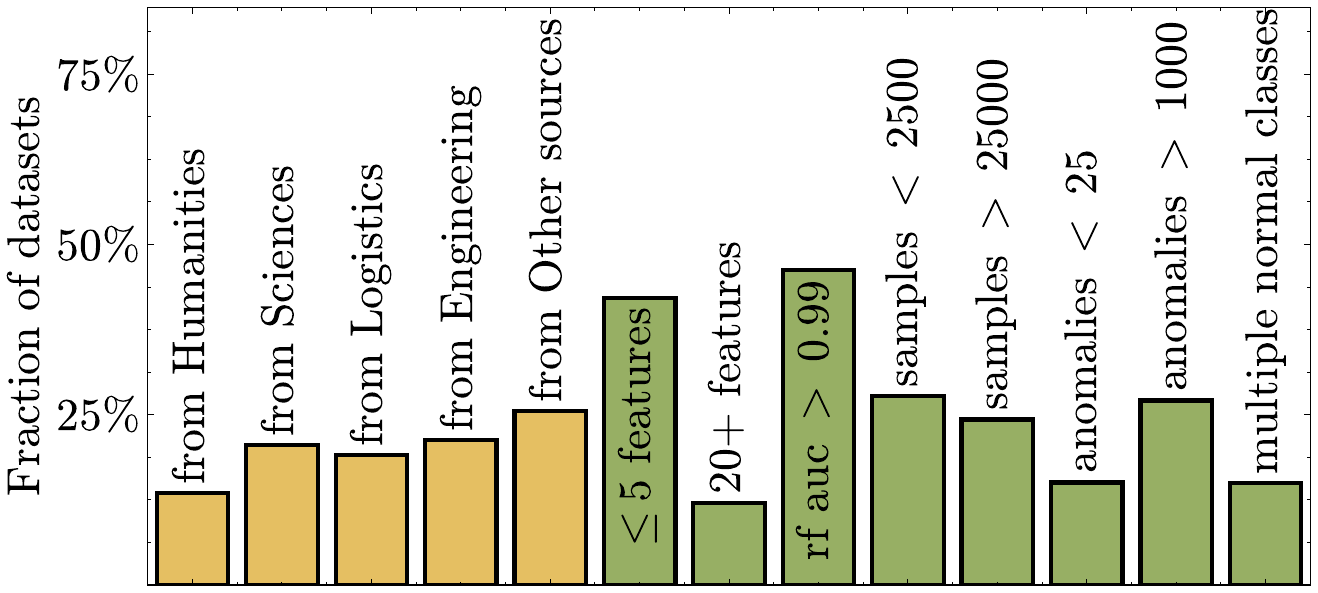}
        \label{fig:Tags_odd}
    }
    \hfill
    \hspace{-0.05in}
    \subfigure[\ovrbench]{
        \includegraphics[width=0.79\linewidth]{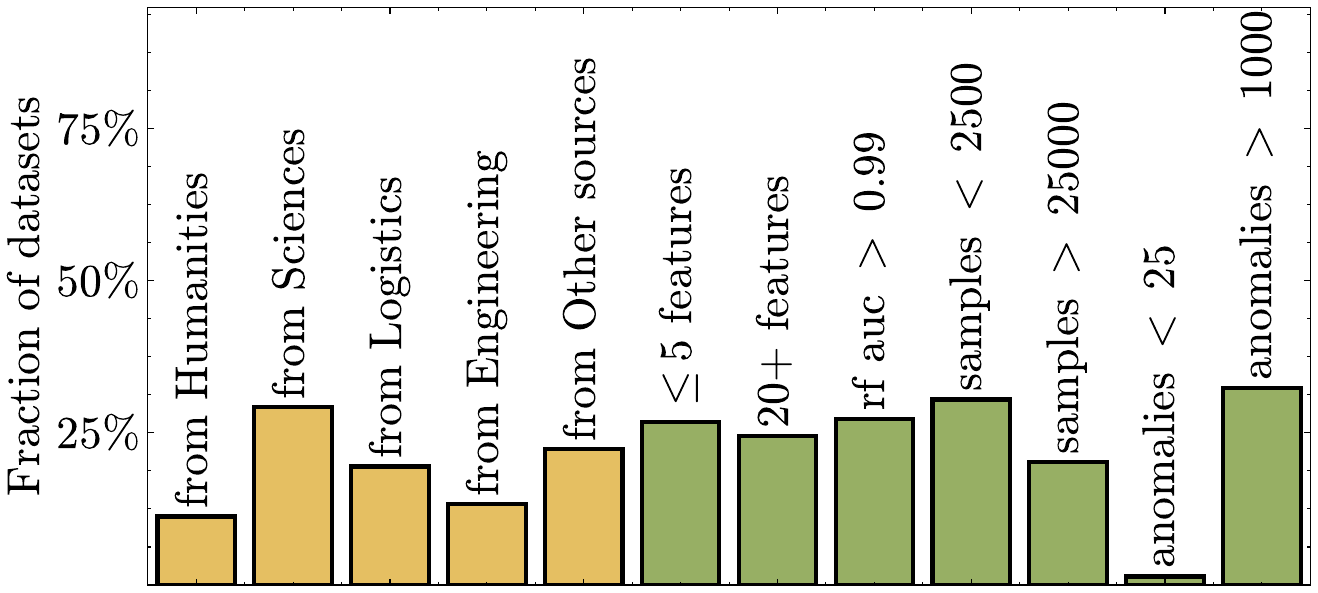}
        \label{fig:Tags_ovr}
    }
    \vspace{-0.15in}
    \caption{Dataset origin (yellow) and situational tags (green) of (a) \oddbench and (b) \ovrbench, enabling specialized research for many different settings.}
    \label{fig:Tags}
    \vspace{-0.1in}
\end{figure}

Figure ~\ref{fig:NvsD} demonstrates that both \oddbench and \ovrbench encompass a wide range of different dataset characteristics. These characteristics are both semantic and statistical (Figure ~\ref{fig:Tags}), ensuring the suitability of our benchmarks across a wide range of outlier detection applications.

\begin{figure}[!t]
    \centering
    \hspace{-0.075in}\subfigure[\oddbench]{
        \includegraphics[width=0.79\linewidth]{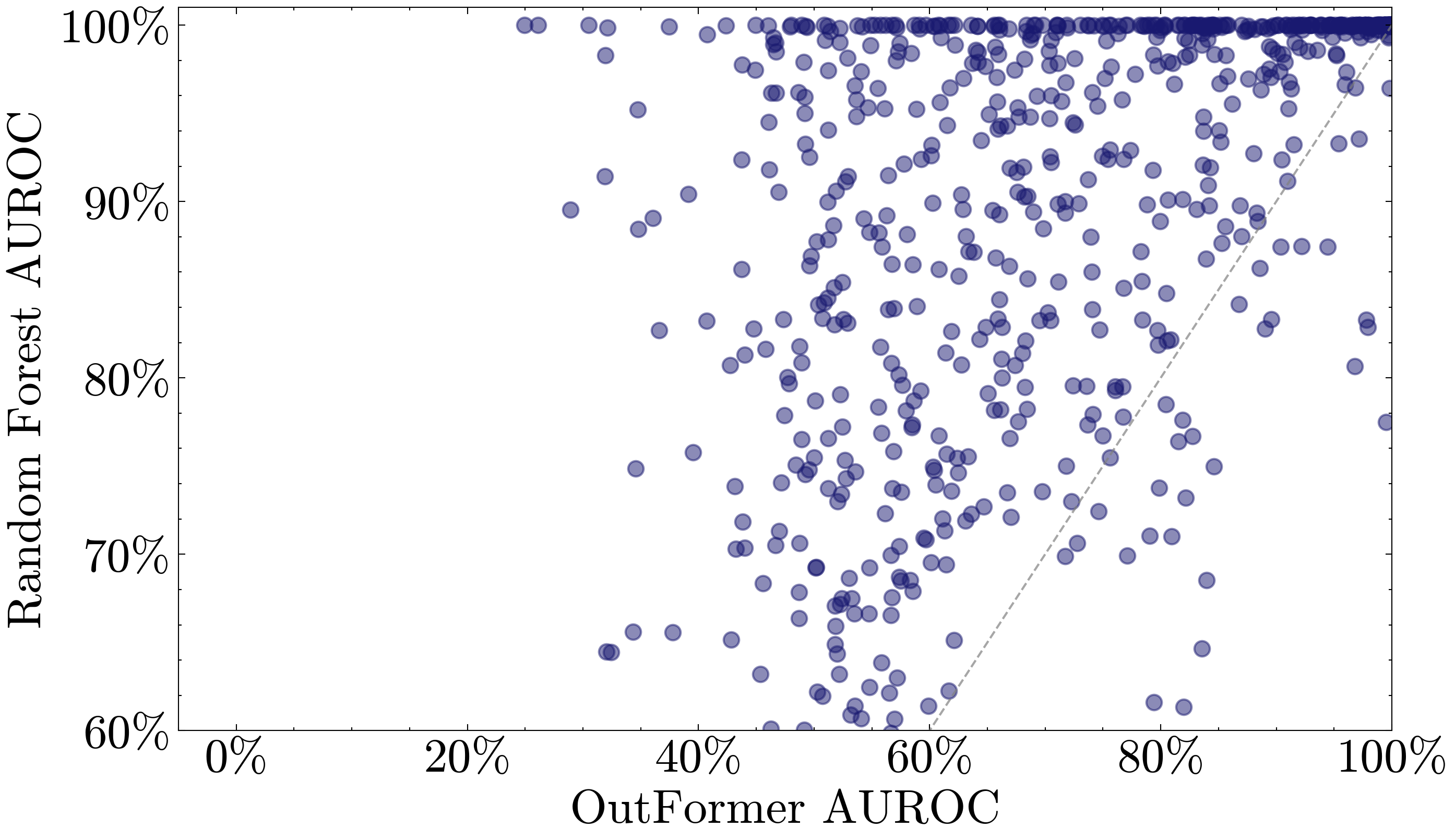}
        \label{fig:RFperf_odd}
    }
    \hfill
    \hspace{-0.05in}
    \subfigure[\ovrbench]{
        \includegraphics[width=0.79\linewidth]{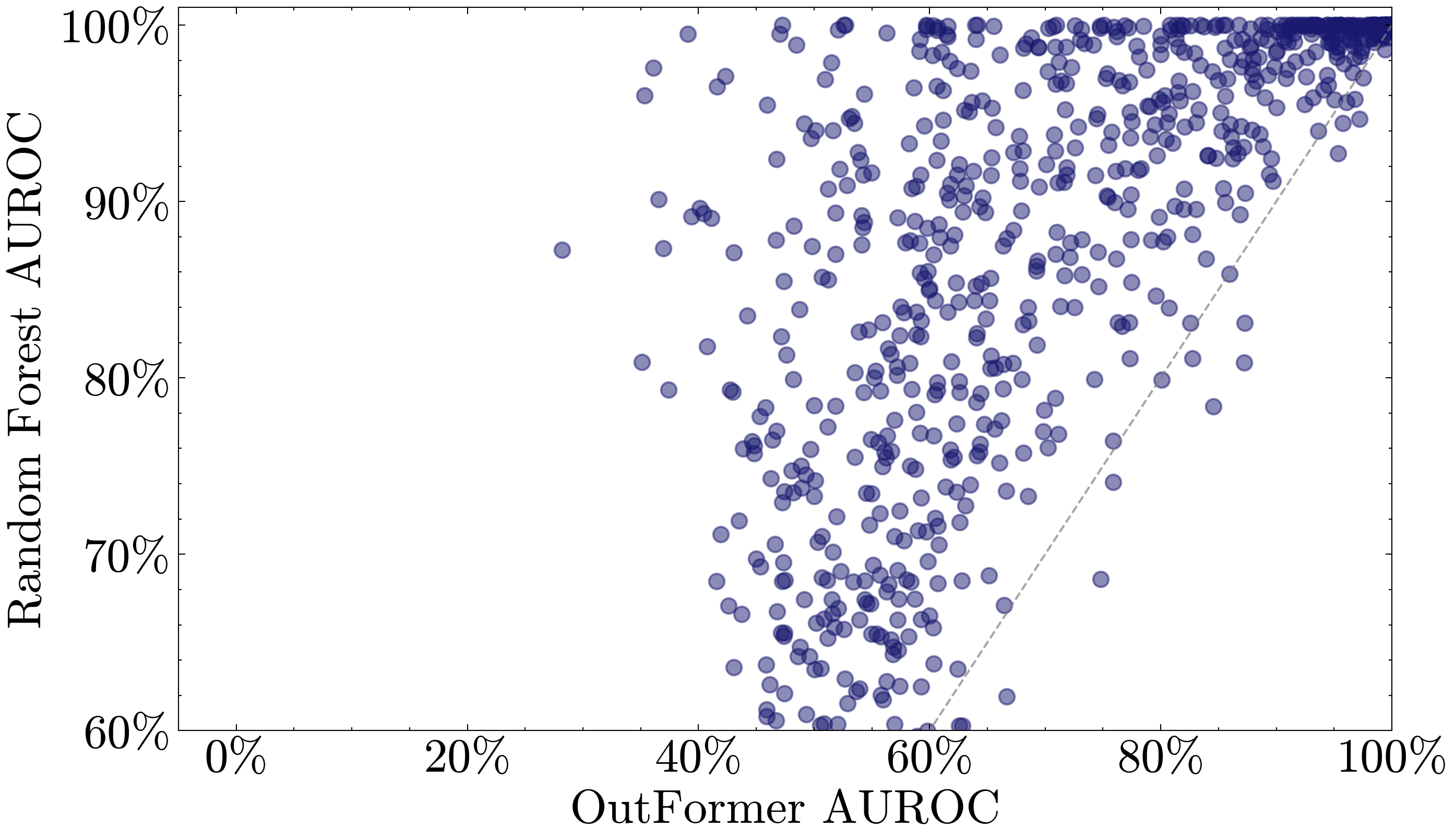}
        \label{fig:RFperf_ovr}
    }
    \vspace{-0.15in}
    \caption{Random Forest (supervised classification) performance versus 
    \outform (one-class outlier detection) performance on (a) \oddbench and (b) \ovrbench show the high potential for improvement of unsupervised methods that our benchmarks allow.}
    \label{fig:RFperformance}
    \vspace{-0.1in}
\end{figure}

Further, we evaluate the performance of the best one-class algorithm of our study (\outform, see Sec.~\ref{sec:experiments}) against a simple supervised baseline (a random forest~\cite{randomForests}) in Figure~\ref{fig:RFperformance}. Despite the theoretical advantage of one-class methods in low-anomaly regimes (Figure ~\ref{fig:crownjewel}), the random forest still consistently outperforms the one-class approach. This performance gap highlights a significant deficiency in current outlier detection methods and positions our benchmarks as critical tools for guiding future algorithmic advancements.

\subsection{Synthetic Benchmark:  \synbench} \label{appx:synbench}

We simulate 800 datasets spanning five outlier categories (GMM, SCM-Measurement, SCM-Structural, Copula-Probabilistic, and Copula-Dependence) with 160 datasets per category. Within each category, we diversify both inlier and outlier distributions by sampling feature dimensions from [2,100], outlier ratios from 
5\% to 15\%, and dataset sizes between 1,000
and 6,000 samples.

\subsubsection{Gaussian Mixtures}

\paragraph{Inliers.} We simulate inliers by sampling from multivariate Gaussian mixture models (GMMs) with varying numbers of components and dimensions. Each dataset is generated from an $m$-component GMM in $d$ dimensions, where component means $\boldsymbol{\mu}^{(k)} \in [-5,5]^d$ and diagonal covariances $\boldsymbol{\Sigma}^{(k)}$ satisfy $\Sigma^{(k)}_{jj} \in (0,5]$ for all $k \in [m]$ and $j \in [d]$. For each dataset, we sample $m$ and $d$ uniformly, with up to $M=5$ mixture components and $D=100$ dimensions. To increase diversity, we apply a linear transformation
$
T(\mathbf{x}) = \mathbf{W}\mathbf{x} + \mathbf{b}
$,
where $\mathbf{W} \in \mathbb{R}^{d \times d}$ and $\mathbf{b} \in \mathbb{R}^d$ have entries independently drawn from $\mathrm{Unif}[-1,1]$. This transformation induces a new GMM with transformed means $T(\boldsymbol{\mu}^{(k)}) = \mathbf{W}\boldsymbol{\mu}^{(k)} + \mathbf{b}$ and full (non-diagonal) covariances
$
T(\boldsymbol{\Sigma}^{(k)}) = \mathbf{W}\boldsymbol{\Sigma}^{(k)}\mathbf{W}^\top$, $\forall k \in [m]$.

\paragraph{Outliers.} We generate \textit{contextual subspace outliers} by first randomly selecting a GMM component $k$ and a subset of dimensions $\mathcal{S} \subseteq [d]$. We then inflate the variances of the selected component along the chosen subspace by scaling the diagonal entries as $s\,\Sigma^{(k)}_{jj}$ for all $j \in \mathcal{S}$, while keeping the component mean unchanged. Samples are drawn from this variance-inflated GMM, and points are labeled as outliers if their Mahalanobis distance exceeds the $90^{\text{th}}$ percentile of the original (uninflated) GMM. The severity of the generated outliers is controlled by the subspace fraction $|\mathcal{S}| = \alpha d$ and the inflation factor $s$, which govern the dimensional extent and magnitude of the perturbation, respectively. Table \ref{tab:gmmhps} provides the range of hyperparameter configurations for sampling datasets from the GMM prior.

\begin{table*}[h!]
\centering
\caption{Hyperparameters for Gaussian Mixture Model (GMM) Data Prior}
\vspace{-0.1in}
\begin{tabular}{lll}
\toprule
\textbf{Hyperparameter} & \textbf{Values} & \textbf{Description} \\
\midrule
$m$ & $[1,5]$ & Number of mixture components \\
$d$ & $[2,100]$ & Dimensionality of data \\
$\bmu_{j}^{(k)}$ & $[-5,5]$ & Component means \\
$\bSigma_{jj}^{(k)}$ & $(0,5]$ & Diagonal variances \\ \hline 
$\alpha$ & $[\frac{1}{d},1]$ & Fraction of dimensions for variance inflation \\
$s$ & $[5,10]$ & Variance inflation factor for subspace outliers \\
$r$ & $[0.02,0.2]$ & Outlier rate of contamination \\
\bottomrule
\end{tabular}
\label{tab:gmmhps}
\end{table*}

\subsubsection{Structural Causal Models}
\paragraph{Inliers.} Structural causal models (SCMs) characterize data-generating processes by explicitly modeling causal dependencies via directed acyclic graphs (DAGs) and structural equations. Formally, an SCM is defined by a graph \( G = (V, E) \), where each node \( j \in V \) represents a variable governed by the structural equation
$
X_j = f_j\!\left(X_{\mathrm{Pa}(X_j; G)}, \epsilon_j\right)$, with \( \mathrm{Pa}(X_j; G) \) denoting the set of parent variables of \( X_j \) in \( G \), and \( \epsilon_j \) capturing exogenous noise. To generate SCM inliers, we first instantiate the causal graph \( G \) using a multilayer perceptron (MLP). Sparsity is induced by applying a binary weight mask with a ratio sampled from \( (0.4, 0.6) \), yielding a sparse dependency structure over a selected set of \( d \) nodes and producing \( d \)-dimensional observations. Nonlinear causal mechanisms are modeled via an activation function \( a(\cdot) \). After initialization, the MLP weights are frozen to simulate a fixed causal mechanism. Inlier samples are generated by forward-propagating through the MLP and subsequently flattened into multi-dimensional feature vectors.

\paragraph{Measurement Outliers.} To generate measurement outliers, we randomly select a variable \( X_j \in \mathcal{X} \) and resample its exogenous noise as \( \epsilon_j \sim \mathcal{N}(0, s) \), thereby inflating its variance by \( s \). The induced perturbation is then propagated to all descendant variables through the fixed structural equations, producing a multivariate outlier. This process models exogenous shocks while preserving the underlying causal structure and mechanisms. For each SCM, outliers are generated by applying this procedure to a random fraction \( r \) of the nodes.

\begin{figure}[htbp]
    \centering
    \hspace{-0.2in}
    \begin{minipage}[t]{0.45\linewidth}
        \centering
        \includegraphics[width=\linewidth]{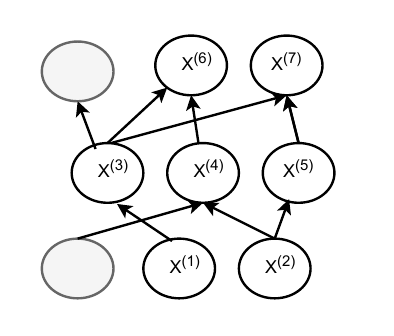}
    \end{minipage}
    \hspace{-0.2in}
    \begin{minipage}[t]{0.46\linewidth}
        \centering
        \includegraphics[width=\linewidth]{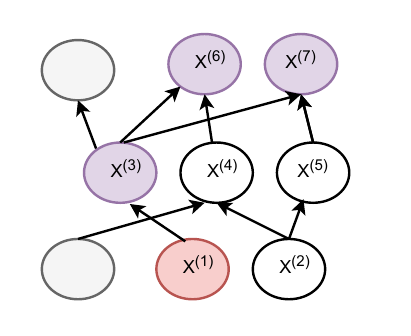}
    \end{minipage}
    \hspace{-0.2in}
    \begin{minipage}[t]{0.47\linewidth}
        \centering
        \includegraphics[width=\linewidth]{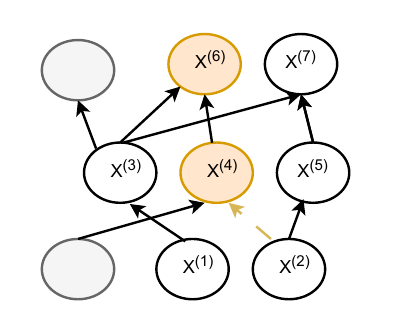}
    \end{minipage}
    \caption{\textbf{(Top-Left):} We construct an SCM and select a subset of seven nodes, from which 7-dimensional inlier samples are generated under a frozen causal structure. 
\textbf{(Top-Right):} Measurement outliers are produced by inflating the variance of node \( x^{(1)} \), causing the perturbation to propagate to \( x^{(3)} \), \( x^{(6)} \), and \( x^{(7)} \). 
\textbf{(Bottom):} Structural outliers are generated by breaking the causal link between \( x^{(2)} \) and \( x^{(4)} \), thereby altering the dependency node \( x^{(6)} \).
 }
    \label{fig:scm_demonstration}
\end{figure}

\paragraph{Structural Outliers.} Structural outliers arise from changes in the underlying causal mechanisms. We generate such outliers by intervening on the MLP-based SCM, either by \emph{breaking} a causal edge (setting its weight to zero) or by \emph{reversing} its direction (multiplying the weight by $-1$). These interventions modify parent--child relationships and thus alter the data-generating process. We apply edge perturbations with total probability $0.2$ (split evenly between breaking and reversing) and enforce a filtering criterion that ensures at least one leaf node is affected. Figure \ref{fig:scm_demonstration} provides an example of generated SCMs, and Table \ref{tab:scmhps} provides the hyperparameter settings of the SCM priors.

\begin{table*}[h!]
\centering
\caption{Hyperparameters for Structural Causal Model (SCM) based Data Priors}
\vspace{-0.1in}
\begin{tabular}{lll}
\toprule
\textbf{Hyperparameter} & \textbf{Values} & \textbf{Description} \\
\midrule
MLP depth & $[3,5]$ & Number of layers used to generate the DAG \\
MLP width & $[20,40]$ & Number of nodes per layer \\
Edge-drop rate & $[0.4,0.6]$ & Fraction of edges randomly removed to form the DAG \\
$d$ & $[2,100]$ & Number of selected nodes as features $\mathcal{X}$ \\
$a(\cdot)$ & $\{\text{ReLU, tanh, sigmoid}\}$ & Activation function for structural equations  \\
\hline 
$s$ & $[5,10]$ & Variance inflation factor for measurement outliers \\
$p_{\text{break}}$ & $[0.1]$ & Probability of removing a causal edge for structural outliers \\
$p_{\text{flip}}$ & $[0.1]$ & Probability of reversing a causal edge for structural outliers \\
$r$ & $[0.02,0.2]$ & Outlier rate of contamination \\
\bottomrule
\end{tabular}
\label{tab:scmhps}
\end{table*}
\begin{table*}[htbp]
\caption{Hyperparameters for Copula Data Prior}
\vspace{-0.1in}
\centering
\scalebox{0.9}{
\begin{tabular}{l c l}
\hline
\textbf{Hyperparameter} & \textbf{Values} & \textbf{Description} \\
\toprule
$d$ & $[2,100]$ & Number of features / copula dimension \\
$C$  & $\{\text{Gaussian, Vine (bivariate)}\}$ & Copula parameter family \\
$\alpha_{\text{indp}}$ & $[0.1, 0.3]$ & Fraction of features set independent in  copula\\
Bivariate vine & {Gauss., Stu., Clayton, Gumbel, Frank, Joe} & Bivariate family for each Vine edge \\
Marginal & $\{\text{Gauss., Beta, Exp., Stu. t, Power-law, Log-logistic}\}$ & Marginal family for each feature \\
$\mu$/$\sigma$ & $(-1, 1)$/$(0.5, 1.0)$ & mean/variance for Gaussian\\
$a$/$b$ & $(1, 5)$/$(1,5)$ & for Beta\\
$loc, scale$  & -5, 10 & Variables rescaled as $loc+scale\times x_{\text{beta}}$ for effective range $[-5, 5]$\\
$\lambda$ / $loc$ & $(0.5, 1.0)$, -5 & Exponential distribution scale / shift as $loc+x_{\text{exp}}$ for effective range \\
df & $(3, 10)$ & Degrees of freedom for Student's $t$\\
$loc, scale$  & $[-1, 1]$, $[0.5, 1.0]$  & Variables rescaled as $loc+scale\times x_{\text{stu}}$ for effective range\\
$a$ / $loc$ \& $scale$ & $(0.5, 5)$ / -5 \& 5& Power law exponent, location \& scale for effective range \\
$c$ / $loc$ \& $scale$ & $(0.5, 5)$ / -5 \& 5& Log-logistic shape, location \& scale for effective range 
\\\midrule
$\gamma_{\text{perturb}}$ & $[0.02,0.2]$ & Fraction of dimensions to perturb for probabilistic outliers \\
$u_{\text{perturb}}$ &$[0.1,0.3]$ or $[0.7,0.9]$& 
CDF perturbation for probabilistic outliers \\
Copula perturbation & $\{ \text{inverse\_corr}, \text{random\_permutation} \}$ & Types of dependence outliers \\
$k_{\textbf{invcorr}}$ & $[1,67)$ & Num of dimensions to invert for dependence outliers \\ 
\bottomrule
\end{tabular}
}
\label{tab:copulahps}
\end{table*}

\subsubsection{Copulas}

\paragraph{Inliers.}
Real-world univariate features often exhibit skewed, heavy-tailed distributions~\cite{clauset2009power}. Although SCMs yield non-Gaussian marginals, they do not offer explicit control over feature-wise distributions; we therefore employ copula models~\cite{nelsen2006introduction,houssou2022generation}. By Sklar’s theorem~\citep{sklar1959fonctions}, any joint CDF of continuous variables $\{X_j\}_{j=1}^d$ with marginals $\{F_j\}$ can be written as
$
F(x_1,\ldots,x_d) = C\!\left(F_1(x_1),\ldots,F_d(x_d)\right)
$,
allowing the marginal distributions and dependence structure to be specified independently. We draw each marginal from a diverse set of parametric families (Gaussian, Beta, Exponential, Student’s $t$, Power-law, and Log-logistic) and model dependence using either a Gaussian or (bivariate) vine copula. Inliers are generated by sampling $\mathbf{u} \sim C$ on $[0,1]^d$ and applying the inverse transforms $x_j = F_j^{-1}(u_j)$. By varying both the marginals and the copula, we synthesize a wide range of realistic tabular inlier distributions.

\paragraph{Probabilistic Outliers.} To generate probabilistic outliers, we randomly select a subset of features and perturb their copula coordinates by shifting them toward the boundaries, replacing $u_j$ with small values $u_{\text{perturb}} \in [0.1,0.3]$ or large values $u_{\text{perturb}} \in [0.7,0.9]$. The perturbation is applied to a fraction $\gamma_{\text{perturb}} \in [0.02,0.2]$ of the dimensions, with the constraint that at least one feature is always perturbed.

\paragraph{Dependence Outliers.}To generate dependence outliers, we alter the copula coordinates to break the original dependence structure, either by inverting selected dimensions (i.e., \( u_j \leftarrow 1 - u_j \)) or by randomly permuting them. The number of perturbed dimensions is sampled as
$
k_{\text{invcorr}} \in \left[ \left\lfloor 1 + \tfrac{d}{3} \right\rfloor,\; \min\!\left( \left\lfloor 1 + \tfrac{2d}{3} \right\rfloor,\; d \right) \right),
$ corresponding to approximately 33\%--67\% of the total dimensions. Table \ref{tab:copulahps} provides the hyperparameter settings of the Copula priors.

\section{Experiment Setup Details}
\label{appx:exp_details}

\subsection{Selected Baselines and HP Configurations}
\label{appx:baselines}

Table \ref{tab:baselines} gives an overview of the benchmark OD baselines. Below we briefly describe each of the baselines and our training/evaluation configurations, with a detailed list of HPs in Table \ref{tab:hps}.

\paragraph{OCSVM} \citep{OCSVM} learns a decision boundary by maximizing the margin between the origin and the normal data, effectively enclosing the region of high data density. The kernel (linear vs. RBF) controls whether this boundary is linear or nonlinear, $\nu$ specifies an upper bound on the fraction of outliers (and a lower bound on support vectors), and $\gamma$ (RBF only) determines the smoothness of the boundary, with larger values yielding more localized, complex decision surfaces.

\paragraph{KNN} \citep{knn} assigns an anomaly score to each instance based on its distance to the 
$k$-th nearest neighbor, with larger distances indicating greater anomaly likelihood.

\paragraph{LOF} \citep{LOF} measures the degree of anomaly of a sample by comparing its local density to that of its neighbors. The hyperparameter 
$N_\text{neighbors}$ specifies the size of the local neighborhood used for density estimation, where smaller values focus on finer local structure and larger values capture broader neighborhood trends.

\paragraph{CBLOF} \citep{cblof} computes anomaly scores by first clustering the data and then measuring each sample's distance to large, dense clusters. The hyperparameter $N_{\text{clusters}}$ controls the clustering granularity, while $\alpha$ and $\beta$ define the criteria for separating large and small clusters, thereby modulating the anomaly score assigned to samples in small clusters.

\paragraph{IForest} \citep{Iforest} isolates anomalies by recursively partitioning the data using random feature selection and random split values. The hyperparameter $N_{\text{estimators}}$ specifies the number of trees in the ensemble, improving stability with larger values, while $\texttt{max\_samples}$ controls the subsample size used to build each tree, affecting both detection sensitivity and computational cost.

\paragraph{EGMM} \citep{EGMM} scores anomalies by averaging the log-likelihoods assigned to each sample across a retained ensemble of Gaussian mixture models, where lower likelihood indicates higher anomaly. Models with poor out-of-bag likelihood are pruned using a \texttt{discard threshold}, and the hyperparameter $k$ controls the number of mixture components in each GMM. Since EGMM is computationally expensive due to repeated GMM fitting, for large datasets in \fraudbench and \ovrbench we cap the context length at 50{,}000 and restrict the number of features to at most 200.

\paragraph{GOAD} \citep{goad} trains a classifier to distinguish among many random linear transformations of normal data—where $N_\text{rots}$ is the number of such transformations, 
$m$ is the margin enforcing separation in the transformation-consistency loss, and $\lambda$ balances that loss against cross-entropy. At test time, samples that fail to produce transformation-discriminative embeddings (i.e., lie far from learned transformation clusters) are scored as anomalous.

\paragraph{ICL}\citep{ICL} learns anomaly-sensitive representations via contrastive learning by distinguishing genuine feature relationships from corrupted (negative) samples, with anomalies yielding higher contrastive loss. Key hyperparameters include the temperature $\tau$, which controls the sharpness of the contrastive objective, and the number of negative samples, which governs discrimination strength. The window size $k$ defines the number of neighboring samples used to form the local context; since varying $k$ can trigger out-of-memory errors on large datasets, we retain the default setting.

\paragraph{DTE-C \& DTE-NP} \citep{LivernocheJHR24} are diffusion-based anomaly detection methods that score anomalies via reconstruction error under a learned diffusion process, with DTE-NP modeling continuous data non-parametrically and DTE-C extending the framework to categorical features via discrete diffusion. In DTE-NP, $k$ controls the neighborhood size used for local density estimation during scoring. DTE-C extends diffusion-based anomaly detection to categorical data, where $N_\text{categories}$
 determines the discrete state space modeled by the diffusion process, and the diffusion timestep controls the noise schedule length.

\paragraph{NPT-AD} \citep{NPTAD} is a deep anomaly detection method for tabular data that leverages Non-Parametric Transformers (NPTs) to jointly model both feature–feature and sample–sample dependencies. In a reconstruction framework, an NPT is trained to reconstruct masked features of normal samples using the entire training set non-parametrically, and at inference the model’s ability (or failure) to reconstruct masked portions yields an anomaly score, with larger reconstruction errors indicating anomalies. Following \citep{NPTAD}, we adopt the same learning schedule and train the model for 2000–4000 steps. For small datasets, we vary $r$, the number of features masked simultaneously, in $\{1,2,3,4\}$, while for large datasets we fix $r=1$ due to computational constraints. Since large $m$ (the number of masking patterns) leads to prohibitive inference cost, we keep $m$ small. To fit memory limits, we subsample training data to at most 50{,}000 instances and 150 features for large datasets, and batch test-time evaluations accordingly.

\paragraph{\fomo} \cite{shen2025fomod} is a foundation model specifically trained of OD tasks. Due to context length and feature dimension limitation, during inference on all benchmarks, we will take up to 5,000 inlier points as contexts and sub-sample 100 dimensions randomly if the dataset has more than 100 dimension.

\paragraph{TabPFN-OD} TabPFN-OD repurposes TabPFN~\citep{hollmann2023tabpfn} for outlier detection using a feature-wise self-prediction scheme. Given normal training data $\mathbf{x}_{\text{train}} \in \mathbb{R}^d$ and test data $\mathbf{x}_{\text{test}}$, we randomly select a subset of features and, for each feature $j$, train a TabPFN regressor to predict $x_j$ from $\mathbf{x}_{-j}$ using only normal samples; anomaly scores are computed as the average absolute prediction error across selected features. This method leverages inter-feature dependencies in normal data, which anomalies tend to violate. For large datasets we sample up to 100 features and fix the context length to 5000.

\paragraph{\outform} \citep{ding2026outformer} is a recent foundation model that improves over \fomo, by training with mixed priors and self-evolving curriculum learning. Following the paper, we use up to 1,000 inlier points and sub-sample 100 dimensions randomly, ensembling across 50 forward runs.

\begin{table*}[h!]
\centering
\caption{Overview of benchmarked OD baselines on \bench, listed chronologically, which are selected from Top-2 methods across various survey and benchmarking studies as cited in column 5.}
\vspace{-0.1in}
\begin{tabular}{l l l l l l l}
\hline
\textbf{Method} & \textbf{Abbrev.} & \textbf{Year} & \textbf{Type} & \textbf{Top-2 in:} &  \textbf{Code} & \textbf{Usage} \\
\hline
One-class Support Vector Machine \citep{OCSVM} & OCSVM & 1999 & \textcolor{darkgreen}{Shallow}  & \citep{domingues2018comparative} & PyOD & One Class \\
K-th Nearest Neighbors \citep{knn} & KNN & 2000 & \textcolor{darkgreen}{Shallow}  & \citep{campos2016evaluation,goldstein2016comparative,shen2025fomod,NPTAD} & PyOD & Un- \& One Class \\
Local Outlier Factor \citep{LOF} & LOF & 2000 & \textcolor{darkgreen}{Shallow}  & \citep{emmott2015meta,campos2016evaluation,goldstein2016comparative} & PyOD & Un- \& One Class \\
Clustering Based Local Outlier Factor \citep{cblof} & CBLOF & 2003 & \textcolor{darkgreen}{Shallow}  & \citep{han2022adbench} & PyOD & Un- \& One Class \\
Isolation Forest \citep{Iforest} & IForest & 2008 & \textcolor{darkgreen}{Shallow}  & \citep{emmott2013systematic,emmott2015meta,domingues2018comparative,han2022adbench} & PyOD & Un- \& One Class \\
Ensemble Gaussian Mixture Model \citep{EGMM} & EGMM & 2013 & \textcolor{darkgreen}{Shallow}  & \citep{emmott2013systematic} & Ours & Un- \& One Class \\
Classification Based AD using General Data \citep{goad} & GOAD & 2020 & \textcolor{darkblue}{Deep}  & \citep{ICL} & Paper & Un- \& One Class \\
Internal Contrastive Learning \citep{ICL} & ICL & 2022 & \textcolor{darkblue}{Deep}  & \citep{shen2025fomod,ICL} & Paper & One Class \\
Prior-data Fitted Network \citep{hollmann2023tabpfn} & TabPFN-OD & 2022 & \textcolor{darkred}{Foundation}  & - & Paper+Ours & One Class \\
Non-Parametric Transformers for AD  \citep{NPTAD} & NPT-AD & 2023 & \textcolor{darkblue}{Deep}  & \citep{NPTAD} & Paper & One Class \\
Diffusion Time Estimation (Non Parametric) \citep{LivernocheJHR24} & DTE-NP & 2024 & \textcolor{darkgreen}{Shallow}  & \citep{LivernocheJHR24,shen2025fomod} & Paper & Un- \& One Class \\
Diffusion Time Estimation (Categorical) \citep{LivernocheJHR24} & DTE-C & 2024 & \textcolor{darkblue}{Deep}  & \citep{LivernocheJHR24} & Paper & One Class \\
Zero-shot Tabular Outlier Detection \citep{shen2025fomod} & \fomo & 2025 & \textcolor{darkred}{Foundation}  & \citep{shen2025fomod} & Paper & One Class \\
Mixed Prior Tabular Outlier Detection \citep{ding2026outformer} & \outform & 2026 & \textcolor{darkred}{Foundation}  & \citep{ding2026outformer} & Paper & One Class \\
\hline
\end{tabular}
\label{tab:baselines}
\end{table*}

\begin{table*}[htbp]
\caption{Hyperparameter configurations for the baselines. }
\vspace{-0.15in}
\scalebox{0.95}{
\begin{tabular}{ccc}
\toprule
\textbf{Model} & \textbf{List of Possible HPs} & \textbf{\#Models} \\
\midrule
OCSVM & \makecell[c]{kernel: [\textbf{rbf}, linear]\\
$\nu$: [0.05, 0.2, \textbf{0.5}, 0.8]\\
$\gamma$: [\textbf{scale}, 0.1, 1.0] (rbf only)} & 16 \\  \hline
KNN & $k$: [\textbf{5}, 10, 20, 50, 100] & 5 \\ \hline
LOF & 
$N_\text{neighbors}$: [10, \textbf{20}, 50, 100]
& 4 \\ \hline
CBLOF &
\makecell[c]{
$N_\text{clusters}$: [4, \textbf{8}, 16, 32]\\
$\alpha$ : [1, 0.5, \textbf{0.9}, 1] \\
$\beta$ : [2, \textbf{5}, 10]}
& 48 \\ \hline 
IForest & 
\makecell[c]{
$N_\text{estimators}$: [50, \textbf{100}, 200] \\
max\_samples: [64, 128, \textbf{256}] }
& 9 \\\hline
EGMM & \makecell[c]{
k : [(1,2,3,4,5), \textbf{(6,7,8,9,10)}] \\
$N_\text{bootstrap}$ [5, \textbf{15}] \\
Discard threshold: [\textbf{0.85}]} 
& 4 \\ \hline
GOAD & \makecell[c]{
$m$ : [0.5, \textbf{1}], $\lambda$:[\textbf{0.1}, 0.5] \\
$N_\text{rots}$: [128, \textbf{256}, 512] \\
Embedding dimension: [\textbf{32}, 64] \\
MLP layers: [\textbf{2}, 3], Leaky ReLU: [\textbf{0.1}, 0.2] \\ 
Epochs: [\textbf{1}, 5, 10],
Batch size: 64, LR: 0.001, Optimizer: Adam \\}
& 288 \\ \hline
ICL & \makecell[c]{Window\_size
$k\in \{2,10,  d - 150 \}$ (dimension-based) \\ Embed\_dim: [\textbf{200}, 100] \\
Loss\_temperature $\tau$: [\textbf{0.1}, 0.5] \\
F/G-network layers: [2,\textbf{3}], BatchNorm \\
Num\_negatives: [100, \textbf{1000}] \\
Num\_epochs: [500, \textbf{2000}], Lr = [\textbf{1e-3}, 1e-4] \\
Activation: LeakyReLU [0.1, \textbf{0.2}], Optimizer:Adam}
&  128  \\ \hline
NPT-AD & \makecell[c]{$r$: [1, 2, 3, 4] (dataset specific) \\ 
Embed\_dim $e$: [16, 32] \\
m: [1, 2, 4],$p_\text{mask}$ = [0.15, 0.25] \\
Num\_transformer blocks: [2, 4] \\
Num\_steps: [2000, 4000]\\
Batch\_size:32, LR:1e-3 \\ Optimizer: LAMB(0.9, 0.999) with Lookahead} & 192 \\ \hline
DTE-NP & $k$: [\textbf{5}, 10, 20, 50, 100] & 5 \\  \hline
DTE-C &  \makecell[c]{$N_\text{categories}$: [5, \textbf{7}, 10] \\ Diffusion\_timestep: [100, \textbf{300}, 500] \\
Hidden\_layers: [\textbf{\{256,512,256\}},  \{256,256\} \} \\
Dropout: [0.3, \textbf{0.5}] \\
Activation: ReLU \\
Epochs: [350, \textbf{400}], Lr: 1e-4} &  72 \\
\bottomrule
\end{tabular}
}
\label{tab:hps}
\vspace{-0.15in}
\end{table*}

\subsubsection{Training Details}
For shallow, deep, and foundation models, all experiments are conducted on six NVIDIA A6000 GPUs with AMD EPYC 7742 64-core processors. For shallow methods, we exhaustively evaluate all hyperparameter (HP) configurations and report performance for each setting. In contrast, deep models typically involve much larger HP spaces and substantially longer training times; therefore, we randomly sample five HP configurations from Table~\ref{tab:hps} to evaluate their performance.

\subsection{Performance Metrics and Statistical Tests}
\label{assec:metrics}

\subsubsection{Ranking Based Metrics}
For ranking and group based metrics, we adopt 5 metrics following the previous FM literature \cite{zhang2025mitra} and give a brief introduction of the metrics.

\paragraph{Average Rank.}
Let $\mathcal{M}$ be the set of models and $\mathcal{D}$ the set of evaluation datasets. 
For each model $m \in \mathcal{M}$ and dataset $\delta \in \mathcal{D}$, let $\mathrm{rank}_{\delta}(m)$ denote the performance rank of $m$ on $\delta$, with smaller values indicating better performance. 
The average rank of model $m$ is defined as
\begin{equation}
\mathrm{AvgRank}(m)
\;=\;
\frac{1}{|\mathcal{D}|}
\sum_{\delta \in \mathcal{D}}
\mathrm{rank}_{\delta}(m).
\end{equation}

\paragraph{Elo Rating.}
Elo rating aggregates pairwise win-loss outcomes into a global ranking by treating each model as a player~\citep{elo1967uscf}. 
For each dataset, models are compared in a round-robin manner, with match outcomes determined by relative performance and ties allowed within a tolerance. 
Ratings are updated using the standard Elo scheme with initial rating $R_0=1000$, K-factor $K=32$, and tie threshold $\epsilon=0.5$, yielding final scores that reflect a model’s relative strength across all datasets.

\paragraph{Winrate.}
Winrate quantifies the fraction of pairwise comparisons in which a model outperforms others across datasets. 
For a model $m \in \mathcal{M}$, it is defined as
\begin{align}
\mathrm{Winrate}(m)
&=
\frac{1}{|\mathcal{D}| (|\mathcal{M}| - 1)}
\sum_{\delta \in \mathcal{D}}
\sum_{\substack{m' \in \mathcal{M} \\ m' \neq m}} \\ \nonumber
&\Big(
\mathbb{I}[E_{\delta}(m) < E_{\delta}(m')]
+
\frac{1}{2}\,\mathbb{I}[E_{\delta}(m) = E_{\delta}(m')]
\Big),
\end{align}
where $E$ denotes an error metric (e.g., $E=1-\mathrm{AUROC}$), $\mathbb{I}[\cdot]$ is the indicator function, and ties contribute half a win. 
Although defined using AUROC, this metric applies to other performance measures such as AUPRC.

\paragraph{Rescaled AUC (rAUC)}
rAUC mitigates differences in dataset difficulty by normalization of model errors within each dataset relative to the best and worst performers. 
For dataset $\delta$, the rescaled AUC of model $m$ is defined as
\begin{equation}
\mathrm{rAUC}_{\delta}(m)
=
1 - \frac{E_{\delta}(m) - \min_{m} E_{\delta}(m)}
{\max_{m} E_{\delta}(m) - \min_{m} E_{\delta}(m)}.
\end{equation}
The  rAUC is calculated by averaging $\mathrm{rAUC}_{\delta}(m)$ across all datasets.

\paragraph{Champion Delta}
Let $m^{*}=\arg\min_{m} E(m)$ denote the best-performing (champion) model.
The Champion Delta of model $m$ is defined as
\begin{equation}
\mathrm{C\Delta}(m)
=
\left(1-\frac{E(m^{*})}{E(m)}\right)\times 100,
\end{equation}
measuring the relative percentage performance gap between model $m$ and the champion.

\subsubsection{Statistical Tests}

\paragraph{Permutation Test.}
Given two models evaluated on the same datasets, we compute per-dataset performance differences $d_i = s_i^{(A)} - s_i^{(B)}$ and discard ties.
The test statistic is $T=\sum_i d_i$, and under the null hypothesis of equal performance, randomly flipping the signs of $d_i$ yields a reference distribution
\[
T^{(b)}=\sum_i \epsilon_i d_i,\quad \epsilon_i\in\{-1,+1\}.
\]
The one-sided $p$-value is estimated as $p=\Pr(T^{(b)}\ge T)$, quantifying the probability that model~A outperforms model~B by chance; unlike a sign test, this permutation test accounts for the magnitude of performance differences. Compared to the Wilcoxon signed-rank test, this permutation test makes fewer distributional assumptions and directly models the null hypothesis via random sign flips of observed differences. It is particularly well-suited for benchmark comparisons with heterogeneous effect magnitudes, where Wilcoxon’s ranking can discard meaningful scale information.

A value $\geq 0.95$ indicates that the column model wins significantly, while a value $\leq 0.05$ indicates that the row model wins significantly; values in between imply no statistically significant difference. In general, smaller values provide stronger evidence in favor of the row model, whereas values close to $0.5$ indicate essentially no difference between the two models.

\section{Additional Results}
\label{asec:moreresults}

\begin{figure*}[!h]
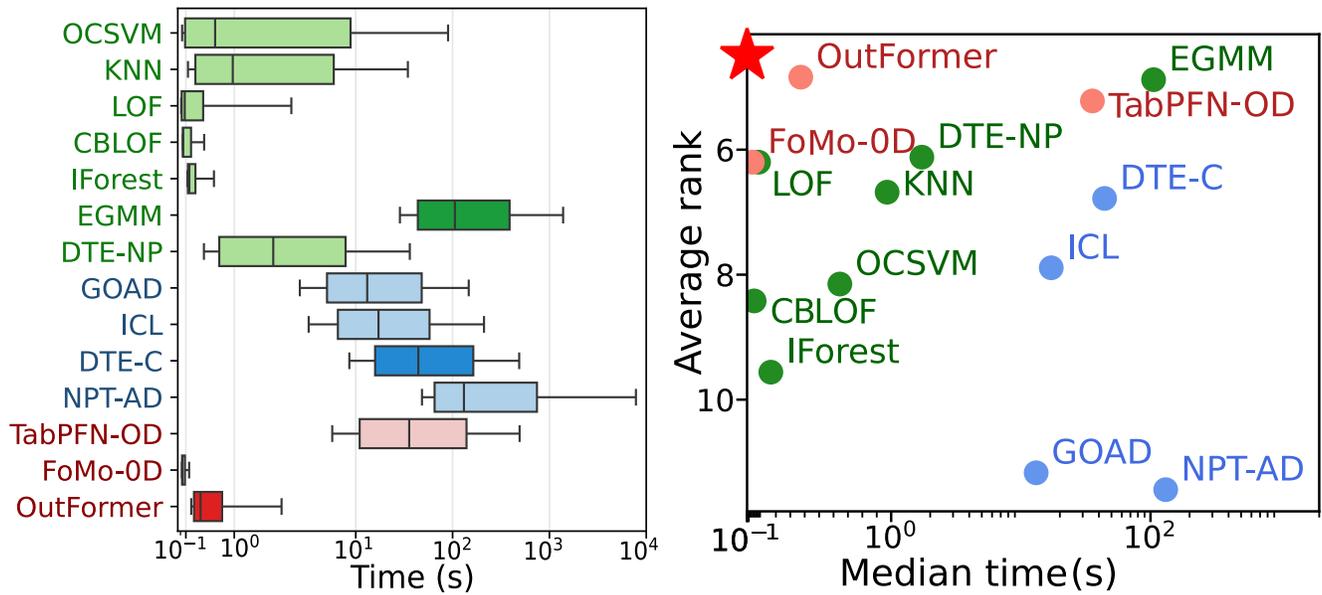

\vspace{0.05in}
    \centering
   \begin{minipage}[t]{0.49\linewidth}
        \centering
        \includegraphics[width=\linewidth]{time_boxplot_macrodata-2.pdf}
    \end{minipage}
    \begin{minipage}[t]{0.5\linewidth}
        \centering
        \includegraphics[width=\linewidth]{time_avg_rank-2.pdf}
    \end{minipage}
   
    \caption{(left) Running time comparison of OD methods across all datasets. Feed-forward-only foundation models (FMs) incur as low latency as certain shallow models such as iForest, while achieving significantly better performance. Deep models take considerably longer, while competitive shallow models EGMM, DTE-NP and kNN are also slow relative to FMs. (right) Median total time (fitting+inference)  vs. Avg rank w.r.t. AUPRC across all 2,446 datasets combined. Foundation models \fomo and \outform occupy the Pareto front, offering the best overall performance-time trade-off. EGMM and DTE-NP achieve competitive performance while taking considerable running time. }
    \label{fig:total_time_appx}
    
 \end{figure*}

We provide additional results on individual benchmark's AUPRC and AUROC performances, with time analysis shown in Figure \ref{fig:total_time_appx}.

\subsection{AUPRC Results on Individual Benchmarks}
\label{assec:indivpr}
\begin{figure*}[!htbp]
    \centering
    \includegraphics[width=0.7 \linewidth]{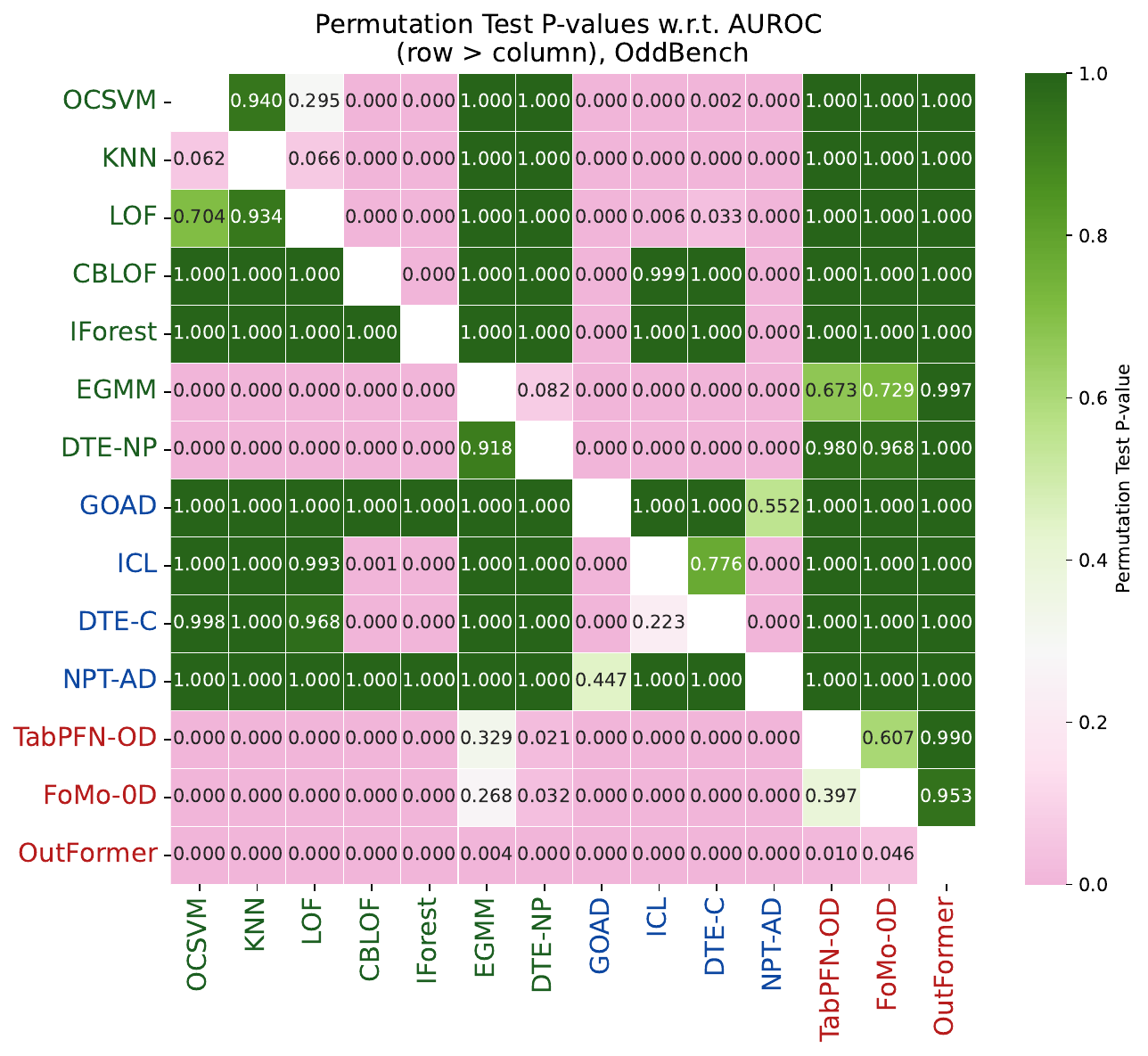}
    \vspace{-0.2in}
    \caption{Paired permutation test results w.r.t. AUPRC across 690 public \fraudbench datasets. All foundation models achieve strong performance (last three rows), with \outform outperforming all baselines significantly ($p$$\leq$$0.05$). EGMM and DTE-NP emerge as the best-performing shallow methods, while deep models do not achieve competitive results, mainly due to hyperparameter sensitivity. }
    \label{fig:permutation_oddbench_aupr}
\end{figure*}

Table \ref{tab:oddbench_results_auprc} and Figure \ref{fig:permutation_oddbench_aupr} show the performance comparison and pairwise permutation test results w.r.t. AUPRC on \fraudbench. The results underscore the effectiveness of foundation models for OD (see the last three rows): \outform outperforms all baselines with all $p\leq 0.05$. \tabod and \fomo follow with similar results, outperforming all shallow and deep baselines, except EGMM. Notably, deep models do not achieve competitive performance mainly due to their hyperparameter (HP) sensitivity as we present in the following (Recall that results are based on average performance across HP configurations). EGMM and DTE-NP stand out as competitive shallow (classical) methods.

\begin{figure*}[!htbp]
    \centering
    \includegraphics[width=0.7\linewidth]{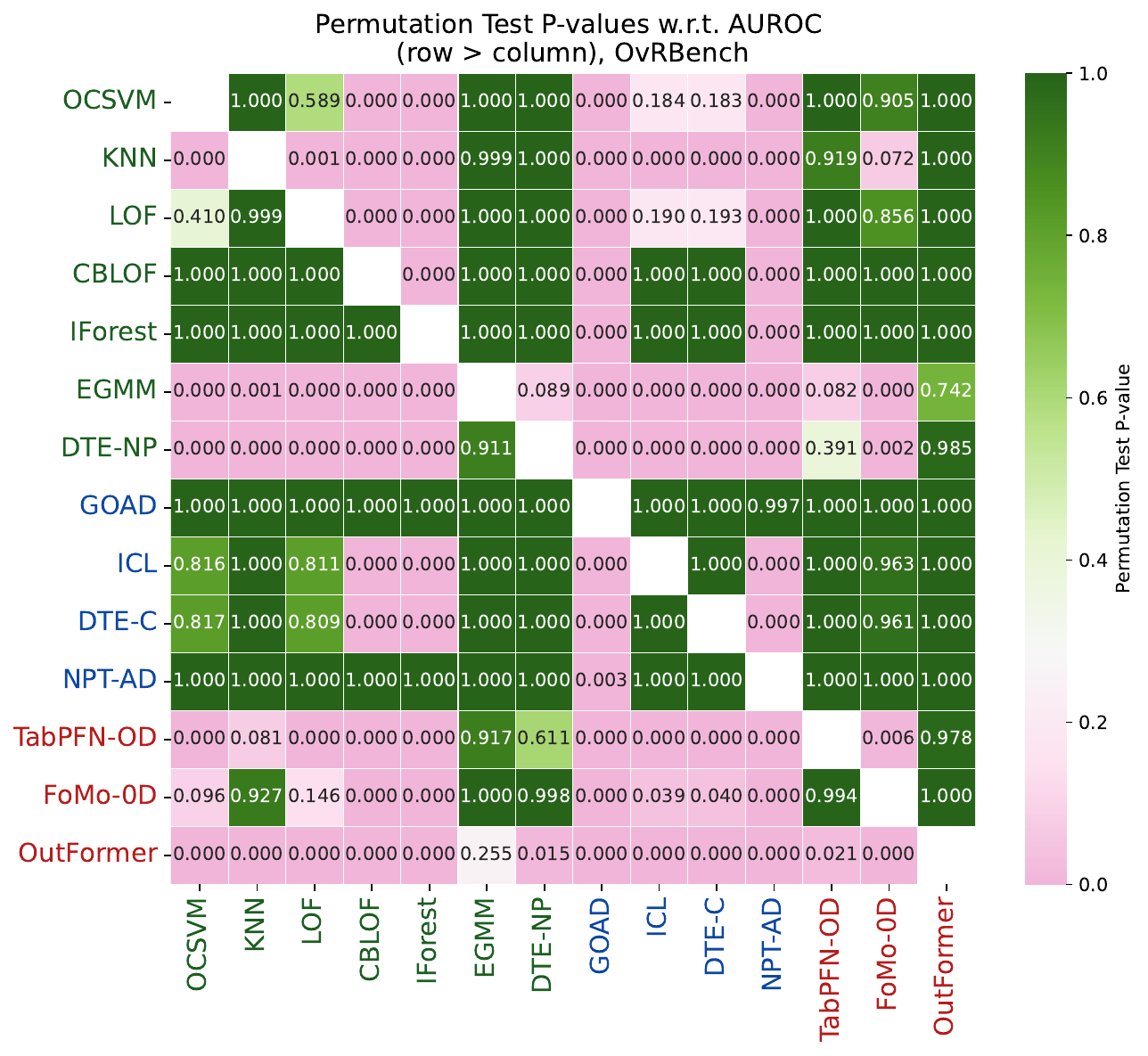}
    \vspace{-0.2in}
    \caption{Paired permutation test results w.r.t. AUPRC across 756 public \ovrbench datasets. All foundation models achieve strong performance (last three rows), with \outform outperforming all baselines  significantly ($p$$\leq$$0.05$), except EGMM ($p=0.255$).  Deep models do not achieve competitive results, mainly due to hyperparameter sensitivity. }
    \label{fig:permutation_ovrbench_aupr}
\end{figure*}

Table \ref{tab:ovrbench_results_auprc} and Figure \ref{fig:permutation_ovrbench_aupr} show the performance comparison and pairwise permutation test results on \onevsrestbench. 
EGMM followed by DTE-NP stand out as two competitive classical methods, and outperform \fomo significantly. All foundation models are competitive overall, with \outform outperforming all baselines ($p\leq0.05$) except EGMM ($p=0.255$). These results on our two large-scale real-world benchmarks \oddbench and \ovrbench highlight the prowess of foundation models for OD.

\begin{figure*}[!htbp]
    \centering
    \includegraphics[width=0.7 \linewidth]{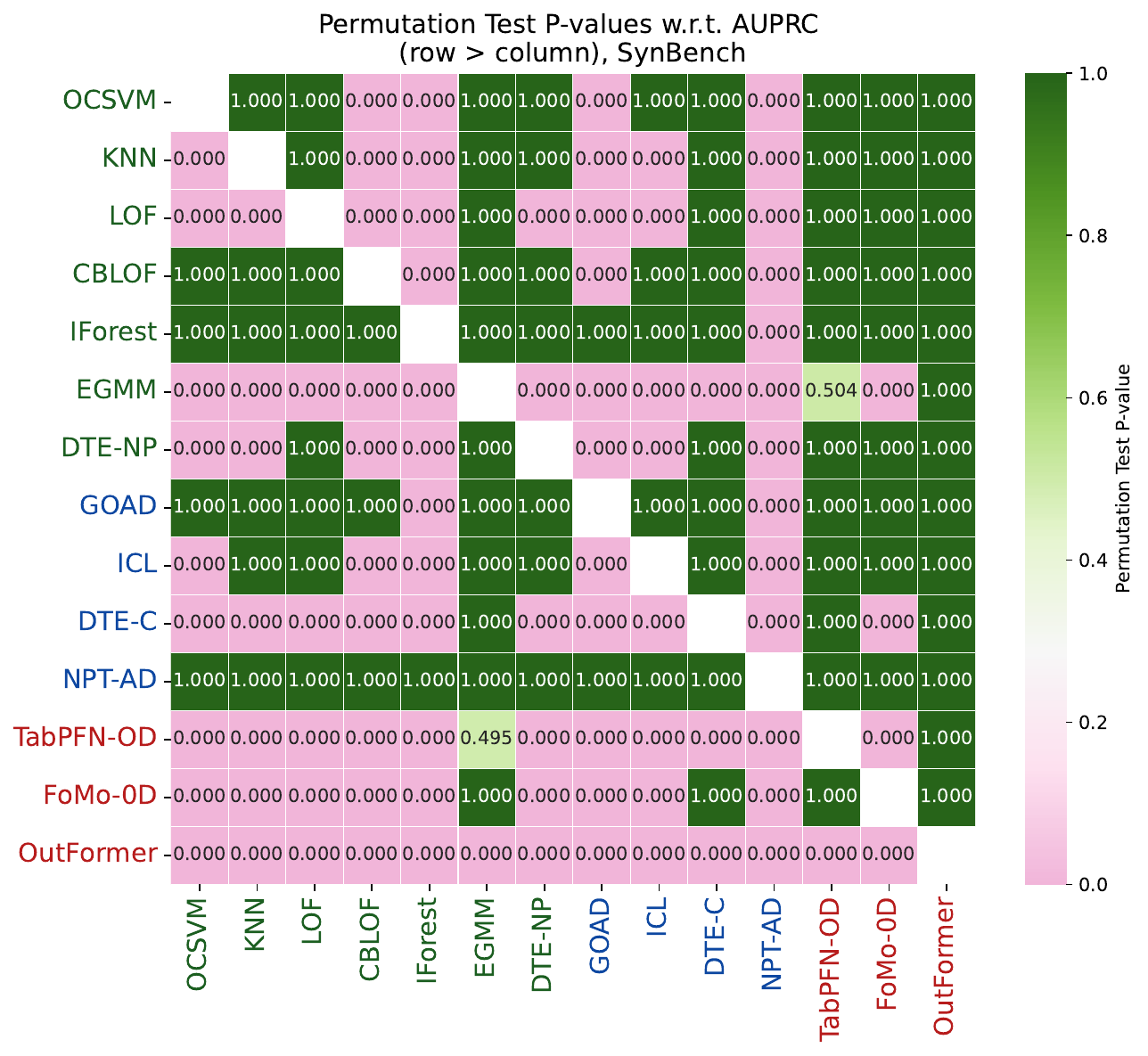}
    \vspace{-0.2in}
    \caption{Paired permutation test results w.r.t. AUPRC across 800 public \synbench datasets. All foundation models achieve competitive performance (last three rows), with \outform outperforming all baselines  significantly ($p$$\leq$$0.05$). 
    }
\label{fig:permutation_synbench_aupr}
\end{figure*}

Finally, Table \ref{tab:synbench_results_auprc} and Figure \ref{fig:permutation_synbench_aupr} show the results on our simulated datasets in \synbench, with similar findings.

\subsection{AUROC Results Combined and on Individual Benchmarks}
\label{assec:roc}

\begin{figure*}[!htbp]
    \centering
    \includegraphics[width=0.7 \linewidth]{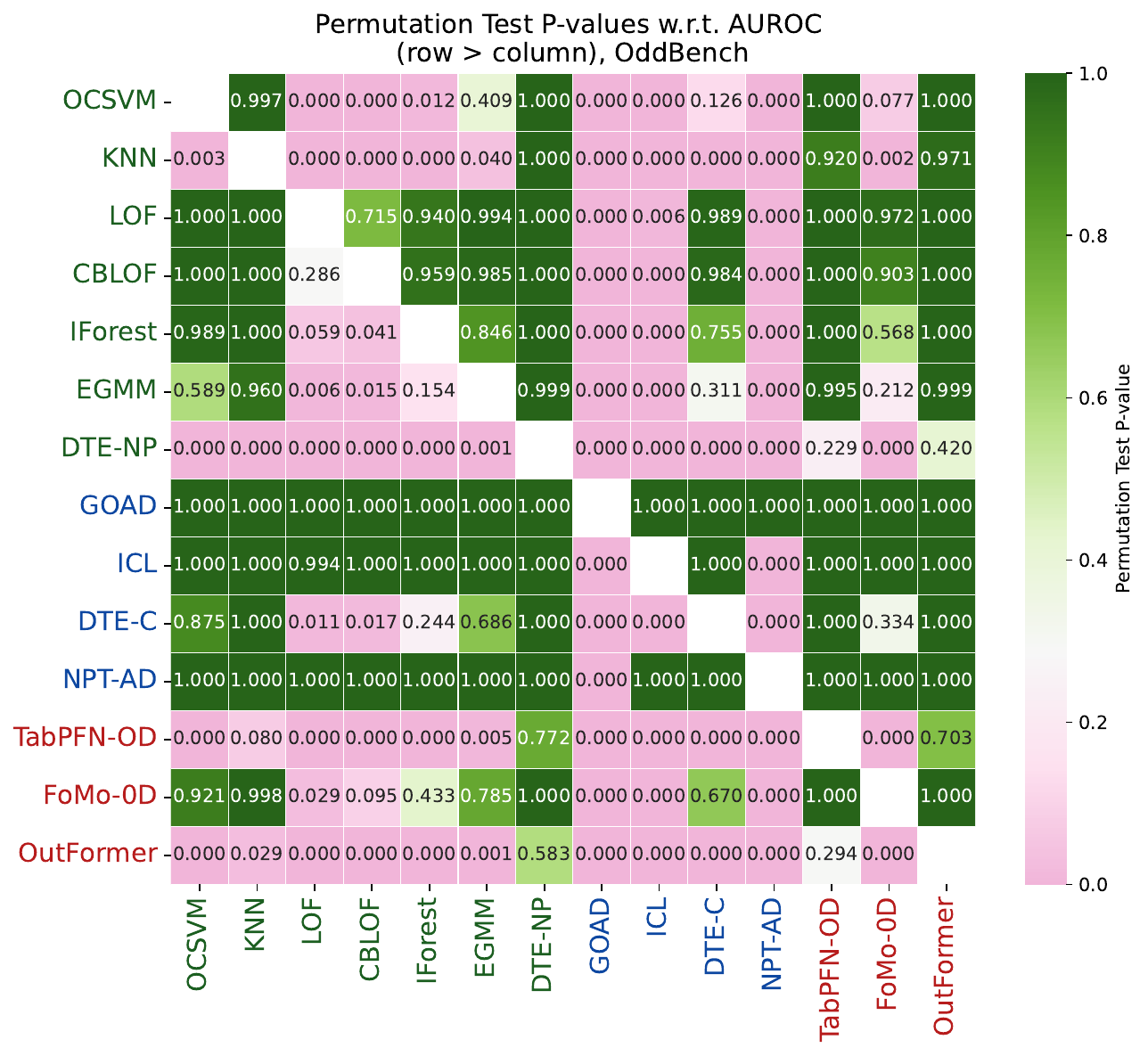}
    \vspace{-0.2in}
    \caption{Paired permutation test results w.r.t. AUROC across 690 public \fraudbench datasets.  
     Among shallow methods, DTE-NP emerges as the strongest performer, followed by EGMM and KNN. Deep models generally underperform, mainly due to their hyperparameter sensitivity. Foundation models \outform and \tabod achieve strong performance, both outperforming all other baselines significantly ($p$$\leq$$0.05$), except DTE-NP ($p=0.583$ and $p=0.772$, respectively).}
    \label{fig:permutation_oddbench_auroc}
\end{figure*}

Table~\ref{tab:oddbench_results} and Figure~\ref{fig:permutation_oddbench_auroc} present the AUROC performance and permutation test results on \fraudbench. 
\outform and DTE-NP achieve the strongest
overall performance, with foundation models generally offering the leading edge.
Among shallow methods, DTE-NP
emerges as the strongest performer, followed by EGMM and KNN. Deep models generally underperform, mainly due to their
hyperparameter sensitivity. Foundation models \outform and \tabod achieve strong performance, both outperforming
all other baselines significantly $(p<0.05)$, except DTE-NP ($p=0.583$ against \outform, and $p= 0.772$ against \tabod).

\begin{figure*}[!htbp]
    \centering
    \includegraphics[width=0.7 \linewidth]{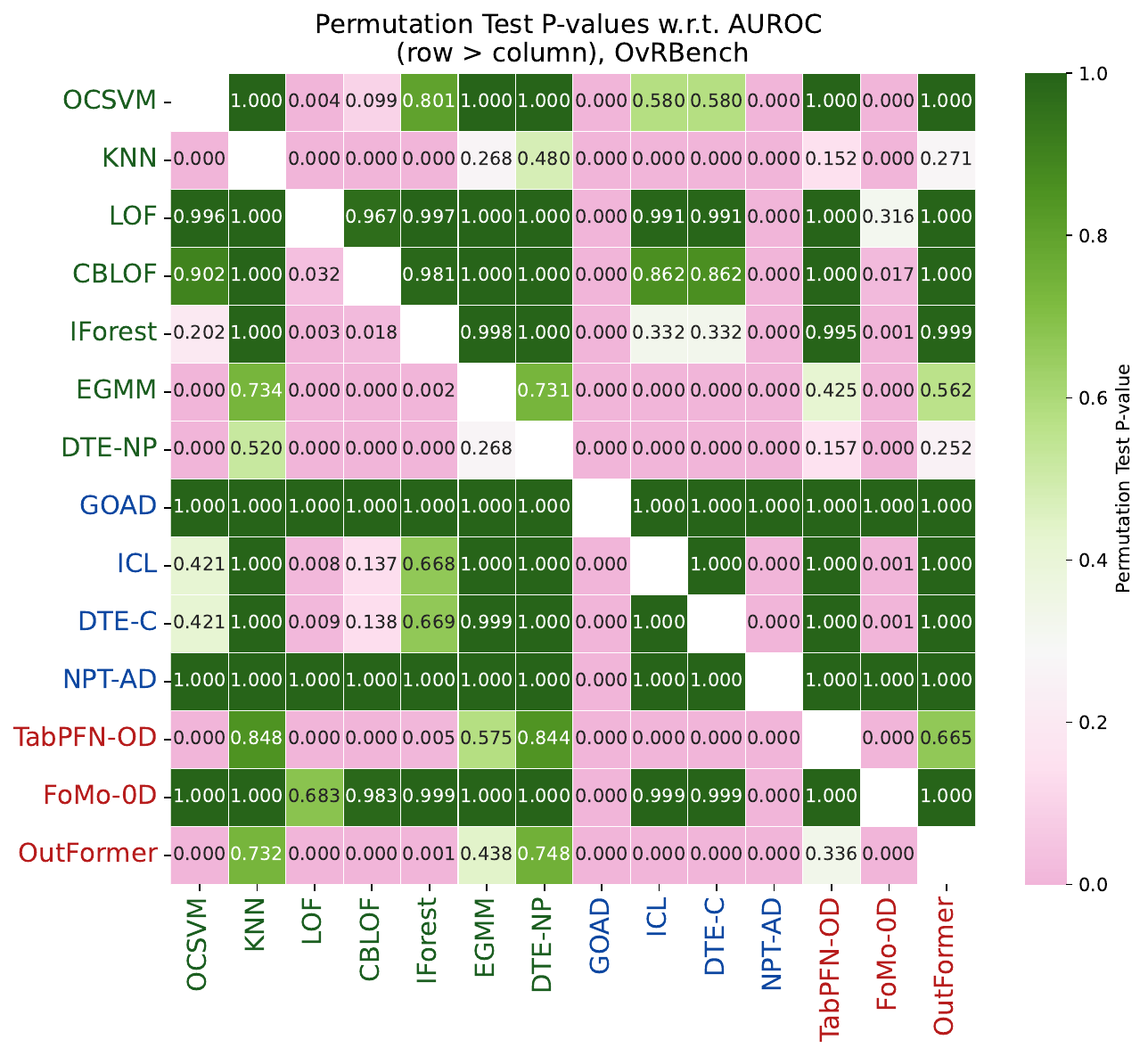}
    \vspace{-0.2in}
    \caption{Paired permutation test results w.r.t. AUROC across 756 public \ovrbench datasets. 
Among shallow models, EGMM, DTE-NP and KNN achieve strong and comparable performance, which are also statistically similar to two  foundation models,  \outform and \tabod. }
    \label{fig:permutation_ovrbench_auroc}
\end{figure*}

Table~\ref{tab:ovrbench_results} and Figure~\ref{fig:permutation_ovrbench_auroc} report the corresponding AUROC results on \onevsrestbench, where the shallow methods KNN, EGMM, and DTE-NP,
as well as the foundation models \outform and \tabod achieve the strongest overall results, significantly outperforming
the rest of the baselines while being comparable among themselves.

\begin{figure*}[!htbp]
    \centering
    \includegraphics[width=0.7 \linewidth]{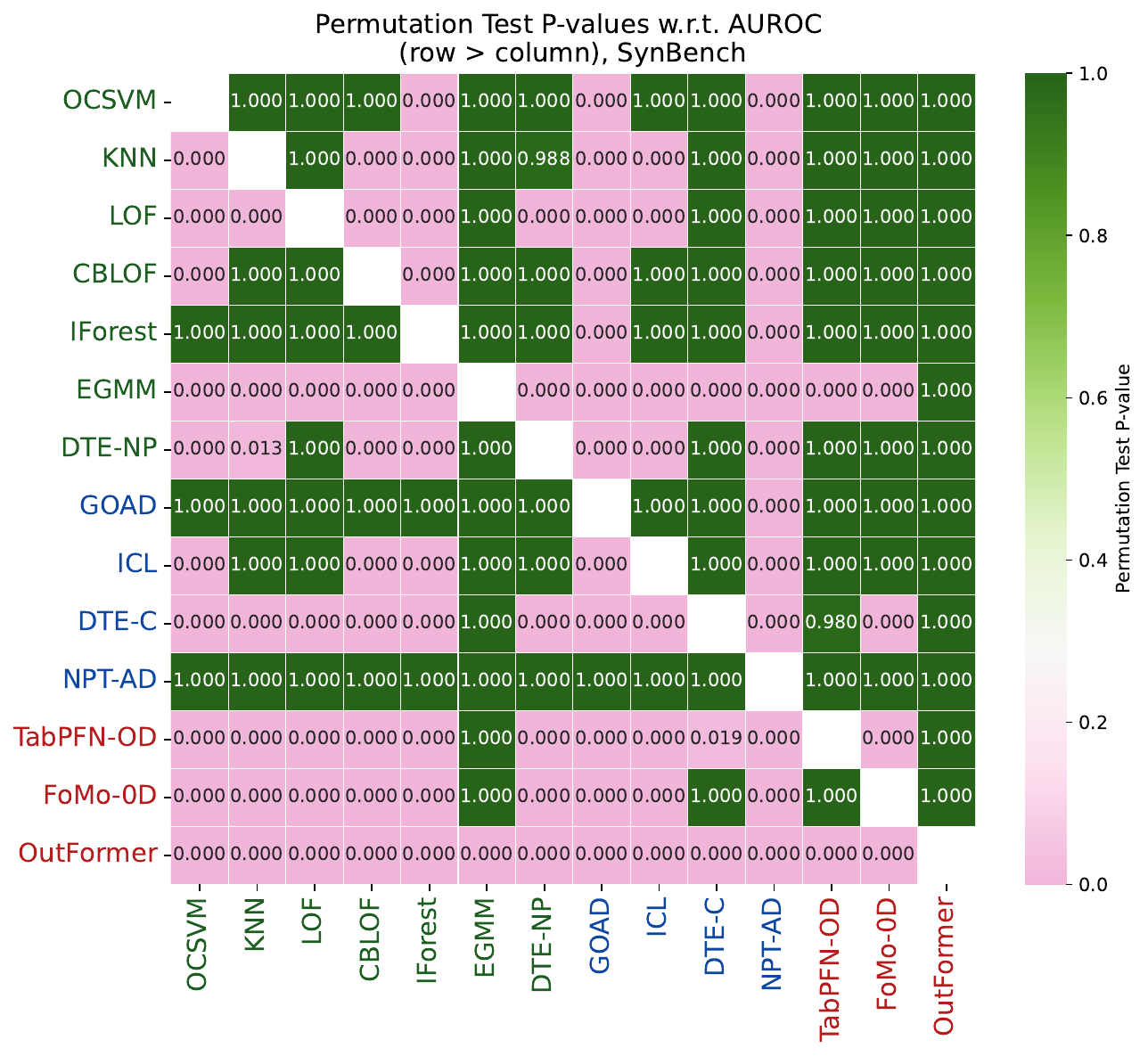}
    \vspace{-0.2in}
    \caption{Paired permutation test results w.r.t. AUROC across 800 public \synbench datasets. EGMM is the strongest performer among shallow and deep methods. Foundation models achieve competitive performance (last three rows), with \outform outperforming all baselines  significantly ($p$$\leq$$0.05$).}
    \label{fig:permutation_synbench_auroc}
\end{figure*}

Table~\ref{tab:synbench_results} and Figure~\ref{fig:permutation_synbench_auroc} report AUROC performance comparison on \synbench. \outform, \tabod and EGMM achieve 
strong results, significantly outperforming the rest of the baselines, where \outform
outperforms all baselines significantly $(p<0.05)$.

\begin{figure*}[!htbp]
    \centering
    \includegraphics[width=0.7 \linewidth]{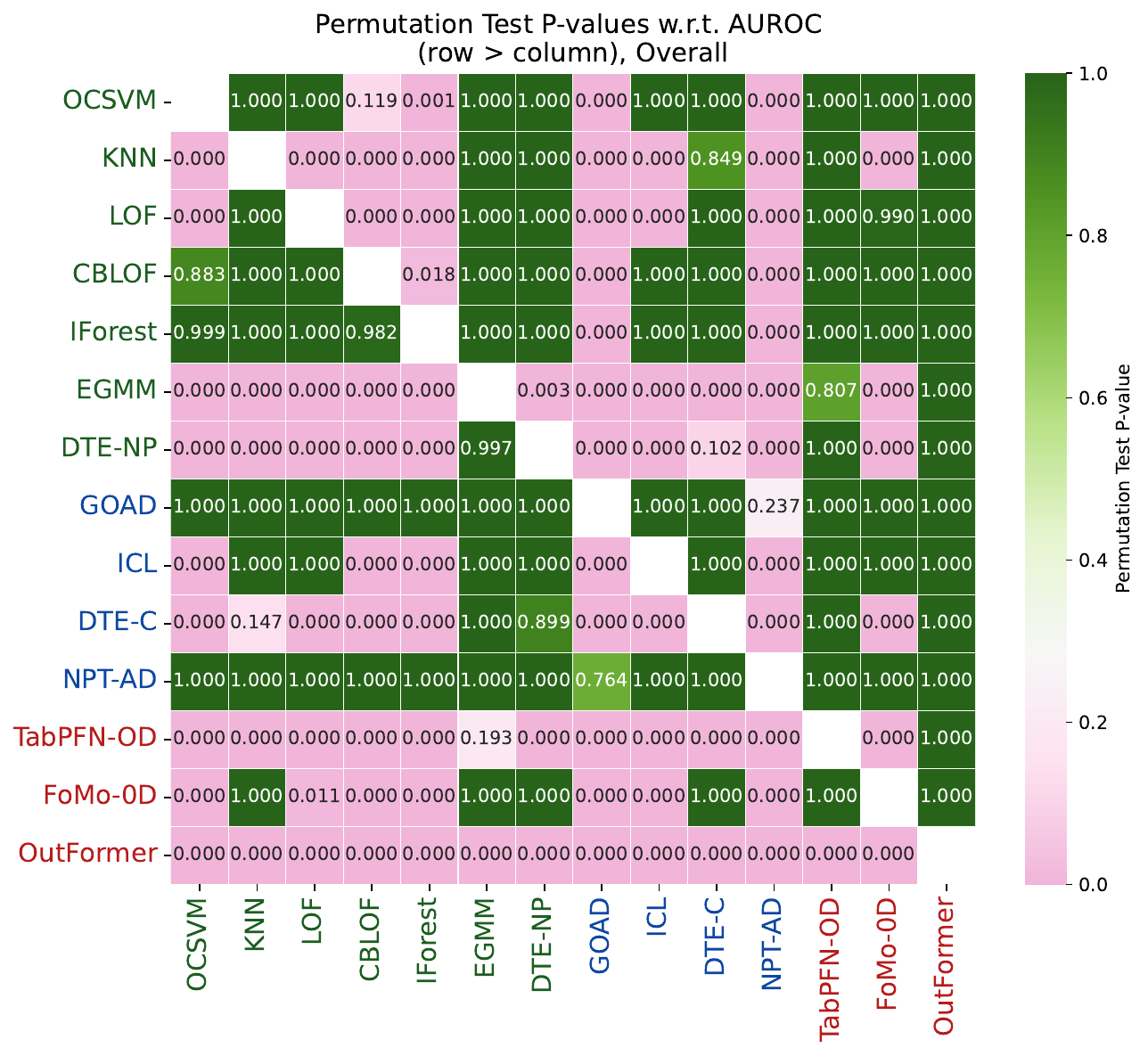}
    \vspace{-0.2in}
    \caption{Paired permutation test results w.r.t. AUROC across overall 2,446 datasets. Foundation models (last three rows) demonstrate  competitive performance, as well as classical methods EGMM and DTE-NP while the latter being much slower. \outform significantly outperforms all baselines ($p$$\leq$$0.05$).}
    \label{fig:permutation_overall_auroc}
\end{figure*}

Table~\ref{tab:overall_results_auroc} and Figure~\ref{fig:permutation_overall_auroc} summarize AUROC performance across all 2,446 datasets. \outform and \tabod achieve strong results, where \outform outperforms all baselines significantly ($p<0.05$) - establishing FM as the dominant approach to OD.

\subsection{Analysis on Results}
\label{assec:additional_analysis}
 Real-world datasets contain heterogeneous anomaly types and unknown generative mechanisms, making it difficult to precisely attribute which models are best suited for which anomaly types. To address this, we complement our analysis with controlled studies where such factors can be better isolated. We conduct preliminary analyses on: dataset size and dimensionality, performance differences across OddBench vs OvRBench, and more detailed diagnostic analysis on SynBench, where the data-generating process is known.

\paragraph{Effect of Dimensionality.}
We investigate whether dimensionality affects the relative performance of shallow and deep outlier detection methods on OddBench. Specifically, we split the datasets into high-dimensional datasets ($d > 100$) and low-dimensional datasets ($d < 100$), and compare the average ranks of shallow and deep methods, where lower rank indicates better performance. As shown in Table~\ref{tab:dim_effect}, shallow methods outperform deep models in both dimensional regimes. However, the performance gap is smaller in high-dimensional datasets than in low-dimensional datasets ($1.70$ vs. $2.61$), suggesting that deep models become relatively more competitive in high-dimensional regimes, where shallow methods may be more affected by the curse of dimensionality.

\begin{table}[t]
\centering
\caption{Effect of dimensionality on the relative performance of shallow and deep methods on OddBench.}
\label{tab:dim_effect}
\begin{tabular}{lcc}
\toprule
Dimension & Avg. Rank (Shallow) & Avg. Rank (Deep)\\
\midrule
$d > 100$ & 7.04 & \textbf{8.74} \\
$d < 100$ & \textbf{6.85} & 9.46 \\
\bottomrule
\end{tabular}
\end{table}

\paragraph{Foundation Models with Small vs. Large Contexts.}
We further analyze the behavior of foundation models by splitting OddBench datasets according to dimensionality and context size, considering high-dimensional datasets ($d > 100$) versus low-dimensional datasets ($d < 100$), and large-context datasets ($N > 50{,}000$) versus small-context datasets ($N < 50{,}000$). Table~\ref{tab:fm_context_dim} reports the average rank of the top-two foundation models, where lower rank indicates better performance. TabPFN-OD outperforms OutFormer in both high-dimensional and large-context regimes, while OutFormer performs better in low-dimensional and small-context settings. This is likely because TabPFN-OD builds on TabPFN-v2.5, whose pretraining covers substantially larger contexts and feature dimensions, whereas OutFormer is pretrained with a smaller context and dimensionality budget. These results suggest that context selection and scaling remain critical factors for improving foundation model performance in outlier detection.

\begin{table}[t]
\centering
\caption{Average ranks of foundation models on OddBench under different dimensionality and context-size regimes.}
\label{tab:fm_context_dim}
\begin{tabular}{lcc}
\toprule
Dataset & OutFormer & TabPFN-OD \\
\midrule
$d > 100$ & 6.75 & \textbf{5.25} \\
$d < 100$ & \textbf{5.81} & 6.27 \\
$N > 50{,}000$ & 6.10 & \textbf{5.50} \\
$N < 50{,}000$ & \textbf{5.80} & 6.31 \\
\bottomrule
\end{tabular}
\end{table}

\paragraph{OddBench vs. OvRBench.}
We compare model performance across OddBench and OvRBench using AUPRC, following Tables~11--12. We observe that performance improves for nearly all models on OvRBench, for example OutFormer improves from $0.739$ to $0.821$, and OCSVM improves from $0.694$ to $0.791$. This suggests that OvRBench is generally easier and more separable than OddBench. Moreover, shallow methods benefit more significantly than deep and foundation models; for example, KNN's Elo score increases from $861$ to $1254$. This indicates that OvRBench aligns better with classical outlier detection assumptions, such as density, clustering, and smooth distributional structure, whereas OddBench contains more heterogeneous anomaly types that require stronger priors and better generalization, thereby favoring foundation models.

\paragraph{Evaluating SynBench Priors.}
Finally, we evaluate representative models across different SynBench subsets to understand how model performance varies under different synthetic data-generating priors. Table~\ref{tab:synbench_priors} reports the average rank of each method, where lower rank indicates better performance. TabPFN-OD excels on Copula/SCM due to its TabPFN-v2.5 backbone trained on diverse dependencies, making these settings effectively in-distribution. EGMM is competitive across all subsets, as it models data via Gaussian mixtures and detects anomalies as low-likelihood regions. Since SynBench anomalies predominantly manifest as density deviations, EGMM remains effective without modeling the underlying generative mechanisms. DTE-C performs relatively well on Copula and Gaussian data, as its diffusion-based assumption (data lies on a smooth manifold) aligns with these settings, where data forms continuous structures.

Overall, model performance strongly depends on alignment between dataset characteristics and model assumptions. When aligned, even simple models perform well. More importantly, our benchmarks provide multiple data tags and transparent synthetic data generation processes, enabling a systematic analysis for its users.

\begin{table}[t]
\centering
\caption{Average ranks of models across SynBench subsets. }
\label{tab:synbench_priors}
\resizebox{1.0\linewidth}{!}{
\begin{tabular}{lcccc}
\toprule
Types & TabPFN-OD & OutFormer & EGMM & DTE-C \\
\midrule
Copula-Dependence & \textbf{1.57} & 2.52 & 2.48 & 3.61 \\
Copula-Prob. & \textbf{1.39} & 1.88 & 2.79 & 4.01 \\
SCM-Measurement & 3.26 & 4.35 & \textbf{2.96} & 5.14 \\
SCM-Structural & \textbf{3.10} & 6.76 & 3.40 & 7.66 \\
Gaussian & 8.40 & \textbf{1.34} & 4.70 & 5.71 \\
\bottomrule
\end{tabular}
}
\end{table}

\subsection{HP Sensitivity Analysis}
\label{assec:hp_sensitivity}

\begin{figure*}[htbp]
    \centering
    \hspace{-0.2in}
    \begin{minipage}[t]{0.495\linewidth}
        \centering
        \includegraphics[width=\linewidth]{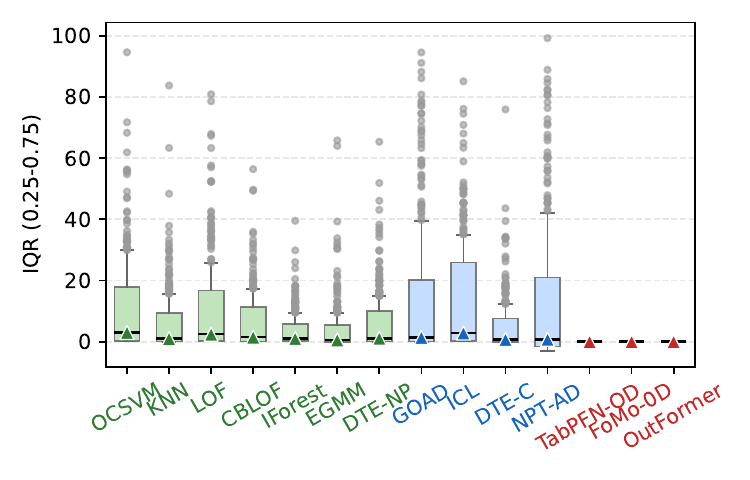}
        {\small\textbf{(a)} AUPRC on \fraudbench}
    \end{minipage}
    \hspace{-0.2in}
    \begin{minipage}[t]{0.495\linewidth}
        \centering
        \includegraphics[width=\linewidth]{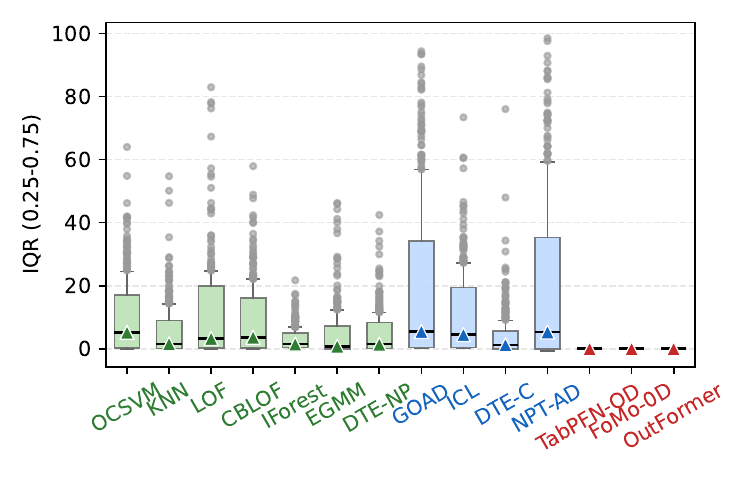}
        {\small\textbf{(b)} AUROC on \fraudbench}
    \end{minipage}

    \begin{minipage}[t]{0.495\linewidth}
        \centering
        \includegraphics[width=\linewidth]{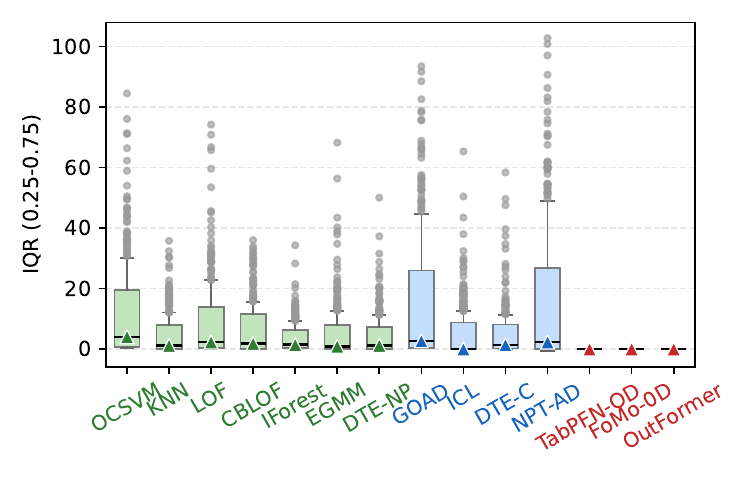}
        {\small\textbf{(c)} AUPRC on \ovrbench}
    \end{minipage}
    \hspace{-0.2in}
    \begin{minipage}[t]{0.495\linewidth}
        \centering
        \includegraphics[width=\linewidth]{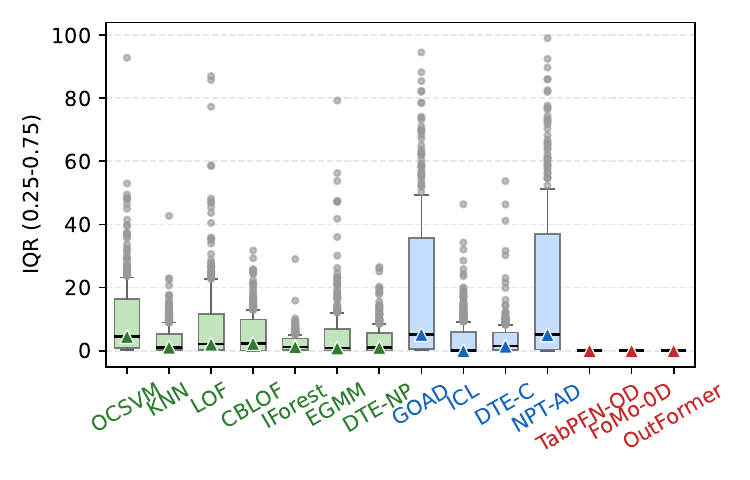}
        {\small\textbf{(d)} AUROC on \ovrbench}
    \end{minipage}

    \begin{minipage}[t]{0.495\linewidth}
        \centering
        \includegraphics[width=\linewidth]{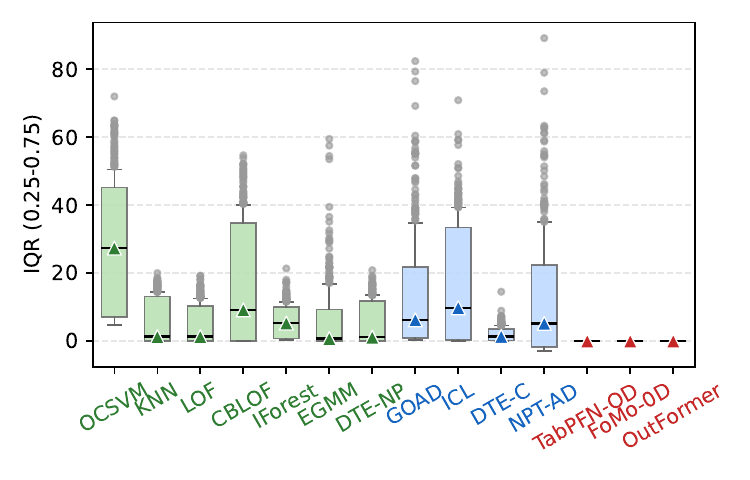}
        {\small\textbf{(e)} AUPRC on \synbench}
    \end{minipage}
    \hspace{-0.2in}
    \begin{minipage}[t]{0.495\linewidth}
        \centering
        \includegraphics[width=\linewidth]{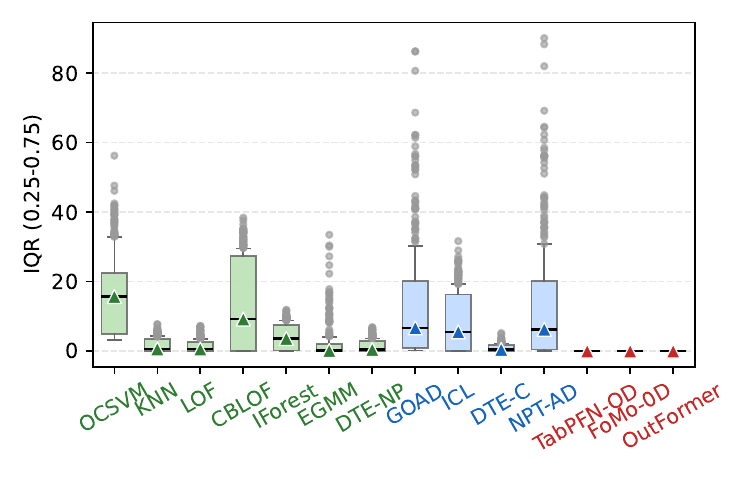}
        {\small\textbf{(f)} AUROC on \synbench}
    \end{minipage}

    \vspace{-0.05in}
    \caption{Hyperparameter sensitivity as
measured by inter-quartile 25\%-75\% performance; the higher, the
more HP-sensitive. 
    Subfigures (a)--(f) report AUPRC and AUROC performance variation under different hyperparameter configurations across all datasets (boxplot) per baseline,
    on \fraudbench, \ovrbench, and \synbench.}
    \label{fig:hp_sensitivity_ind}
\end{figure*}

Figure~\ref{fig:hp_sensitivity_ind} illustrates the inter-quartile range of performance variations induced by different hyperparameter configurations, highlighting each method’s sensitivity  to hyperparameter (HP) choices. Among shallow models, ensemble based IForest and  EGMM exhibit relatively strong HP robustness. In contrast, deep models are generally more sensitive to HP settings, due to the large hyperparameter spaces introduced by architectural design, optimization and regularization. Foundation models (FMs) fully mitigate this issue by adopting a unified ``one-model-for-all'' paradigm, eliminating the need for per-task hyperparameter selection. Combined with low-latency inference, this makes FMs truly plug-and-play for practical deployment.

One may wonder whether strong conclusions about the relative underperformance of deep OD methods are affected by evaluating each method with only five random hyperparameter configurations. We acknowledge that computational constraints are a practical limitation, as our benchmark spans more than 2,400 datasets. Nevertheless, even under this limited setting, deep OD models already exhibit substantial performance variation across hyperparameter choices, highlighting their pronounced sensitivity. Importantly, increasing the number of evaluated configurations would expose more of the underlying hyperparameter-performance landscape, and is therefore expected to reveal greater, rather than smaller, sensitivity. This observation is also consistent with our prior work, where we showed that deep OD models remain highly sensitive when evaluated over much larger hyperparameter grids \cite{ding2022hyperparameter}.

\begin{table*}[htbp]
\centering
\caption{Overall AUPRC performance comparison across models on the public partition of \fraudbench (690 datasets). Colors depict OD method categories, where  
\textcolor{darkgreen}{green}, \textcolor{darkblue}{blue}, and \textcolor{darkred}{red} respectively denote shallow (classical), deep, and foundation models.
Best results of each category are shown in bold, while global best is \underline{underlined}. \outform achieves the strongest overall performance, underscoring the effectiveness of foundation models. }
\vspace{-0.1in}
\label{tab:oddbench_results_auprc}
\resizebox{0.65\linewidth}{!}{
\begin{tabular}{clccccc}
\toprule
& \textbf{Model} & \textbf{Avg. Rank ($\downarrow$)} & \textbf{ELO ($\uparrow$)} &
\textbf{Winrate ($\uparrow$)} & \textbf{rAUC ($\uparrow$)} &
\textbf{$C_{\Delta}$ ($\downarrow$)} \\
\midrule

\multirow{7}{*}{\rotatebox{90}{\textbf{Shallow}}}
& \textbf{\textcolor{darkgreen}{OCSVM}} & \pmv{6.42}{2.8} & \textbf{1180} & 0.57 & \pmv{0.694}{0.26} & 0.35 \\
& \textbf{\textcolor{darkgreen}{KNN}} & \pmv{6.76}{3.3} & 861 & 0.53 & \pmv{0.684}{0.28} & 0.31 \\
& \textbf{\textcolor{darkgreen}{LOF}} & \pmv{6.86}{3.9} & 915 & 0.53 & \pmv{0.676}{0.28} & 0.33 \\
& \textbf{\textcolor{darkgreen}{CBLOF}} & \pmv{7.98}{3.6} & 786 & 0.44 & \pmv{0.632}{0.30} & 0.35 \\
& \textbf{\textcolor{darkgreen}{IForest}} & \pmv{8.43}{4.1} & 1072 & 0.43 & \pmv{0.589}{0.31} & 0.39 \\
& \textbf{\textcolor{darkgreen}{EGMM}} & \textbf{\pmv{5.83}{4.1}} & 1135 & \textbf{0.60} & \textbf{\pmv{0.724}{0.30}} & \textbf{0.25} \\
& \textbf{\textcolor{darkgreen}{DTE-NP}} & \pmv{5.91}{3.3} & 1018 & \textbf{0.60} & \pmv{0.710}{0.28} & 0.30 \\

\midrule
\multirow{4}{*}{\rotatebox{90}{\textbf{Deep}}}
& \textbf{\textcolor{darkblue}{GOAD}} & \pmv{10.98}{3.6} & 741 & 0.22 & \pmv{0.483}{0.29} & 0.41 \\
& \textbf{\textcolor{darkblue}{ICL}} & \pmv{8.10}{3.7} & 1105 & 0.44 & \pmv{0.642}{0.28} & 0.36 \\
& \textbf{\textcolor{darkblue}{DTE-C}} & \textbf{\pmv{7.64}{3.8}} & \textbf{1199} & \textbf{0.47} & \textbf{\pmv{0.651}{0.29}} & \textbf{0.33} \\
& \textbf{\textcolor{darkblue}{NPT-AD}} & \pmv{10.38}{4.1} & 671 & 0.27 & \pmv{0.502}{0.31} & 0.40 \\

\midrule
\multirow{3}{*}{\rotatebox{90}{\textbf{FM}}}
& \textbf{\textcolor{darkred}{TabPFN-OD}} & \pmv{6.12}{4.0} & 929 & 0.58 & \pmv{0.716}{0.29} & 0.26 \\
& \textbf{\textcolor{darkred}{FoMo-0D}} & \pmv{6.51}{4.4} & \underline{\textbf{1207}} & 0.56 & \pmv{0.721}{0.29} & 0.25 \\
& \textbf{\textcolor{darkred}{OutFormer}} & \underline{\textbf{\pmv{5.43}{3.9}}} & 1180 & \underline{\textbf{0.63}} & \underline{\textbf{\pmv{0.739}{0.28}}} & \underline{\textbf{0.23}} \\
\bottomrule
\end{tabular}
}
\end{table*}

\begin{table*}[htbp]
\centering
\caption{Overall AUPRC performance comparison across models on the public partition of \onevsrestbench (756 datasets). Colors depict OD method categories, where  
\textcolor{darkgreen}{green}, \textcolor{darkblue}{blue}, and \textcolor{darkred}{red} respectively denote shallow (classical), deep, and foundation models.
Best results of each category are shown in bold, while global best is \underline{underlined}. EGMM and  \outform achieve the strongest overall results, significantly outperforming the rest of the baselines.}
\vspace{-0.1in}
\label{tab:ovrbench_results_auprc}
\resizebox{0.65\linewidth}{!}{
\begin{tabular}{clccccc}
\toprule
& \textbf{Model} & \textbf{Avg. Rank ($\downarrow$)} & \textbf{ELO ($\uparrow$)} &
\textbf{Winrate ($\uparrow$)} & \textbf{rAUC ($\uparrow$)} &
\textbf{$C_{\Delta}$ ($\downarrow$)} \\
\midrule

\multirow{7}{*}{\rotatebox{90}{\textbf{Shallow}}}
& \textbf{\textcolor{darkgreen}{OCSVM}} & \pmv{6.97}{2.9} & 1127 & 0.53 & \pmv{0.791}{0.18} & 0.30 \\
& \textbf{\textcolor{darkgreen}{KNN}} & \pmv{6.51}{3.4} & \underline{\textbf{1254}} & 0.56 & \pmv{0.805}{0.20} & 0.26 \\
& \textbf{\textcolor{darkgreen}{LOF}} & \pmv{6.97}{3.9} & 1062 & 0.53 & \pmv{0.775}{0.21} & 0.28 \\
& \textbf{\textcolor{darkgreen}{CBLOF}} & \pmv{7.83}{3.7} & 1221 & 0.47 & \pmv{0.758}{0.22} & 0.31 \\
& \textbf{\textcolor{darkgreen}{IForest}} & \pmv{7.99}{4.2} & 1167 & 0.46 & \pmv{0.744}{0.23} & 0.33 \\
& \textbf{\textcolor{darkgreen}{EGMM}} &  \underline{\textbf{\pmv{5.57}{3.8}}} & 1195 & \underline{\textbf{0.64}} & \underline{\textbf{\pmv{0.828}{0.20}}} & \textbf{0.22} \\
& \textbf{\textcolor{darkgreen}{DTE-NP}} & \pmv{5.94}{3.4} & 1233 & 0.61 & \pmv{0.816}{0.19} & 0.25 \\

\midrule
\multirow{4}{*}{\rotatebox{90}{\textbf{Deep}}}
& \textbf{\textcolor{darkblue}{GOAD}} & \pmv{11.27}{3.4} & 581 & 0.21 & \pmv{0.590}{0.24} & 0.40 \\
& \textbf{\textcolor{darkblue}{ICL}} & \textbf{\pmv{7.26}{3.5}} & \textbf{951} & \textbf{0.43} & \textbf{\pmv{0.770}{0.20}} & \textbf{0.29} \\
& \textbf{\textcolor{darkblue}{DTE-C}} & \textbf{\pmv{7.26}{3.5}} & 939 & \textbf{0.43} & \textbf{\pmv{0.770}{0.19}} & \textbf{0.29} \\
& \textbf{\textcolor{darkblue}{NPT-AD}} &  \pmv{10.31}{4.2} & 356 & 0.28 & \pmv{0.627}{0.26} & 0.38 \\

\midrule
\multirow{3}{*}{\rotatebox{90}{\textbf{FM}}}
& \textbf{\textcolor{darkred}{TabPFN-OD}} & \pmv{6.22}{4.0} & 877 & 0.59 & \pmv{0.815}{0.20} & 0.23 \\
& \textbf{\textcolor{darkred}{FoMo-0D}} & \pmv{7.18}{4.5} & 997 & 0.52 & \pmv{0.783}{0.21} & 0.26\\
& \textbf{\textcolor{darkred}{OutFormer}} &
\textbf{\pmv{5.88}{3.9}} & \textbf{1039} & \textbf{0.61} & \textbf{\pmv{0.821}{0.19}} & \underline{\textbf{0.21}} \\

\bottomrule
\end{tabular}
}
\end{table*}

\begin{table*}[htbp]
\centering
\caption{Overall AUPRC performance comparison across models on the public partition of \synbench (800 datasets). Colors depict OD method categories, where  
\textcolor{darkgreen}{green}, \textcolor{darkblue}{blue}, and \textcolor{darkred}{red} respectively denote shallow (classical), deep, and foundation models.
Best results of each category are shown in bold, while global best is \underline{underlined}.  \outform and \tabod achieve the strongest overall results, significantly outperforming the rest of the baselines.}
\vspace{-0.1in}
\label{tab:synbench_results_auprc}
\resizebox{0.65\linewidth}{!}{
\begin{tabular}{clccccc}
\toprule
& \textbf{Model} & \textbf{Avg. Rank ($\downarrow$)} & \textbf{ELO ($\uparrow$)} &
\textbf{Winrate ($\uparrow$)} & \textbf{rAUC ($\uparrow$)} &
\textbf{$C_{\Delta}$ ($\downarrow$)} \\
\midrule

\multirow{7}{*}{\rotatebox{90}{\textbf{Shallow}}}
& \textbf{\textcolor{darkgreen}{OCSVM}} & \pmv{10.77}{1.8} & 370 & 0.25 & \pmv{0.687}{0.16} & 0.88 \\
& \textbf{\textcolor{darkgreen}{KNN}} & \pmv{6.77}{2.2} & 1036 & 0.54 & \pmv{0.800}{0.18} & 0.74 \\
& \textbf{\textcolor{darkgreen}{LOF}} & \pmv{4.92}{1.9} & 1266 & 0.68 & \pmv{0.855}{0.15} & 0.69 \\
& \textbf{\textcolor{darkgreen}{CBLOF}} & \pmv{9.37}{2.8} & 893 & 0.34 & \pmv{0.638}{0.30} & 0.80 \\
& \textbf{\textcolor{darkgreen}{IForest}} & \pmv{12.02}{1.6} & 633 & 0.15 & \pmv{0.553}{0.30} & 0.89 \\
& \textbf{\textcolor{darkgreen}{EGMM}} & \textbf{\pmv{3.40}{2.1}} & \textbf{1538} & \textbf{0.80 }& \textbf{\pmv{0.931}{0.11}} & \textbf{0.53} \\
& \textbf{\textcolor{darkgreen}{DTE-NP}} & \pmv{6.47}{1.9} & 1026 & 0.56 & \pmv{0.814}{0.17} & 0.75 \\

\midrule
\multirow{4}{*}{\rotatebox{90}{\textbf{Deep}}}
& \textbf{\textcolor{darkblue}{GOAD}} & \pmv{11.23}{2.2} & 673 & 0.21 & \pmv{0.591}{0.29} & 0.87 \\
& \textbf{\textcolor{darkblue}{ICL}} & \pmv{8.31}{2.8} & \textbf{1181} & 0.44 & \pmv{0.744}{0.21} & 0.81 \\
& \textbf{\textcolor{darkblue}{DTE-C}} & \textbf{\pmv{5.58}{2.8}} & 1091 & \textbf{0.64} & \textbf{\pmv{0.904}{0.10}} & \textbf{0.71} \\
& \textbf{\textcolor{darkblue}{NPT-AD}} & \pmv{13.43}{1.3} & 56 & 0.04 & \pmv{0.426}{0.25} & 0.92 \\

\midrule
\multirow{3}{*}{\rotatebox{90}{\textbf{FM}}}
& \textbf{\textcolor{darkred}{TabPFN-OD}} & \pmv{3.51}{3.5} & \underline{\textbf{1601}} & 0.79 & \pmv{0.920}{0.17} & \underline{\textbf{0.29}} \\
& \textbf{\textcolor{darkred}{FoMo-0D}} & \pmv{5.01}{2.7} & 1280 & 0.68 & \pmv{0.885}{0.14} & 0.63 \\
& \textbf{\textcolor{darkred}{OutFormer}} &
\underline{\textbf{\pmv{3.34}{3.1}}} & 1356 & \underline{\textbf{0.82}} & \underline{\textbf{\pmv{0.984}{0.04}}} & 0.41 \\

\bottomrule
\end{tabular}
}
\end{table*}

\begin{table*}[htbp]
\centering
\caption{Overall AUROC performance comparison across models on the public partition of \fraudbench (690 datasets). Colors depict OD method categories, where  
\textcolor{darkgreen}{green}, \textcolor{darkblue}{blue}, and \textcolor{darkred}{red} respectively denote shallow (classical), deep, and foundation models.
Best results of each category are shown in bold, while global best is \underline{underlined}. \outform and DTE-NP achieve the strongest overall performance, with foundation models generally offering the leading edge. }
\vspace{-0.1in}
\label{tab:oddbench_results}
\resizebox{0.65\linewidth}{!}{
\begin{tabular}{clccccc}
\toprule
& \textbf{Model} & \textbf{Avg. Rank ($\downarrow$)} & \textbf{ELO ($\uparrow$)} &
\textbf{Winrate ($\uparrow$)} & \textbf{rAUC ($\uparrow$)} &
\textbf{$C_{\Delta}$ ($\downarrow$)} \\
\midrule
\multirow{7}{*}{\rotatebox{90}{\textbf{Shallow}}}
& \textbf{\textcolor{darkgreen}{OCSVM}} & \pmv{7.44}{3.0} & 1007 & 0.49 & \pmv{0.871}{0.11} & 0.52 \\
& \textbf{\textcolor{darkgreen}{KNN}} & \pmv{6.10}{3.4} & 990 & 0.58 & \pmv{0.879}{0.15} & 0.43 \\
& \textbf{\textcolor{darkgreen}{LOF}} & \pmv{7.14}{4.0} & 954 & 0.50 & \pmv{0.848}{0.16} & 0.50 \\
& \textbf{\textcolor{darkgreen}{CBLOF}} & \pmv{7.73}{3.5} & 958 & 0.46 & \pmv{0.853}{0.14} & 0.50 \\
& \textbf{\textcolor{darkgreen}{IForest}} & \pmv{7.43}{3.9} & 865 & 0.50 & \pmv{0.865}{0.13} & 0.50 \\
& \textbf{\textcolor{darkgreen}{EGMM}} & \pmv{5.73}{4.2} & 1018 & 0.61 & \pmv{0.868}{0.19} & \underline{\textbf{0.37}} \\
& \textbf{\textcolor{darkgreen}{DTE-NP}} &
\underline{\textbf{\pmv{5.44}{3.4}}} & \textbf{1140} & \underline{\textbf{0.64}} & \underline{\textbf{\pmv{0.891}{0.15}}} & 0.41  \\
\midrule
\multirow{4}{*}{\rotatebox{90}{\textbf{Deep}}}
& \textbf{\textcolor{darkblue}{GOAD}} & \pmv{11.31}{3.7} & 731 & 0.19 & \pmv{0.658}{0.23} & 0.64 \\
& \textbf{\textcolor{darkblue}{ICL}} & \pmv{8.68}{3.5} & 944 & 0.40 & \pmv{0.828}{0.15} & 0.55 \\
& \textbf{\textcolor{darkblue}{DTE-C}} & \textbf{\pmv{7.11}{3.7}} & \textbf{1126} & \textbf{0.51} & \textbf{\pmv{0.860}{0.16}} & \textbf{0.48} \\
& \textbf{\textcolor{darkblue}{NPT-AD}} & \pmv{10.33}{4.0} & 642 & 0.28 & \pmv{0.736}{0.20} & 0.59 \\
\midrule
\multirow{3}{*}{\rotatebox{90}{\textbf{FM}}}
& \textbf{\textcolor{darkred}{TabPFN-OD}} & \pmv{6.07}{4.0} & 1200 & 0.58 & \pmv{0.886}{0.14} & 0.39 \\
& \textbf{\textcolor{darkred}{FoMo-0D}} & \pmv{7.02}{4.5} & 1165 & 0.51 & \pmv{0.857}{0.16} & 0.43 \\
& \textbf{\textcolor{darkred}{OutFormer}} &
\textbf{\pmv{5.62}{3.9}} & \underline{\textbf{1262}} & \textbf{0.61} & \textbf{\pmv{0.889}{0.14}} & \underline{\textbf{0.37}} \\
\bottomrule
\end{tabular}
    }
\end{table*}

\begin{table*}[htbp]
\centering
\caption{Overall AUROC performance comparison across models on the public partition of \onevsrestbench (756 datasets). Colors depict OD method categories, where  
\textcolor{darkgreen}{green}, \textcolor{darkblue}{blue}, and \textcolor{darkred}{red} respectively denote shallow (classical), deep, and foundation models.
Best results of each category are shown in bold, while global best is \underline{underlined}. The shallow methods KNN, EGMM, and DTE-NP, as well as the foundation models  \outform and \tabod achieve the strongest overall results, significantly outperforming the rest of the baselines while being comparable among themselves. }
\vspace{-0.1in}
\label{tab:ovrbench_results}
\resizebox{0.65\linewidth}{!}{
\begin{tabular}{clccccc}
\toprule
& \textbf{Model} & \textbf{Avg. Rank ($\downarrow$)} & \textbf{ELO ($\uparrow$)} &
\textbf{Winrate ($\uparrow$)} & \textbf{rAUC ($\uparrow$)} &
\textbf{$C_{\Delta}$ ($\downarrow$)} \\
\midrule

\multirow{7}{*}{\rotatebox{90}{\textbf{Shallow}}}
& \textbf{\textcolor{darkgreen}{OCSVM}} & \pmv{7.71}{3.2} & 1125 & 0.47 & \pmv{0.890}{0.09} & 0.40 \\
& \textbf{\textcolor{darkgreen}{KNN}} & \pmv{6.01}{3.5} & \underline{\textbf{1274}} & 0.60 & \underline{\textbf{\pmv{0.907}{0.12}}} & 0.30 \\
& \textbf{\textcolor{darkgreen}{LOF}} & \pmv{7.04}{3.8} & 1033 & 0.52 & \pmv{0.874}{0.15} & 0.37 \\
& \textbf{\textcolor{darkgreen}{CBLOF}} & \pmv{7.60}{3.5} & 1230 & 0.48 & \pmv{0.886}{0.12} & 0.38 \\
& \textbf{\textcolor{darkgreen}{IForest}} & \pmv{7.30}{4.0} & 1223 & 0.51 & \pmv{0.895}{0.11} & 0.37 \\
& \textbf{\textcolor{darkgreen}{EGMM}} & \underline{\textbf{\pmv{5.66}{4.0}}} & 1217 & \underline{\textbf{0.63}} & \pmv{0.906}{0.14} & \underline{\textbf{0.28}} \\
& \textbf{\textcolor{darkgreen}{DTE-NP}} & \pmv{5.85}{3.5} & 1211 & 0.62 & \pmv{0.906}{0.12} & 0.31 \\

\midrule
\multirow{4}{*}{\rotatebox{90}{\textbf{Deep}}}
& \textbf{\textcolor{darkblue}{GOAD}} & \pmv{11.48}{3.5} & 403 & 0.19 & \pmv{0.704}{0.20} & 0.55 \\
& \textbf{\textcolor{darkblue}{ICL}} & \textbf{\pmv{6.76}{3.4}} & \textbf{1021} & \textbf{0.47} & \textbf{\pmv{0.885}{0.12}} & \textbf{0.36} \\
& \textbf{\textcolor{darkblue}{DTE-C}} & \textbf{\pmv{6.76}{3.4}} & 1009 & \textbf{0.47} & \textbf{\pmv{0.885}{0.12}} & \textbf{0.36} \\
& \textbf{\textcolor{darkblue}{NPT-AD}} & \pmv{10.57}{3.9} & 443 & 0.26 & \pmv{0.777}{0.16} & 0.49 \\

\midrule
\multirow{3}{*}{\rotatebox{90}{\textbf{FM}}}
& \textbf{\textcolor{darkred}{TabPFN-OD}} & \pmv{6.39}{4.0} & 880 & 0.57 & \textbf{\pmv{0.903}{0.12}} & 0.30 \\
& \textbf{\textcolor{darkred}{FoMo-0D}} & \pmv{7.74}{4.4} & 907 & 0.47 & \pmv{0.867}{0.14} & 0.37 \\
& \textbf{\textcolor{darkred}{OutFormer}} & \textbf{\pmv{6.15}{4.0}} & \textbf{1024} & \textbf{0.59} & \textbf{\pmv{0.903}{0.11}} & \textbf{0.29} \\

\bottomrule
\end{tabular}
}
\end{table*}

\begin{table*}[htbp]
\centering
\caption{Overall AUROC performance comparison across models on the public partition of \synbench (800 datasets). Colors depict OD method categories, where  
\textcolor{darkgreen}{green}, \textcolor{darkblue}{blue}, and \textcolor{darkred}{red} respectively denote shallow (classical), deep, and foundation models.
Best results of each category are shown in bold, while global best is \underline{underlined}.  \outform, \tabod and EGMM achieve the strongest overall results, significantly outperforming the rest of the baselines.
}
\vspace{-0.1in}
\label{tab:synbench_results}
\resizebox{0.65\linewidth}{!}{
\begin{tabular}{clccccc}
\toprule
& \textbf{Model} & \textbf{Avg. Rank ($\downarrow$)} & \textbf{ELO ($\uparrow$)} &
\textbf{Winrate ($\uparrow$)} & \textbf{rAUC ($\uparrow$)} &
\textbf{$C_{\Delta}$ ($\downarrow$)} \\
\midrule

\multirow{7}{*}{\rotatebox{90}{\textbf{Shallow}}}
& \textbf{\textcolor{darkgreen}{OCSVM}} & \pmv{11.30}{1.6} & 397 & 0.21 & \pmv{0.827}{0.07} & 0.91 \\
& \textbf{\textcolor{darkgreen}{KNN}} & \pmv{6.52}{2.1} & 1039 & 0.55 & \pmv{0.929}{0.07} & 0.73\\
& \textbf{\textcolor{darkgreen}{LOF}} & \pmv{5.18}{2.1} & 1151 & 0.65 & \pmv{0.941}{0.06} & 0.68 \\
& \textbf{\textcolor{darkgreen}{CBLOF}} & \pmv{9.68}{3.0} & 874 & 0.31 & \pmv{0.839}{0.14} & 0.80 \\
& \textbf{\textcolor{darkgreen}{IForest}} & \pmv{11.42}{1.7} & 698 & 0.20 & \pmv{0.817}{0.14} & 0.89 \\
& \textbf{\textcolor{darkgreen}{EGMM}} & \underline{\textbf{\pmv{3.09}{2.0}}} & \textbf{1515} & \underline{\textbf{0.81}} & \textbf{\pmv{0.978}{0.04}} & \textbf{0.47}  \\
& \textbf{\textcolor{darkgreen}{DTE-NP}} & \pmv{5.84}{1.9} & 1066 & 0.60 & \pmv{0.932}{0.07} & 0.72 \\

\midrule
\multirow{4}{*}{\rotatebox{90}{\textbf{Deep}}}
& \textbf{\textcolor{darkblue}{GOAD}} & \pmv{11.44}{2.2} & 556 & 0.19 & \pmv{0.800}{0.13} & 0.89 \\
& \textbf{\textcolor{darkblue}{ICL}} & \pmv{8.46}{2.7} & 1178 & 0.42 & \pmv{0.888}{0.10} & 0.82 \\
& \textbf{\textcolor{darkblue}{DTE-C}} & \textbf{\pmv{5.01}{2.5}} & \textbf{1415} & \textbf{0.67} & \textbf{\pmv{0.966}{0.04}} & \textbf{0.67} \\
& \textbf{\textcolor{darkblue}{NPT-AD}} & \pmv{13.55}{1.3} & 138 & 0.03 & \pmv{0.664}{0.16} & 0.94 \\

\midrule
\multirow{3}{*}{\rotatebox{90}{\textbf{FM}}}
& \textbf{\textcolor{darkred}{TabPFN-OD}} & \pmv{3.36}{3.2} & \underline{\textbf{1522}} & 0.79 & \pmv{0.969}{0.07} & \underline{\textbf{0.30}} \\
& \textbf{\textcolor{darkred}{FoMo-0D}} & \pmv{5.34}{2.8} & 1104 & 0.65 & \pmv{0.950}{0.06} & 0.63 \\
& \textbf{\textcolor{darkred}{OutFormer}} &
\textbf{\pmv{3.30}{2.8}} & 1347 & \underline{\textbf{0.81}} & \underline{\textbf{\pmv{0.993}{0.01}}} & 0.42 \\
\bottomrule
\end{tabular}
}
\end{table*}

\begin{table*}[htbp]
\centering
\caption{AUROC performance comparison across models on overall 2,446 datasets across three benchmarks.. Colors depict OD method categories, where  
\textcolor{darkgreen}{green}, \textcolor{darkblue}{blue}, and \textcolor{darkred}{red} respectively denote shallow (classical), deep, and foundation models.
Best results of each category are shown in bold, while \underline{overall best} is {underlined}.  
}
\vspace{-0.1in}
\label{tab:overall_results_auroc}
\resizebox{0.65\linewidth}{!}{
\begin{tabular}{clccccc}
\toprule
& \textbf{Model} & \textbf{Avg. Rank ($\downarrow$)} & \textbf{ELO ($\uparrow$)} &
\textbf{Winrate ($\uparrow$)} & \textbf{rAUC ($\uparrow$)} &
\textbf{$C_{\Delta}$ ($\downarrow$)} \\
\midrule

\multirow{7}{*}{\rotatebox{90}{\textbf{Shallow}}}
& \textbf{\textcolor{darkgreen}{OCSVM}} & \pmv{8.91}{3.2} & 397 & 0.38 & \pmv{0.862}{0.09} & 0.62\\
& \textbf{\textcolor{darkgreen}{KNN}} & \pmv{6.22}{3.1} & 1037 & 0.57 & \pmv{0.906}{0.12} & 0.49 \\
& \textbf{\textcolor{darkgreen}{LOF}} & \pmv{6.41}{3.5} & 1153 & 0.56 & \pmv{0.890}{0.13} & 0.52 \\
& \textbf{\textcolor{darkgreen}{CBLOF}} & \pmv{8.38}{3.5} & 874 & 0.41 & \pmv{0.859}{0.13} & 0.57 \\
& \textbf{\textcolor{darkgreen}{IForest}} & \pmv{8.81}{3.8} & 699 & 0.40 & \pmv{0.858}{0.13} & 0.59 \\
& \textbf{\textcolor{darkgreen}{EGMM}} & \underline{\textbf{\pmv{4.76}{3.7}}} & \textbf{1521} & \underline{\textbf{0.69}} & \textbf{\pmv{0.920}{0.14}} & \textbf{0.38} \\
& \textbf{\textcolor{darkgreen}{DTE-NP}} & \pmv{5.72}{3.0} & 1066 & 0.62 & \pmv{0.911}{0.12} & 0.49 \\

\midrule
\multirow{4}{*}{\rotatebox{90}{\textbf{Deep}}}
& \textbf{\textcolor{darkblue}{GOAD}} & \pmv{11.41}{3.2} & 557 & 0.19 & \pmv{0.724}{0.20} & 0.70 \\
& \textbf{\textcolor{darkblue}{ICL}} & \pmv{7.95}{3.3} & 1176 & 0.43 & \pmv{0.869}{0.13} & 0.58 \\
& \textbf{\textcolor{darkblue}{DTE-C}} & \textbf{\pmv{6.24}{3.4}} & \textbf{1411} & \textbf{0.55} & \textbf{\pmv{0.906}{0.12}} & \textbf{0.51} \\
& \textbf{\textcolor{darkblue}{NPT-AD}} & \pmv{11.56}{3.6} & 138 & 0.18 & \pmv{0.724}{0.18} & 0.68 \\

\midrule
\multirow{3}{*}{\rotatebox{90}{\textbf{FM}}}
& \textbf{\textcolor{darkred}{TabPFN-OD}} & \pmv{5.21}{4.0} & \underline{\textbf{1522}} & 0.65 & \pmv{0.921}{0.12} & \underline{\textbf{0.33}} \\
& \textbf{\textcolor{darkred}{FoMo-0D}} & \pmv{6.67}{4.1} & 1108 & 0.55 & \pmv{0.894}{0.13} & 0.48 \\
& \textbf{\textcolor{darkred}{OutFormer}} &
\textbf{\pmv{4.97}{3.8}} & 1342 & \textbf{0.68} & \underline{\textbf{\pmv{0.931}{0.11}}} & 0.36 \\
\bottomrule
\end{tabular}
}
\end{table*}

\end{document}